\documentclass{article}

\usepackage{silence}
\usepackage{microtype}

\PassOptionsToPackage{numbers, sort&compress}{natbib}
\usepackage[preprint]{neurips_2026}

\usepackage{tikz}
\usepackage{graphicx}
\usepackage{xcolor}
\usetikzlibrary{arrows.meta, positioning, fit, calc}
\usetikzlibrary{shapes.geometric}

\makeatletter
\def\blfootnote{\xdef\@thefnmark{}\@footnotetext}
\makeatother

\usepackage{xcolor}

\newcommand{\new}[1]{\textcolor{black}{#1}}
\newcommand{\edited}[1]{\textcolor{black}{#1}}

\usepackage[utf8]{inputenc} 
\usepackage[T1]{fontenc}    
\usepackage{hyperref}       
\usepackage{url}            
\usepackage{booktabs}       
\usepackage{amsfonts}       
\usepackage{nicefrac}       
\usepackage{microtype}      
\usepackage{xcolor}         
\usepackage{graphicx}    
\usepackage{makecell}
\usepackage{float}          
\usepackage{amsmath, amssymb}
\usepackage{cleveref}
\usepackage{tikz}
\usetikzlibrary{arrows.meta, positioning, fit, backgrounds}
\usepackage{subcaption}
\usepackage{multirow}

\title{What's in an Earth Embedding? An Explainability Analysis of Location Encoders}

\author{%
  \begin{minipage}{\linewidth}
    \centering
    \textbf{Livia Betti}$^{1,*}$
    \quad
    \textbf{Sebastian Ricke}$^{1,*}$
    \quad
    \textbf{Ivica Obadic}$^{2,3,*}$
    \quad
    \textbf{Adam J. Stewart}$^{2}$
    \quad
    \textbf{Esther Rolf}$^{1}$
    \\[0.5em]
    \normalfont$^{1}$University of Colorado Boulder
    \quad
    $^{2}$Technical University of Munich
    \\[0.25em]
    $^{3}$Munich Center for Machine Learning
    \\[0.25em]
    $^{*}$Equal Contribution
  \end{minipage}
}

\date{}

\begin{document}

\maketitle

\blfootnote{Correspondence to: \texttt{livia.betti@colorado.edu}}

\begin{abstract}
Geographic implicit neural representations (INRs) learn to map any coordinate on Earth to a location embedding, implicitly encoding geospatial data into the weights of a neural network. Location embeddings are widely used off the shelf as general-purpose geospatial representations, yet users lack principled tools to audit what geographic or semantic information these embeddings capture. In this work, we analyze the information content of geographic INRs through their location embeddings. We decompose these embeddings into human-interpretable features---namely, (i) sparse latent concepts, (ii) natural language concepts, and (iii) visual features. The latent concept embeddings are learned using sparse autoencoders. To recover natural language concepts, we apply sparse linear concept embeddings (SpLiCE) over a predefined geospatial dictionary. Finally, visual features are extracted using saliency maps derived from CLIP Surgery. We show that location embeddings can be decomposed into human-interpretable representations while retaining high reconstruction capability, revealing interpretable geographic structures such as forests, deserts, and urban features. Across methods, sparse decompositions expose systematic differences in encoded information, ranging from urban structures to broader biome and climate signals, and pretraining-space saliency maps further highlight complementary features such as roads and landmarks. We hope this work provides a first step toward interpretable geospatial representations.
\end{abstract}

\section{Introduction}
\label{sec: intro}

The archive of geospatial data is large and growing, including remote sensing images, meteorological measurements, and geotagged social media posts~\cite{zhu2026foundations}. The availability of this data across space and time enables a range of tasks such as weather forecasting~\citep{lam2023learning}, tree canopy mapping~\citep{tolan2024very}, and disaster response~\citep{ivic2019artificial}. The massive amount of geospatial data---generally considered a barrier-to-entry for those without resources---has encouraged the use of machine learning to develop compact, general-purpose geospatial representations \new{known as Earth embeddings}. \new{These representations can be learned either directly from Earth observation data or indirectly through geographic implicit neural representations (INRs)~\citep{klemmer2025earth}.} \new{In this work, we focus on geographic INRs,} which encode geospatial data in the weights of a neural network called a \textit{location encoder} that maps geographic coordinates (latitude, longitude) to a representation called a \textit{location embedding}~\cite{mai2022review,klemmer2025earth}. Location embeddings serve as succinct representations of the geospatial information at a given location, and are increasingly used in downstream geospatial tasks, such as temperature mapping~\citep{klemmer2025satclip}, air quality prediction~\cite{wang2026proxy}, and species presence and identification~\citep{sinr}. \new{However, the information content of location embeddings is particularly hard to interpret as these embeddings do not correspond to a single input image, but rather to a specific latitude/longitude coordinate, and currently, there is no principled way to determine what information these representations encode.}

Explainability research in remote sensing has largely focused on traditional supervised image models, leaving geographic INRs underexplored \edited{despite the fundamental differences inherent to training such architectures}. \edited{For geographic INRs, the dominant method for assessing embedding quality relies on predictive performance on benchmark tasks, which does not address \textit{why} location embeddings perform well, or \textit{when} they may fail on unseen prediction tasks. For example, in applications such as poverty mapping, it can be unclear whether predictions derived from Earth embeddings use meaningful indicators of poverty or instead rely on more predictable proxies such as infrastructure, which is seen more readily in satellite imagery~\citep{aiken2023fairness}.} \new{To address this ambiguity, we require more direct ways of interrogating what information location embeddings contain.}

We aim to interpret the information content of location embeddings through the decomposition into human-interpretable features, providing three perspectives through 1) latent concepts, 2) natural language concepts, and 3) visual features (\Cref{fig:method_overview}). We first extract latent concepts using sparse dictionary learning methods and find that the resulting sparse decompositions can differentiate between geographic concepts while reliably reconstructing the original location embedding (\Cref{sec: monosemantic emb}). We then extend these explanations to use natural language concepts in an aligned location--text space. Qualitative analysis emphasizes fine-grained urban signals for some location encoders, and broad geographic attributes like climate or biome for others (\Cref{sec: text concepts}). Finally, we examine the visual features learned from the training data by using saliency maps, highlighting \edited{the differences in attention between natural and satellite imagery} (\Cref{sec: image space}). 
Taken together, these three explainability methods provide complementary but consistent interpretations of the types of information contained within location embeddings. We conclude in \Cref{sec: discussion} with a discussion of what this means for the current landscape of explainability methods that can be readily applied to understand Earth embeddings and location encoders, and where gaps exist that future work should address.

\definecolor{encblue}{RGB}{83,74,255}
\definecolor{embred}{RGB}{153,60,29}
\definecolor{coralred}{RGB}{153,60,29}

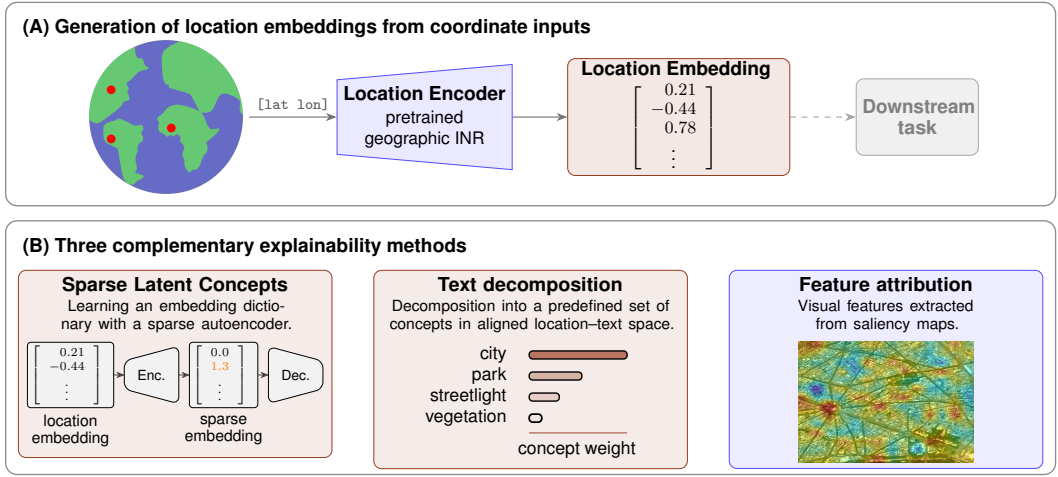
\begin{figure}[t!]
\centering
\resizebox{\textwidth}{!}{%
\begin{tikzpicture}[
    font=\normalsize\sffamily,
    >=Stealth,
    encoder/.style={
        shape=trapezium,
        trapezium angle=85,
        shape border rotate=90,
        draw=encblue,
        fill=encblue!12,
        line width=0.4pt,
        minimum width=18mm,
        minimum height=14mm,
        align=center
    },
    embedding/.style={
        draw=embred, fill=embred!12, rounded corners=4pt,
        align=center,
        minimum height=14mm,
        inner sep=1pt,
        line width=0.4pt
    },
    method/.style={
        rounded corners=4pt, text width=55mm,
        align=center, minimum height=1mm, line width=0.4pt, font=\normalsize\sffamily
    },
    methodred/.style={method, draw=embred,  fill=embred!10},
    methodblue/.style={method, draw=encblue,   fill=encblue!10},
    downstream/.style={
        draw=black!35, fill=black!6, rounded corners=4pt, align=center,
        minimum height=14mm, text width=20mm, line width=0.4pt,
        text=black!50, font=\normalsize\sffamily
    },
    output/.style={rounded corners=3pt, inner sep=2pt, line width=0.4pt},
    arrow/.style={  ->, line width=0.6pt, color=black!55},
    barrow/.style={ ->, line width=0.5pt, color=black!45},
    lbl/.style={font=\footnotesize, color=black!65},
    top_container/.style={draw=black!45, rounded corners=6pt, line width=0.7pt, inner sep=1mm, minimum height=36mm},
    bottom_container/.style={draw=black!45, rounded corners=6pt, line width=0.7pt, inner sep=1mm, minimum height=12mm},
    contlbl/.style={font=\bfseries\sffamily, color=black!75},
]


\node[opacity=0.6, yshift=-20mm] (globe) {%
    \includegraphics[width=28mm]{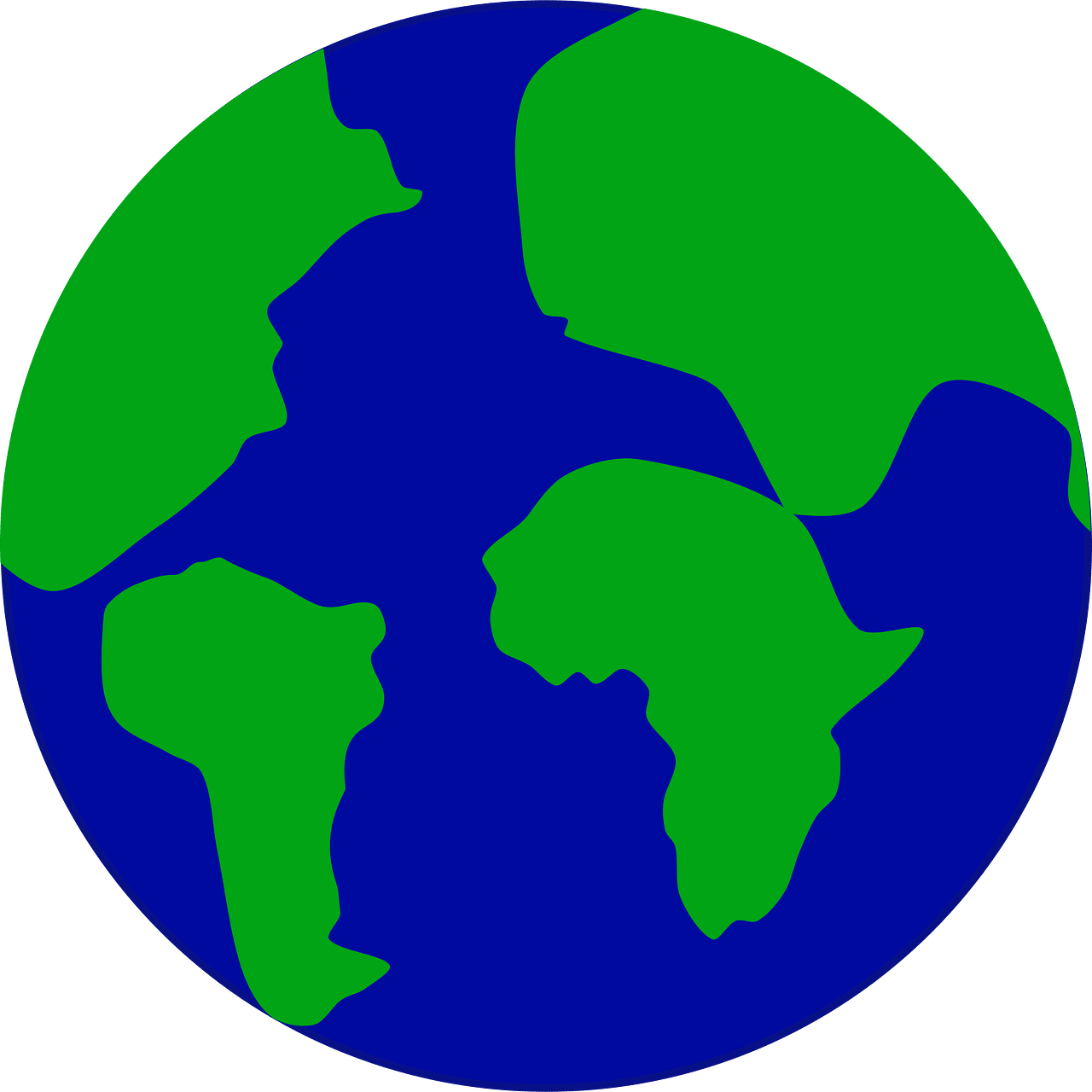}%
};
\fill[red] ($(globe.center)+(1mm,-2mm)$) circle (0.8mm);
\fill[red] ($(globe.center)+(-10mm, 5mm)$) circle (0.8mm);
\fill[red] ($(globe.center)+(-10mm,-4mm)$) circle (0.8mm);

\node[encoder, right=16mm of globe] (enc) {%
    \textbf{Location Encoder}\\[1pt]
    \small pretrained \\ \small geographic INR %
};

\node[embedding, right=10mm of enc] (emb) {%
\begin{tabular}{c}
\textbf{Location Embedding}\\
{\small
$
\left[
\begin{array}{r}
0.21\\
-0.44\\
0.78\\
\multicolumn{1}{c}{\vdots}\\
\end{array}
\right]
$
}
\end{tabular}
};

\node[downstream, right=12mm of emb] (down) {%
    \textbf{Downstream}\\\textbf{task}%
};

\draw[arrow] (globe.east) -- node[above, lbl]{\footnotesize \texttt{[lat lon]}} (enc.west);
\draw[arrow] (enc.east)   -- (emb.west);
\draw[->, line width=0.6pt, color=black!35, dashed] (emb.east) -- (down.west);


\coordinate (cx) at ($(globe.west)!0.5!(down.east)$);

\coordinate (A_l) at ($(cx)+(-95mm,  0mm)$);
\coordinate (A_r) at ($(cx)+( 95mm,  0mm)$);
\coordinate (B_l) at ($(cx)+(-95mm,-20mm)$);
\coordinate (B_r) at ($(cx)+( 95mm,-20mm)$);

\coordinate (row2) at ($(cx)+(0,-28mm)$);

\node[methodblue, anchor=north] (feat) at ($(row2)+(65mm,0)$) {%
    \textbf{Feature attribution}\\[1pt]
    \footnotesize Visual features extracted from saliency maps.\\[4pt]
    \includegraphics[width=32mm, height=22mm, keepaspectratio=false]{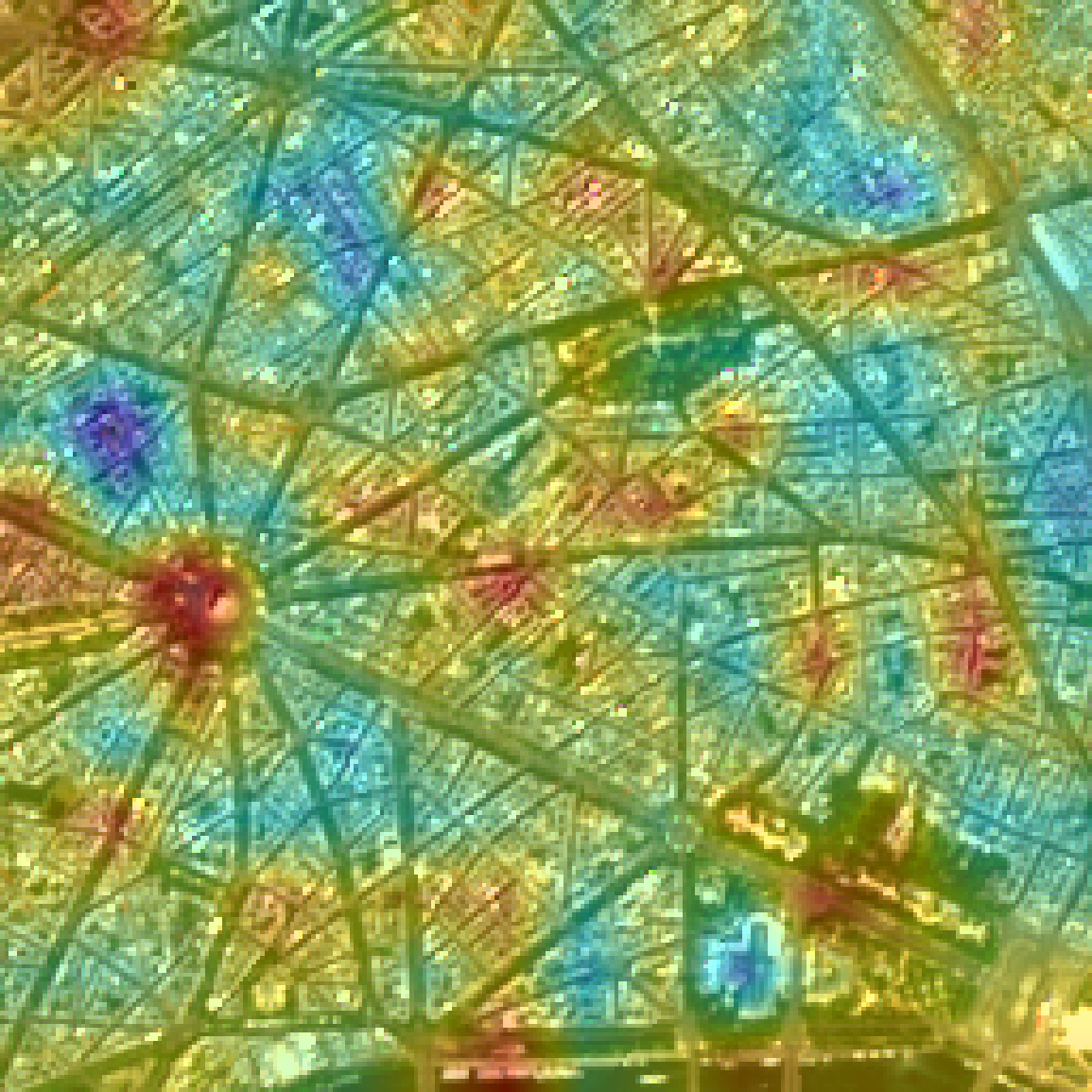}%
};

\node[methodred, anchor=north] (text) at (row2) {%
    \textbf{Text decomposition}\\[1pt]
    \footnotesize Decomposition into a predefined set of concepts in aligned location--text space.\\[4pt]
\resizebox{!}{22mm}{%
\begin{tikzpicture}[
    font=\normalsize\sffamily,
    bar/.style={rounded corners=2pt, fill=coralred!70, line width=0.8pt},
    barlight/.style={rounded corners=2pt, fill=coralred!40, line width=0.8pt},
    barlighter/.style={rounded corners=2pt, fill=coralred!25, line width=0.8pt},
    label/.style={anchor=east}
]

\def\dy{0.42}

\node[label] at (0,0) {city};
\node[label] at (0,-\dy) {park};
\node[label] at (0,-2*\dy) {streetlight};
\node[label] at (0,-3*\dy) {vegetation};

\draw[bar]        (0.35, 0-0.08)        rectangle (2.3, 0+0.08);
\draw[barlight]   (0.35,-\dy-0.08)      rectangle (1.4,-\dy+0.08);
\draw[barlighter] (0.35,-2*\dy-0.08)    rectangle (0.95,-2*\dy+0.08);

\draw[barlighter, fill opacity=0.2, draw opacity=1, line width=1.0pt]
    (0.35,-3*\dy-0.08) rectangle (0.6,-3*\dy+0.08);

\draw[coralred!70, thick] (0.35,-3.7*\dy) -- (2.3,-3.7*\dy);

\node[anchor=north] at (1.3,-3.85*\dy) {concept weight};

\end{tikzpicture}%
}%
};

\node[methodred, anchor=north] (mono) at ($(row2)+(-65mm,0)$) {%
    \textbf{Sparse Latent Concepts}\\[1pt]
    \footnotesize Learning an embedding dictionary with a sparse autoencoder.\\[4pt]
\resizebox{!}{20mm}{%
\begin{tikzpicture}[
    block/.style={
        draw,
        rounded corners=2pt,
        line width=0.4pt,
        fill=gray!10,
        align=center,
        inner sep=2pt
    },
    arrow/.style={->, line width=0.2pt, opacity=0.7},
    lbl/.style={font=\small\sffamily, color=black!65, align=center}
]

\node[block] (emb) {
{\scriptsize
$
\left[
\begin{array}{r}
0.21\\
-0.44\\
\multicolumn{1}{c}{\vdots}\\
\end{array}
\right]
$
}
};
\node[lbl, below=0.3mm of emb, text=black] {location \\ embedding};

\node[
    shape=trapezium,
    trapezium angle=75,
    shape border rotate=90,
    draw,
    fill=gray!10,
    line width=0.4pt,
    minimum width=5mm,
    minimum height=11mm,
    right=2mm of emb
] (enc) {Enc.};

\node[block, right=2mm of enc] (lat) {
{\scriptsize
$
\left[
\begin{array}{r}
0.0\\
\color{orange}{1.3}\\
\multicolumn{1}{c}{\vdots}\\
\end{array}
\right]
$
}
};
\node[lbl, below=0.5mm of lat, text=black] {sparse \\ embedding};

\node[
    shape=trapezium,
    trapezium angle=75,
    shape border rotate=270,
    draw,
    fill=gray!10,
    line width=0.4pt,
    minimum width=5mm,
    minimum height=11mm,
    right=2mm of lat
] (dec) {Dec.};

\draw[arrow] (emb) -- (enc);
\draw[arrow] (enc) -- (lat);
\draw[arrow] (lat) -- (dec);

\end{tikzpicture}%
}%
};

\node[top_container, fit=(globe)(enc)(emb)(down)(A_l)(A_r)] (boxA) {};
\node[bottom_container, fit=(feat)(text)(mono)(B_l)(B_r)] (boxB) {};
\node[contlbl, anchor=north west, text=black] at ($(boxA.north west)+(2mm,-2mm)$) {(A) Generation of location embeddings from coordinate inputs};
\node[contlbl, anchor=north west, text=black] at ($(boxB.north west)+(2mm,-2mm)$) {(B) Three complementary explainability methods};

\end{tikzpicture}
}

\caption{
\new{Overview of our decomposition framework for geographic INRs.
A geographic INR maps coordinate inputs to a location embedding (top). We aim to
interpret pretrained location encoders through three complementary explainability methods (bottom):
sparse latent concepts, text decomposition, and feature attribution.
}}
\label{fig:method_overview}
\end{figure}

\section{Related Work}
\label{sec:related_work}

We situate our work within the literature of geographic INRs and explainability. For the explainability methods, we discuss in greater depth the methods that are most relevant to this work.

\subsection{Location Encoders and Location Embeddings}

A location encoder serves both as the model as well as a compact, queryable database of geospatial information~\citep{klemmer2025earth}. Location encoders typically pass a latitude--longitude coordinate to a positional encoding followed by a (small) neural network~\citep{mai2022review}. Existing location encoders have leveraged various positional encodings, including sinusoidal encodings (e.g., SINR and Climplicit~\citep{mac2019presence, sinr, climplicit}), random Fourier features (RFFs) (e.g., GeoCLIP~\citep{vivanco2023geoclip}), and spherical harmonics basis functions (e.g., SatCLIP~\citep{russwurm2023geographic, klemmer2025satclip}). Popular choices for the neural network that follows the positional encoder include feedforward MLPs, SIREN~\citep{sitzmann2020implicit} or RESIREN~\citep{climplicit} architectures.

\subsection{Explainability Methods for Geospatial ML}

The growing adoption of ML methods for geospatial data analysis has motivated increasing interest in explaining geospatial ML models~\citep{hohl2024survey}. Common explanation methods for image-based geospatial models include class activation mapping (CAM)~\citep{zhou2016learning} and Grad-CAM~\citep{selvaraju2017grad}, Local Interpretable Model-agnostic Explanations (LIME)~\citep{ribeiro2016should}, SHapley Additive exPlanations (SHAP)~\citep{lundberg2017unified}, and the use of attention mechanisms~\citep{bahdanau2015neural, vaswani2017attention}. Beyond image-based models, \citet{jia2025towards} integrate concept-bottleneck models to design interpretable geolocalization models.

While image-based remote sensing models and geolocalization frameworks have attracted some attention in explainability research, geographic INRs and their resulting location embeddings remain largely unexplored in this regard. Some recent work has begun to examine the information of pixelwise Earth embeddings, such as Google's AlphaEarth Foundations (AEF)~\citep{brown2025alphaearth}. \citet{benavides2026earth} apply feature selection methods to analyze the contribution of individual embedding dimensions. Beyond AEF, \citet{rao2025measuring} investigate the intrinsic dimension of various geographic INRs, proposing it as a quantitative measure of the amount of information they contain, but \edited{do not audit \textit{what} information is being represented}. Our work aims to fill gaps left by these previous studies, by developing a more detailed understanding of the types of information encoded in geographic INRs.

\subsection{General Explainability Methods}

\paragraph{Sparse Autoencoders.} Recent work in sparse dictionary learning aims to provide interpretations through latent concepts encoded in vector representations~\citep{shu2025survey}. These methods use sparse autoencoders (SAEs) to learn a set of relevant concepts in neural activations. More specifically, SAEs project embeddings into a higher-dimensional space under a sparsity constraint, encouraging the separation of features that are entangled in the original embedding space. SAEs have been successfully used to identify interpretable, monosemantic features in large language models~\cite{cunningham2023sparse} and vision--language models~\cite{pach2025sparse}, but have not to our knowledge been used to expose structures in location embeddings. Evaluation of these SAE-based methods for explainability typically involves manual inspection. \citet{pach2025sparse} additionally proposed a metric for measuring the monosemanticity of SAE activations on vision tasks, using the similarity of images that activate given neurons.

\paragraph{Natural Language Explanations.} In contrast to sparse dictionary learning, various methods aim to decompose representations into predefined natural language concepts. For example, \citet{hernandez2021natural} propose finding natural language concepts that maximize mutual information with an input. Other methods rely on a CLIP-aligned image--text space, proposing techniques such as aligning embedding to a CLIP space~\citep{moayeri2023text} or using CLIP to dissect a neural network~\citep{oikarinen2022clip}. In this work, we leverage on a recent approach which posits that representations in CLIP-aligned image--text space can be expressed as linear functions of single-concept text embeddings~\citep{bhalla2024interpreting}. Using this hypothesis, sparse linear concept embeddings (SpLiCE) represent CLIP image embeddings as sparse linear combinations of a predefined dictionary of text concepts~\citep{bhalla2024interpreting}.

\paragraph{Feature Attribution and Saliency Maps.} A key aspect of geographic INR explainability is understanding how the pretraining data---often, georeferenced natural or remote sensing imagery---contributes to the resulting location embedding. Saliency maps identify the regions of an image that most strongly influence a model's prediction. Gradient-weighted Class Activation Mapping (Grad-CAM)~\citep{selvaraju2017grad} does so by computing gradients of the class score and has extensions to embedding models~\citep{chen2020adapting}. \new{As most existing location encoders are trained using CLIP-style losses, we leverage methods specific to CLIP. Grad-CAM on CLIP models has been shown to produce misleading or unsatisfactory results, such as highlighting background elements and noisy activations~\citep{li2022eclip, clipsurgery}.} \edited{ECLIP~\cite{li2022eclip} proposes masked max pooling to avoid this semantic shift, but it requires an additional alignment step, limiting its applicability.} We utilize CLIP Surgery~\citep{clipsurgery}, \edited{which instead modifies the} inference architecture of a CLIP-trained model to address these visualization drawbacks, producing better qualitative results without the need for finetuning.

\renewcommand{\vec}[1]{\mathbf{#1}}

\section{Unsupervised Discovery of Sparse Concepts}
\label{sec: monosemantic emb}

To uncover relevant concepts in location embeddings, our first explainability analysis uses sparse dictionary learning to disentangle these embeddings into human-interpretable features. Building on prior work~\cite{cunningham2023sparse, pach2025sparse}, we use a sparse autoencoder (SAE) to learn a sparse dictionary of concepts that enables accurate reconstruction of location embeddings. The sparsity objective promotes interpretability by ensuring that only a small number of neurons in the autoencoder are activated for accurate reconstruction of location embeddings. Uncovering the semantics of these neurons into human-interpretable concepts can reveal the structure present in the location embeddings.

In our analysis, we use the BatchTopK SAE~\cite{bussmann2024batchtopk} method commonly used for interpreting large language models and vision--language models \cite{pach2025sparse}. This method constrains at most $ b\times k$ elements per batch of size $b$ to be non-sparse. As such, it enables a variable, flexible number of non-sparse elements to be activated per sample in the batch. 
To interpret the semantics of sparse neurons, we evaluate their visual and geographic monosemanticity (MS). The visual MS metric defined by \citet{pach2025sparse} measures whether a neuron in the sparse autoencoder layer activates for images with similar visual features. We compute this by iterating over pairs of locations and for each pair computing the correlation between the shared neuron activation at those locations and the visual similarity between the corresponding images. Higher values indicate that the locations activating a neuron share similar visual features. Further, the geographic monosemanticity (GeoMS) metric introduced in our study assesses whether a sparse concept activates over a local or global geographical region. We compute this analogously to the previous metric by replacing visual similarity with the inverse Haversine distance between two points on Earth. Thus, neurons with higher GeoMS values activate more strongly within a local geographic region.

\subsection{Sparse Autoencoder Experimental Setup}
We use the BatchTopK autoencoder model to interpret SatCLIP, GeoCLIP, and Climplicit location encoders in terms of sparse dictionary concepts. To account for the geographical distribution bias in the location encoders, we train the SAE on two sets of 100K locations. The first set is sampled uniformly at random (UAR) across landmasses, similar to the training distribution for the SatCLIP and Climplicit encoders. The second set focuses on human-visited areas and is formed by sampling locations from the Geo-YFCC dataset \citep{dubey2021adaptive}, which has a distribution similar to that used to train the GeoCLIP model. 
We follow the evaluation protocol outlined in Appendix Table \ref{tab:sae_ms_evaluation_setup} and compute the monosemanticity of SAE models trained on the UAR dataset using satellite imagery from the S2-100k dataset \cite{klemmer2025satclip}, and on human-visited locations using natural imagery from the Geo-YFCC dataset.
For training the BatchTopK autoencoder model, we set the sparse dictionary size to the embedding dimension for each location encoder and use $k=20$ to flexibly constrain the number of non-sparse neurons per sample. We use a learning rate of $3e^{-4}$; for the other hyperparameters, we use the default values provided by \citet{pach2025sparse}.

\subsection{Reconstruction Quality and Concept Monosemanticity}
\label{sec:sae_results}

Table ~\ref{tab:sae_reconstruction} compares the reconstruction quality (MSE, $R^2$ score) and the monosemanticity of the BatchTopK sparse autoencoders for three different location encoders on the datasets used in this study. GeoCLIP embeddings yield the lowest reconstruction scores, achieving only about half and two-thirds of the $R^2$ variance explained for the UAR and human-visited locations datasets, respectively. In contrast, SatCLIP and Climplicit demonstrate superior performance, as the BatchTopK SAE achieves close-to-perfect reconstruction of their embeddings on both datasets.
In addition, the sparse neurons for SatCLIP and Climplicit exhibit stronger visual and geographic monosemanticity than those for GeoCLIP, indicating that these neurons activate imagery with more similar visual patterns and regions that are geographically closer. Finally, \Cref{tab:sae_reconstruction} highlights a trade-off in monosemanticity scores driven by the location distribution in the training dataset and the source of imagery used for evaluation. The models trained in UAR locations and evaluated on satellite imagery from the S2-100k exhibit stronger visual monosemanticity. Conversely, SAE models trained on human-visited sites and evaluated on natural imagery from Geo-YFCC achieve higher geographical monosemanticity, with sparse neurons that show increased specificity to local geographical regions. 

\begin{table}[tbp]
\centering
\caption{\textbf{Reconstruction quality of the location embeddings with the BatchTopK SAE method and monosemanticity of the latent neurons.} The Max columns report the monosemanticity of the neuron with the highest monosemanticity (MS) values in the SAE, while the Mean columns report the average monosemanticity over all neurons. The location embeddings of both SatCLIP and Climplicit allow explaining a higher percentage of the variance after SAE reconstruction, and the visual MS values suggest that their sparse neurons activate on more coherent visual concepts than in GeoCLIP. Further, these concepts pertain to more regional geographies.}
\label{tab:sae_reconstruction}
\small
\setlength{\tabcolsep}{3pt}
\begin{tabular}{llcccccc}
\toprule
& & & & \multicolumn{2}{c}{\textbf{Visual MS} $\mathbf{\uparrow}$} & \multicolumn{2}{c}{\textbf{Geo MS} $\mathbf{\uparrow}$} \\
\cmidrule(lr){5-6} \cmidrule(lr){7-8}
\textbf{Dataset} & \textbf{Model} & \textbf{MSE} $\mathbf{\downarrow}$ & $\mathbf{R^2 \uparrow}$ & \textbf{Mean} & \textbf{Max} & \textbf{Mean} & \textbf{Max} \\
\midrule
\multirow{3}{*}{UAR} & GeoCLIP & $0.00 \pm 0.00$ & $0.52 \pm 0.00$ & $0.03 \pm 0.00$ & $0.16 \pm 0.09$ & $0.03 \pm 0.00$ & $0.12 \pm 0.04$ \\
& SatCLIP & $0.01 \pm 0.00$ & $0.96 \pm 0.00$ & $0.13 \pm 0.00$ & $0.42 \pm 0.01$ & $0.07 \pm 0.00$ & $0.31 \pm 0.06$ \\
& Climplicit & $0.01 \pm 0.00$ & $0.92 \pm 0.00$ & $0.17 \pm 0.00$ & $0.59 \pm 0.02$ & $0.19 \pm 0.00$ & $0.59 \pm 0.04$ \\
\midrule
\multirow{3}{*}{\shortstack{Human-\\Visited}} & GeoCLIP & $0.00 \pm 0.00$ & $0.65 \pm 0.00$ & $0.19 \pm 0.00$ & $0.35 \pm 0.03$ & $0.07 \pm 0.00$ & $0.65 \pm 0.07$ \\
& SatCLIP & $0.00 \pm 0.00$ & $0.99 \pm 0.00$ & $0.20 \pm 0.00$ & $0.33 \pm 0.02$ & $0.18 \pm 0.00$ & $0.87 \pm 0.06$ \\
& Climplicit & $0.01 \pm 0.00$ & $0.99 \pm 0.00$ & $0.21 \pm 0.00$ & $0.41 \pm 0.03$ & $0.22 \pm 0.00$ & $0.92 \pm 0.04$ \\
\bottomrule
\end{tabular}
\end{table}

\subsection{Interpreting Sparse Concepts}
\begin{figure}[tbp]
    \centering
   \includegraphics[width=0.98\linewidth]{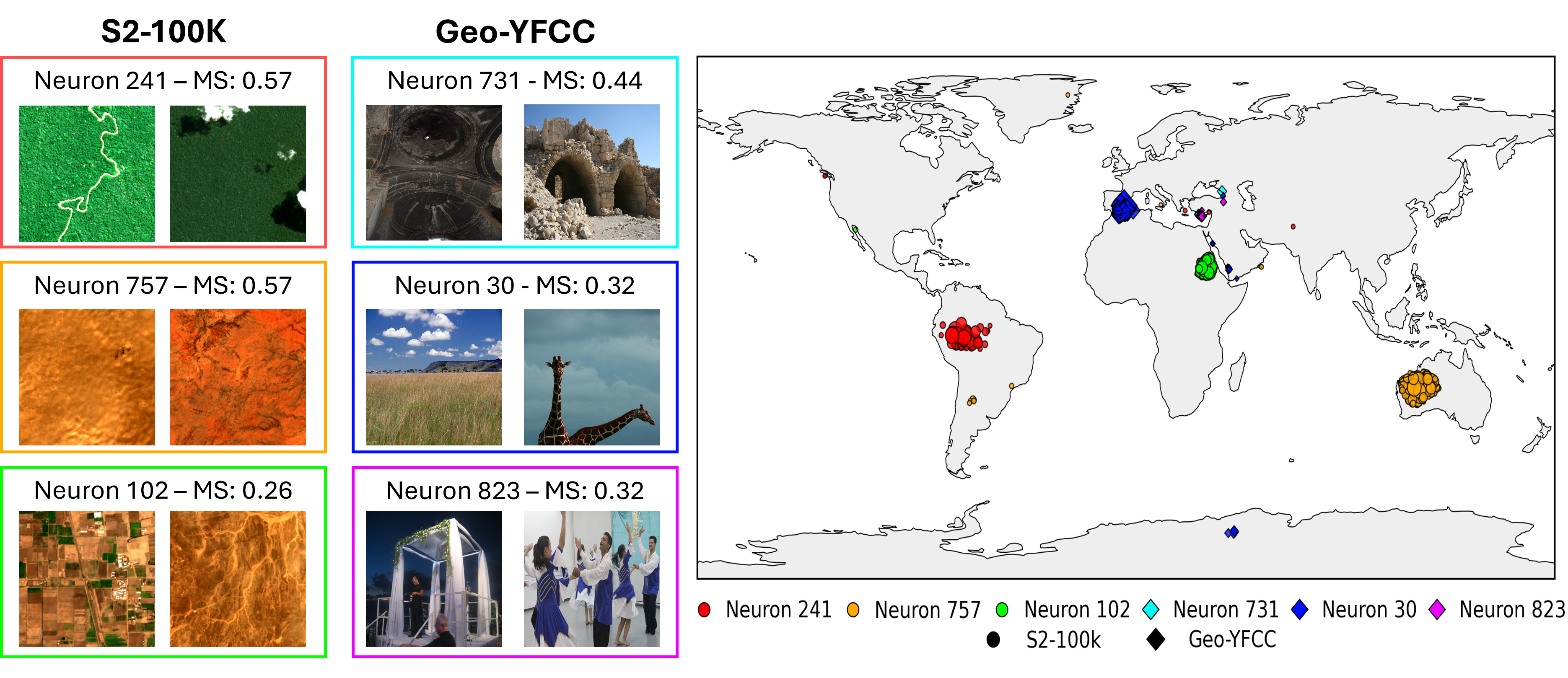}
    \caption{\textbf{Qualitative analysis of selected neurons from the top-10 ranked by visual (top and middle row) and geospatial (bottom row) monosemanticity for SAE reconstruction on the Climplicit encoder.} The left panels display images from the top-10 activating locations from the S2-100k and Geo-YFCC datasets for selected neurons, with the MS values indicating their visual monosemanticity. The corresponding spatial distribution of these neurons is shown on the map on the right, with marker sizes indicating the strength of neurons activations. The visualized neurons typically encode rainforests in the Amazon basin (Neuron 241), deserts in Australia (Neuron 757), and agricultural fields and deserts in the Middle East and the Horn of Africa (Neuron 102) for the S2-100k dataset. Furthermore, the sparse neurons on Geo-YFCC activate to archaeological sites in the Middle East (Neuron 731), coherent visual patterns over a broader area, such as natural landscapes with animals in Europe (Neuron 30), and regional polysemantic visual concepts, such as wedding chuppah and dance scenes in the Middle East (Neuron 823).}
    \label{fig:sae_qualitative}
\end{figure}

To interpret the semantics of neurons in the sparse autoencoder layer, in Figure \ref{fig:sae_qualitative} we visualize the strongest activating samples from the S2-100k and the Geo-YFCC datasets for neurons with the top-10 highest visual and geographic monosemanticity for the reconstruction of the Climplicit embeddings. These samples reveal that the sparse neurons having high monosemanticity scores capture three distinct patterns: 1) \textit{regional monosemantic concepts}, 2) \textit{regional polysemantic concepts}, and 3) \textit{recurring visual patterns over broader area}. The regional monosemantic concepts in Figure \ref{fig:sae_qualitative} are represented by neurons 241 and 757 from the S2-100k dataset, as well as by neuron 731 from the Geo-YFCC dataset. They capture rainforests in the Amazon basin, deserts in Australia, and archaeological sites in the Middle East, respectively. Further, regional polysemantic neurons fire on diverse visual concepts specific to a certain region, such as neuron 102 from the S2-100k dataset, which encodes concepts such as agricultural fields and deserts specific to the Middle East and the Horn of Africa, and neuron 823 from Geo-YFCC, which captures events from the Middle East, such as a wedding chuppah and a dance scene. The third type of neuron encodes recurring visual patterns over a broader area, and is exemplified in \Cref{fig:sae_qualitative} by neuron 30 from the Geo-YFCC dataset that strongly activates in natural areas in Europe, often including animal scenes. Furthermore, we note the existence of sparse neurons with limited interpretability that have lower monosemanticity scores and activate globally across diverse landscapes.   

In addition to improving the interpretability of the location encoders, we found that the sparse autoencoders can also be useful in identifying specific artifacts in the pretraining datasets. In particular, Figure~\ref{fig:sae_visual_artifacts} in the Appendix illustrates that some of the top-10 neurons with the highest visual monosemanticity encode visual artifacts in Sentinel-2 imagery around Greenland from the S2-100k dataset.

\renewcommand{\vec}[1]{\mathbf{#1}}
\newcommand{\mat}[1]{\mathbf{#1}}

\newcommand{\enc}{f_{\mathrm{enc}}}
\newcommand{\dec}{f_{\mathrm{dec}}}

\newcommand{\s}{\sigma}

\section{Decomposition into Predefined Natural Language Concepts}
\label{sec: text concepts}

The next method we use to explain location embeddings also utilizes sparsity to identify embedding structure, but introduces additional inductive bias through a prespecified natural language concept dictionary, following the SpLiCE framework proposed by \citet{bhalla2024interpreting}. In this setting, the set of one or two-word dictionary concepts are constructed a priori through domain knowledge. 

\paragraph{Aligning Location Embeddings and Text Embeddings.} SpLiCE was originally developed for interpreting CLIP image embeddings, which are naturally aligned with text embeddings. Extending this approach to location embeddings requires a shared semantic space of \textit{location} and text embeddings. Some geographic INRs have this prerequisite by design, such as GeoCLIP~\cite{vivanco2023geoclip}, which uses a CLIP location encoder during pretraining. To apply SpLiCE to other location embeddings, we construct this aligned space manually. We achieve this alignment by treating this as a locked-image tuning problem~\citep{zhai2022lit} with a fixed location encoder and a learned text encoder.

As GeoCLIP is trained contrastively with the OpenCLIP ViT-L~\citep{Radford2021LearningTV} image encoder, we can directly replace the image encoder with the OpenCLIP ViT-L text encoder to produce text embeddings in a space aligned with the GeoCLIP location encoder. To develop a shared semantic space of location and text embeddings for all other location encoders, we finetune the OpenCLIP ViT-L model using the Git-10M dataset~\citep{Text2Earth}, which contains 10.5M geolocated remote-sensing images paired with captions generated via the GPT-4o API~\citep{achiam2023gpt}. From Git-10M, we additionally construct a vocabulary of 2000 geospatial concepts by selecting the most frequent one-word nouns appearing in the captions. We then apply SpLiCE~\citep{bhalla2024interpreting} using this fixed concept dictionary. In practice, CLIP's image and text modalities are not fully aligned~\cite{liang2022mind}, which we find applies to the location and text alignment as well. Therefore, we mean-center and re-normalize the location and text embeddings as in \citet{bhalla2024interpreting}. Again following the methodology of \citet{bhalla2024interpreting}, we choose the $\ell_1$ regularization ($\lambda$) to encourage sparsity, aiming for solutions with approximately 5--20 active concepts per example.

\subsection{Reconstructing Location Embeddings from Sparse Decompositions}

As with the sparse autoencoder explanations, we first measure how well the SpLiCE decomposition preserves the geometry of the original embedding space (\Cref{tab:splice_quant}). Unlike reconstruction-based approaches, here the goal is not to perfectly recover $\vec{z}$, but to assess whether a sparse and interpretable concept representation retains directional structure in embedding space.
The results in \Cref{tab:splice_quant} show that SpLiCE achieves reasonable directional reconstruction
(cosine similarity substantially above 0) and overall reconstruction (low MSE) while maintaining a substantially lower number of active concepts compared to the original location embeddings. This indicates that the concept dictionary can be used to capture meaningful structure in the embedding space under sparsity constraints.

\subsection{Qualitative Concept Identification and Comparisons}

\Cref{fig:splice_decompositions} provides a qualitative analysis of SpLiCE decompositions by inspecting the top-4 text concepts associated with four regions representing distinct land cover types.
This qualitative evaluation illustrates how different location encoders distribute semantic information over the geospatial concept set.
\edited{For GeoCLIP embeddings, concept decompositions tend to be more fine-grained and specific in highly photographed locations such as Paris. In contrast, less frequently photographed regions like Siberia demonstrate some interpretability in the top concept ``tundra'', yet other terms represent general, non-geographic concepts. On the other hand, SatCLIP embeddings seem to capture more general trends related to the landcover of the region, as seen in the concepts relevant for Paris. Climplicit shows mixed behavior: the SpLiCE decompositions are partially meaningful for regions such as the Sahara, where coarse land cover signals may be prevalent, but yield less coherent decompositions for Paris and Siberia. Lastly, CSP-fMoW produces concept decompositions that are generally less interpretable across the different regions, with top concepts appearing less consistently aligned with expected geographic attributes. Together, these results suggest that the interpretability of these SpLiCE decompositions varies significantly  across the different location embeddings.}

\begin{table}[tbp]
\centering

\caption{\textbf{Sparse concept decompositions preserve structure in location embedding space.}
We compare original location embeddings $\vec{z}$ with their sparse representations produced by SpLiCE.
Results are obtained from SpLiCE decompositions of the same two sets of points used to evaluate SAEs: \textbf{UAR} and \textbf{Human-Visited}. All values are averaged over all points in the dataset and reported as mean $\pm$ standard deviation.
The sparsity regularization parameter is set to $\lambda=0.25$ for GeoCLIP and SatCLIP, and $\lambda=0.175$ for Climplicit and CSP-fMoW.}

{\setlength{\tabcolsep}{4pt}
\resizebox{\linewidth}{!}{%
\begin{tabular}{l ccc ccc}
\toprule
 & \multicolumn{3}{c}{\textbf{UAR}} & \multicolumn{3}{c}{\textbf{Human-visited}} \\
\cmidrule(lr){2-4} \cmidrule(lr){5-7}
\textbf{Model} &
\textbf{MSE} $\downarrow$ &
\textbf{Cos. Sim.} $\uparrow$ &
\textbf{\# Concepts} &
\textbf{MSE} $\downarrow$ &
\textbf{Cos. Sim.} $\uparrow$ &
\textbf{\# Concepts} \\
\midrule
GeoCLIP & 0.002 $\pm$ 0.000 & 0.435 $\pm$ 0.064 & 8.2 $\pm$ 2.9 & 0.002 $\pm$ 0.000 & 0.377 $\pm$ 0.061 & 7.8 $\pm$ 3.1 \\
SatCLIP & 0.005 $\pm$ 0.001 & 0.360 $\pm$ 0.078 & 9.1 $\pm$ 3.5 & 0.004 $\pm$ 0.001 & 0.462 $\pm$ 0.088 & 8.7 $\pm$ 3.4 \\
Climplicit & 0.001 $\pm$ 0.000 & 0.299 $\pm$ 0.091 & 9.1 $\pm$ 4.3 & 0.001 $\pm$ 0.000 & 0.361 $\pm$ 0.092 & 11.3 $\pm$ 4.8 \\
CSP-fMoW & 0.004 $\pm$ 0.001 & 0.442 $\pm$ 0.192 & 15.4 $\pm$ 5.4 & 0.005 $\pm$ 0.001 & 0.331 $\pm$ 0.127 & 13.8 $\pm$ 6.7 \\
\bottomrule
\end{tabular}
}}

\label{tab:splice_quant}
\end{table}

\begin{figure}[t]
    \new{\centering
    \setlength{\tabcolsep}{2pt}
    \renewcommand{\arraystretch}{0.95}
    \newcommand{\spliceimgbig}[1]{\raisebox{-0.5\height}{\includegraphics[width=0.25\linewidth]{#1}}}
    \newcommand{\splicetag}[1]{%
        \refstepcounter{subfigure}\label{fig:splice_#1}%
        \small(\alph{subfigure})%
    }
    \setcounter{subfigure}{0}
    \begin{tabular}{@{}l@{\hspace{4pt}}c@{\hspace{-3mm}}cc@{\hspace{-3mm}}c@{}}
        & \, \, \, \, \, \small Paris & \, \, \, \, \, \small Sahara & \, \, \, \small Siberia & \, \, \, \small Bali \\
        \raisebox{-0.5cm}{\rotatebox{90}{\small \textbf{GeoCLIP}}} &
            \spliceimgbig{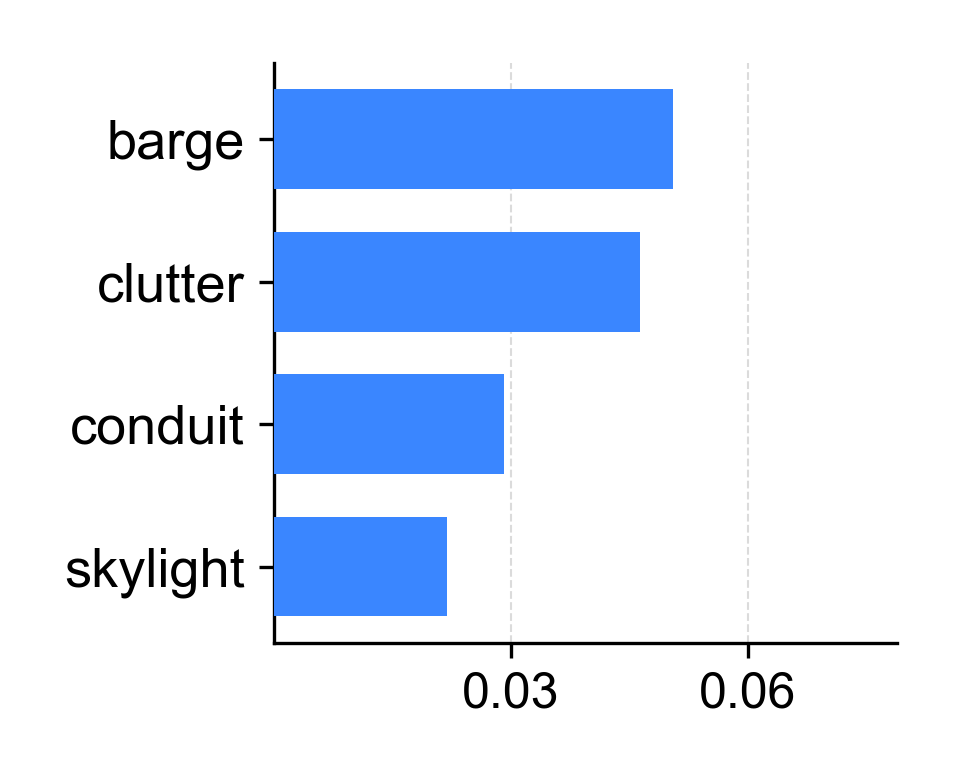} &
            \spliceimgbig{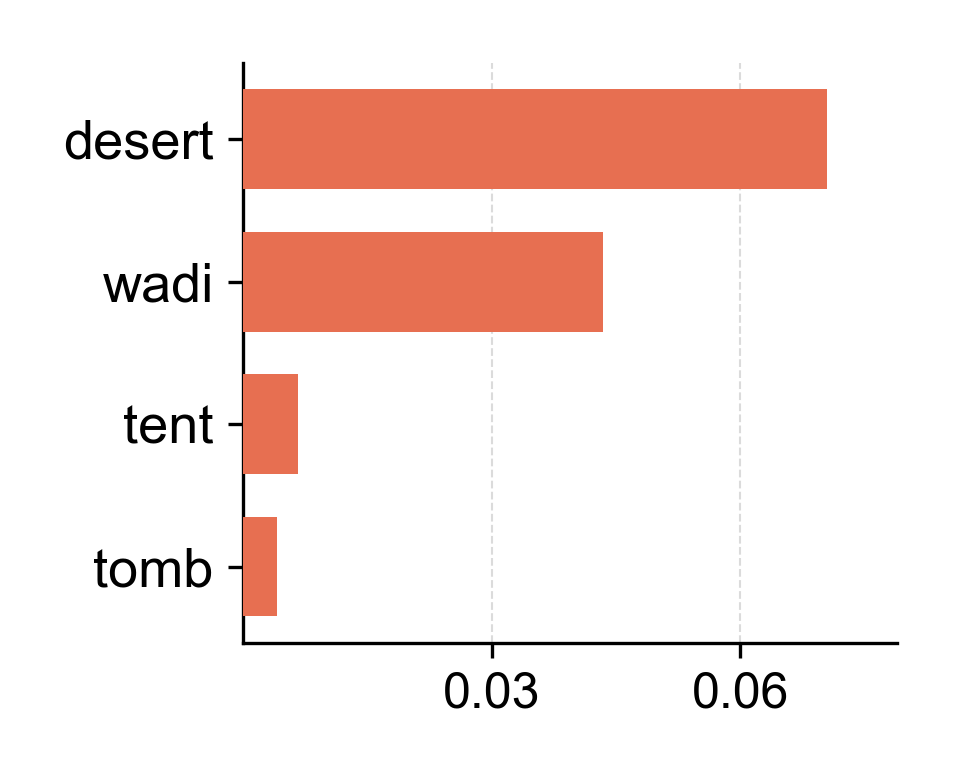} &
            \spliceimgbig{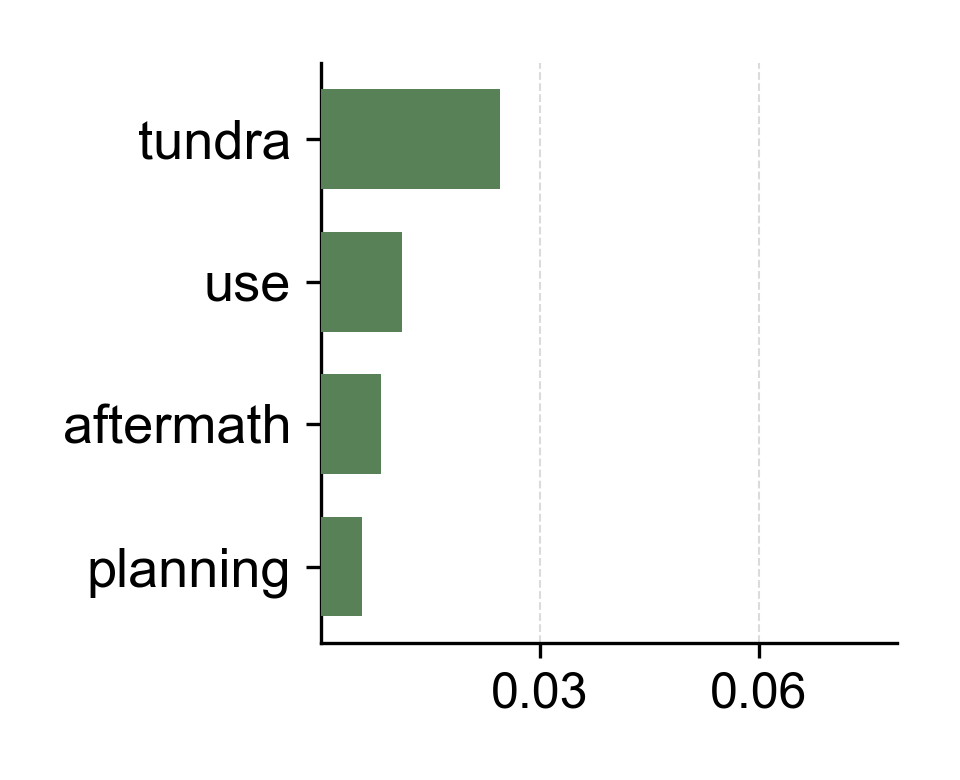} &
            \spliceimgbig{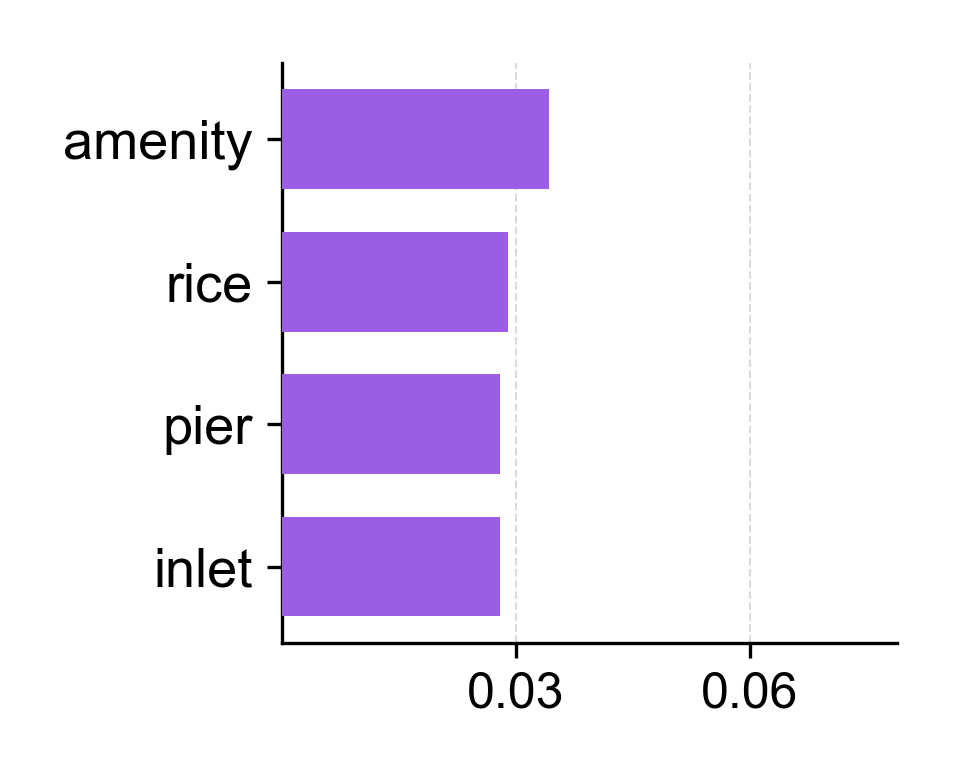} \\
        \raisebox{-0.5cm}{\rotatebox{90}{\small \textbf{SatCLIP}}} &
            \spliceimgbig{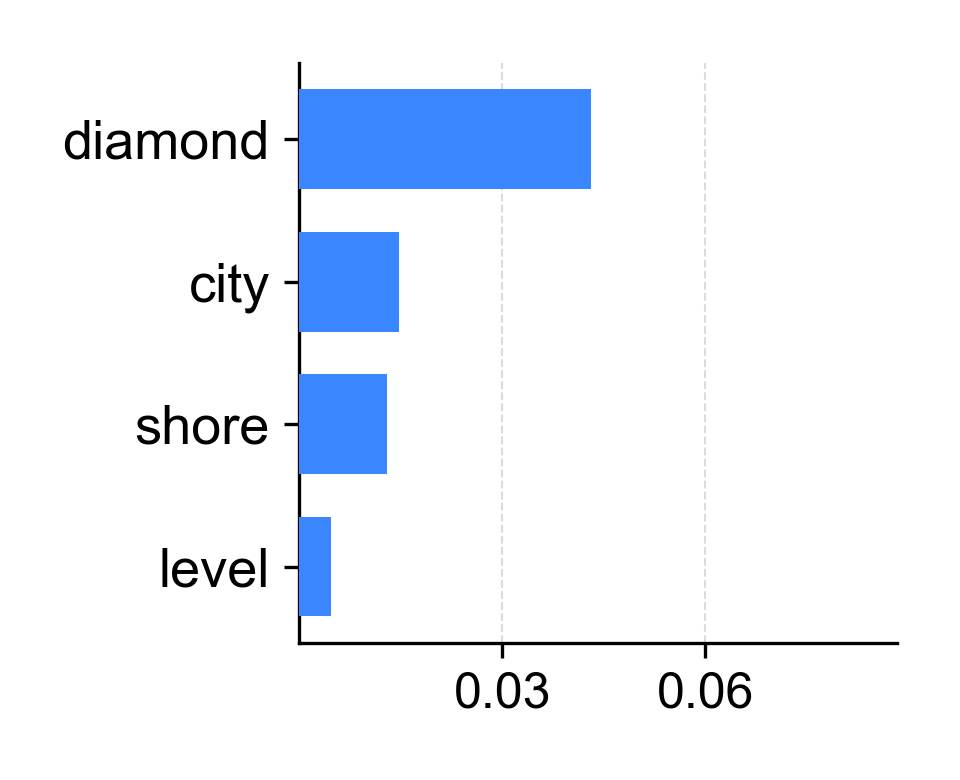} &
            \spliceimgbig{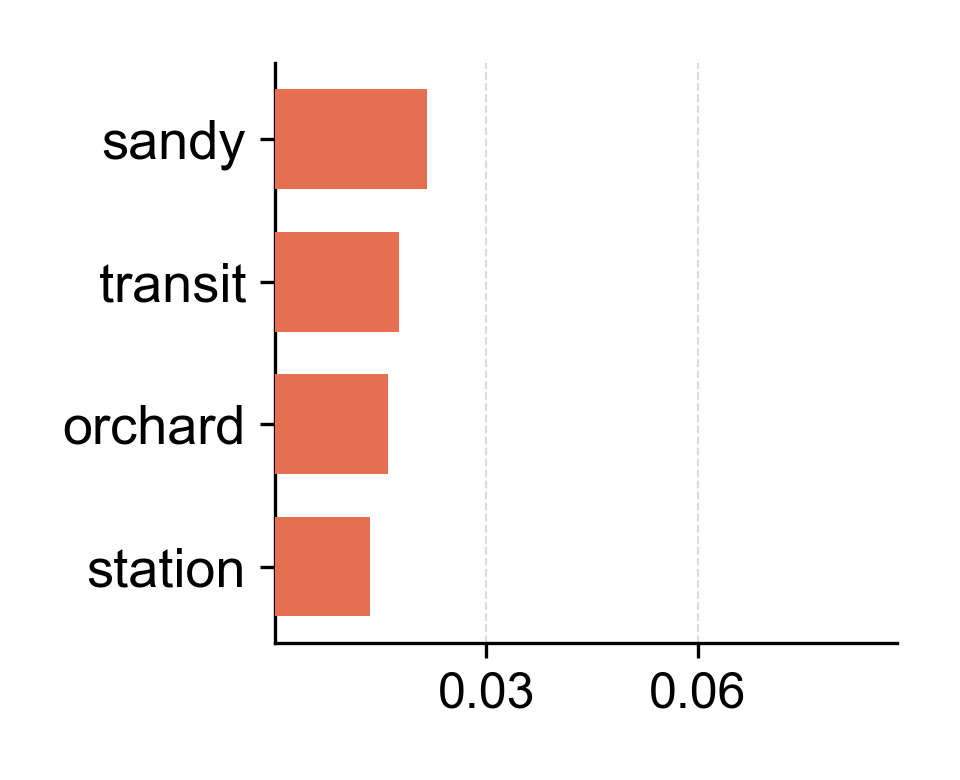} &
            \spliceimgbig{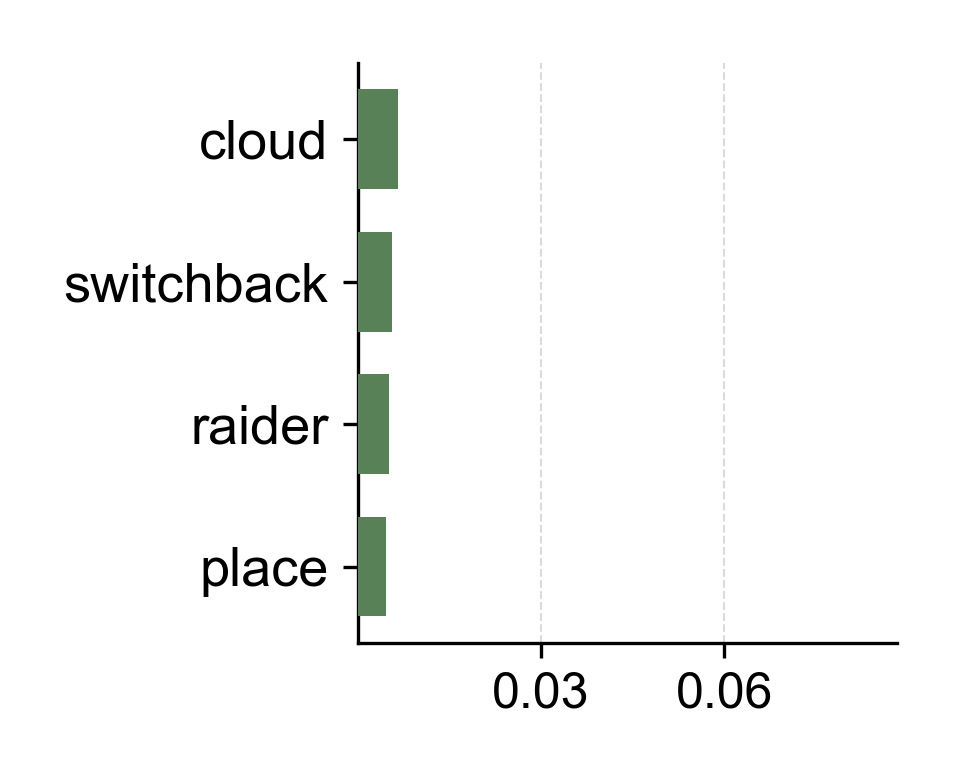} &
            \spliceimgbig{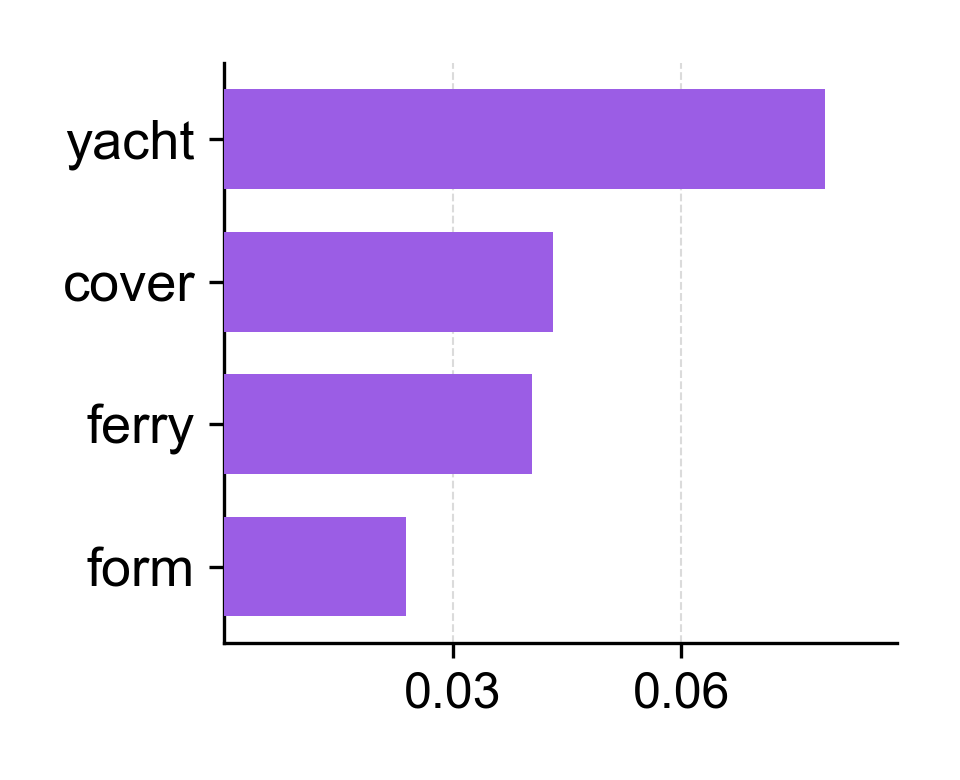} \\
    \end{tabular}}
    \caption{\textbf{SpLiCE decompositions highlight different signals captured by location embeddings.} Bar plots show the concept weights of the top 4 concepts in SpLiCE decompositions for GeoCLIP~(top) and SatCLIP (bottom) across four locations representing different land cover types: Paris, Sahara, Siberia, and Bali. \new{SpLiCE decomposition weights are averaged over a $0.5^\circ$ grid within each region's boundary.
    Results for the other location encoders are included in Appendix~\Cref{apx: sec: splice}.}}
    \label{fig:splice_decompositions}
\end{figure}

\section{Feature Attribution to Extract Visual Features}
\label{sec: image space}

For our third explainability analysis, we deviate from methods that explain embeddings ``off-the-shelf'' and now turn to feature attribution in input space to validate if the information content of location embeddings corresponds to meaningful visual evidence. We leverage CLIP Surgery \citep{clipsurgery} to generate saliency maps for contrastively pretrained location encoders. CLIP surgery  has been shown to produce superior visualizations compared to traditional backpropagation-based methods such as Grad-CAM (Section~\ref{sec:related_work}).
In this approach: (1) all modifications are applied at inference time, requiring access to the pretrained location encoder weights but no fine-tuning; and (2) the original attention computation is preserved, keeping the class token output unchanged and thus leaving general model usage unaffected.

To adapt CLIP Surgery for location encoders, we use location as the unique class and compute the similarity between the location and image embeddings. CLIP Surgery utilizes an empty string to approximate noise \edited{but} location encoders have no analogue to the null class or empty string used in language models \edited{since every location is valid. As a heuristic, we choose the north pole (lat = 90, lon = 0) to approximate a ``null'' embedding, because it lies in a sparsely (or not at all) sampled region for most location encoder training datasets.}

\subsection{Experimental Setup}

\begin{figure}[tbh]
    \centering
    \setlength{\tabcolsep}{2pt}
    \begin{tabular}{@{}cccc@{}}
        \includegraphics[width=0.22\linewidth, height=0.10\textheight, keepaspectratio=false]{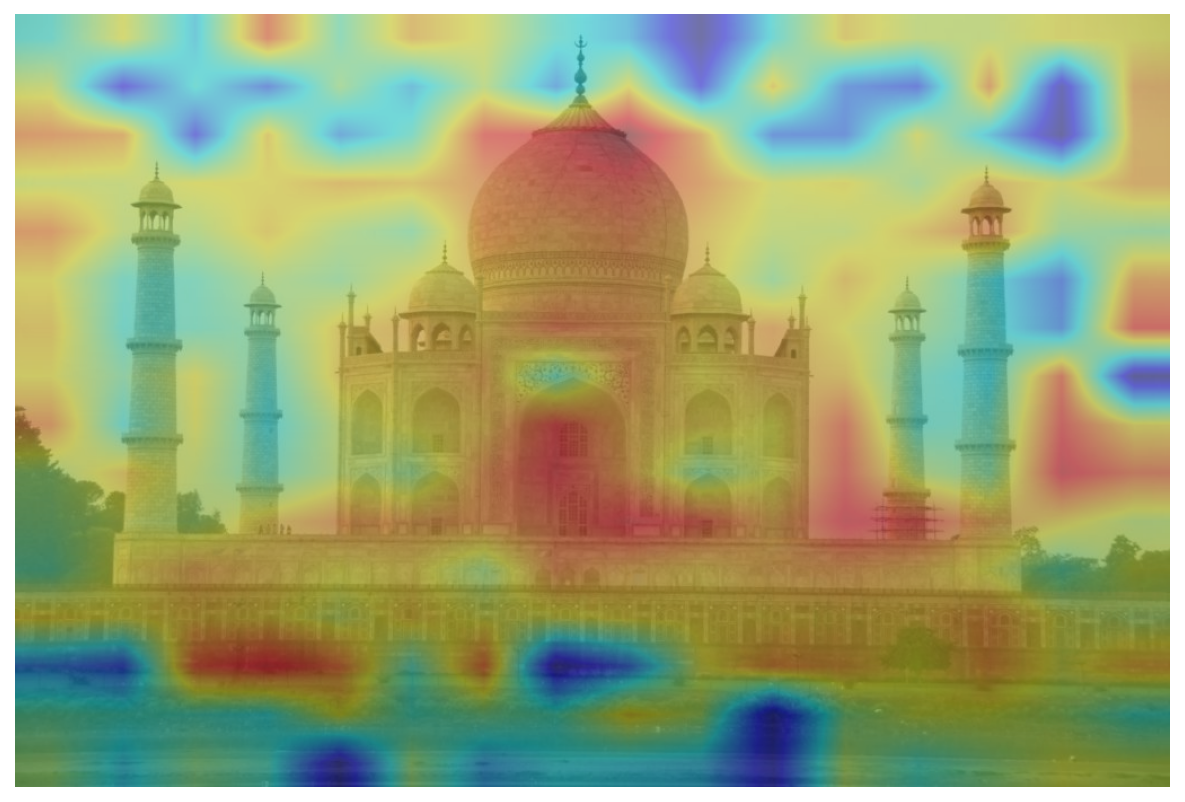} &
        \includegraphics[width=0.22\linewidth, height=0.10\textheight, keepaspectratio=false]{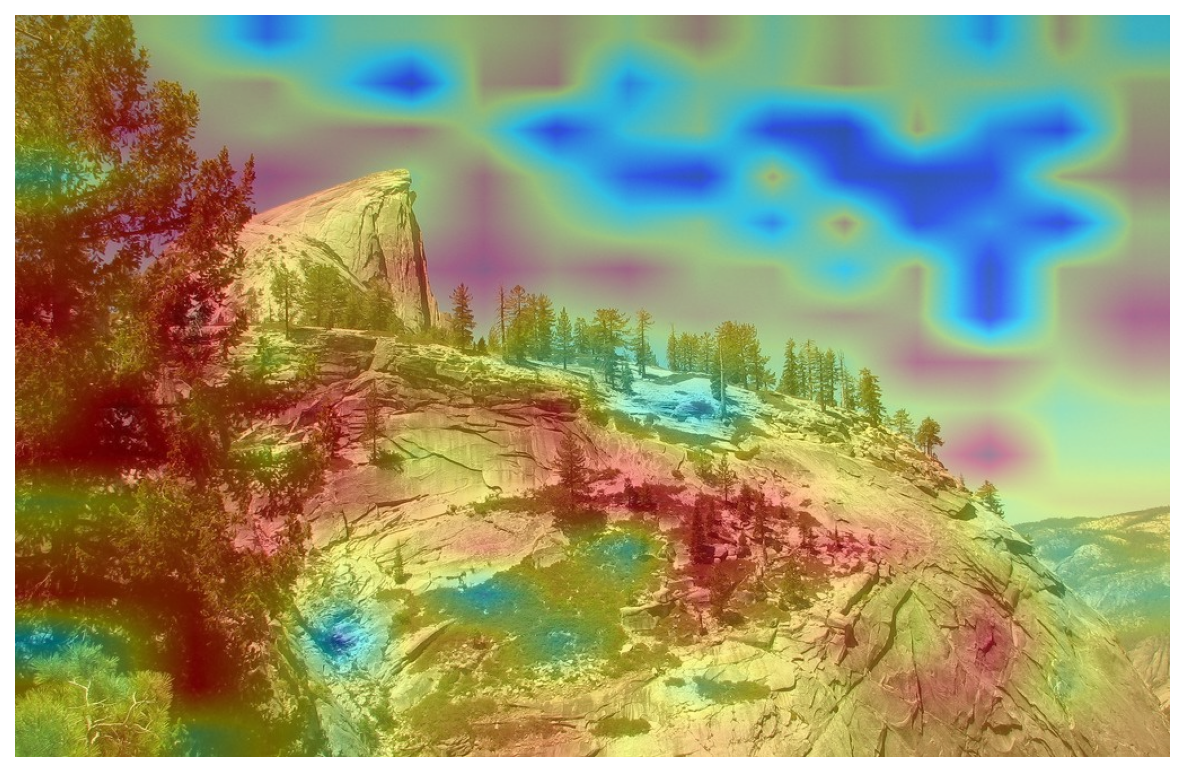} &
        \includegraphics[width=0.22\linewidth, height=0.10\textheight, keepaspectratio=false]{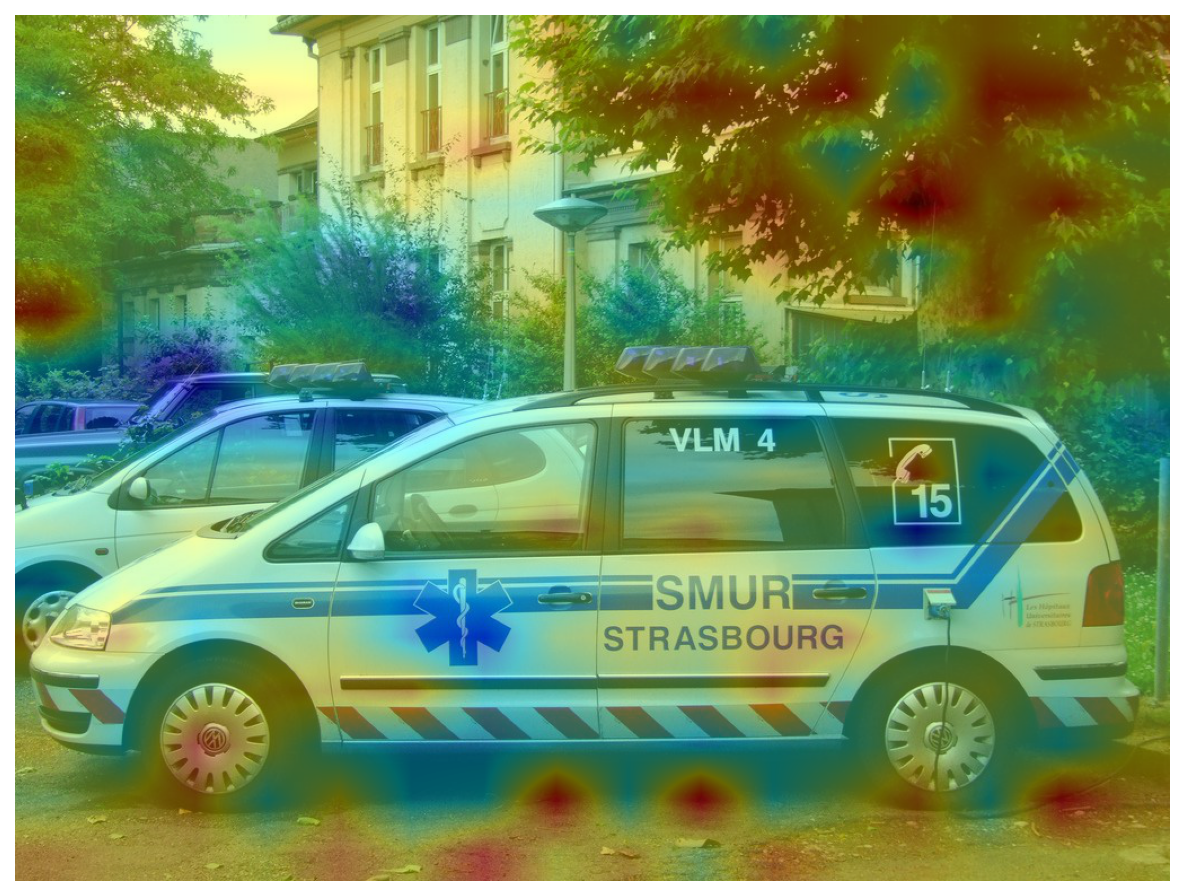} &
        \includegraphics[width=0.22\linewidth, height=0.10\textheight, keepaspectratio=false]{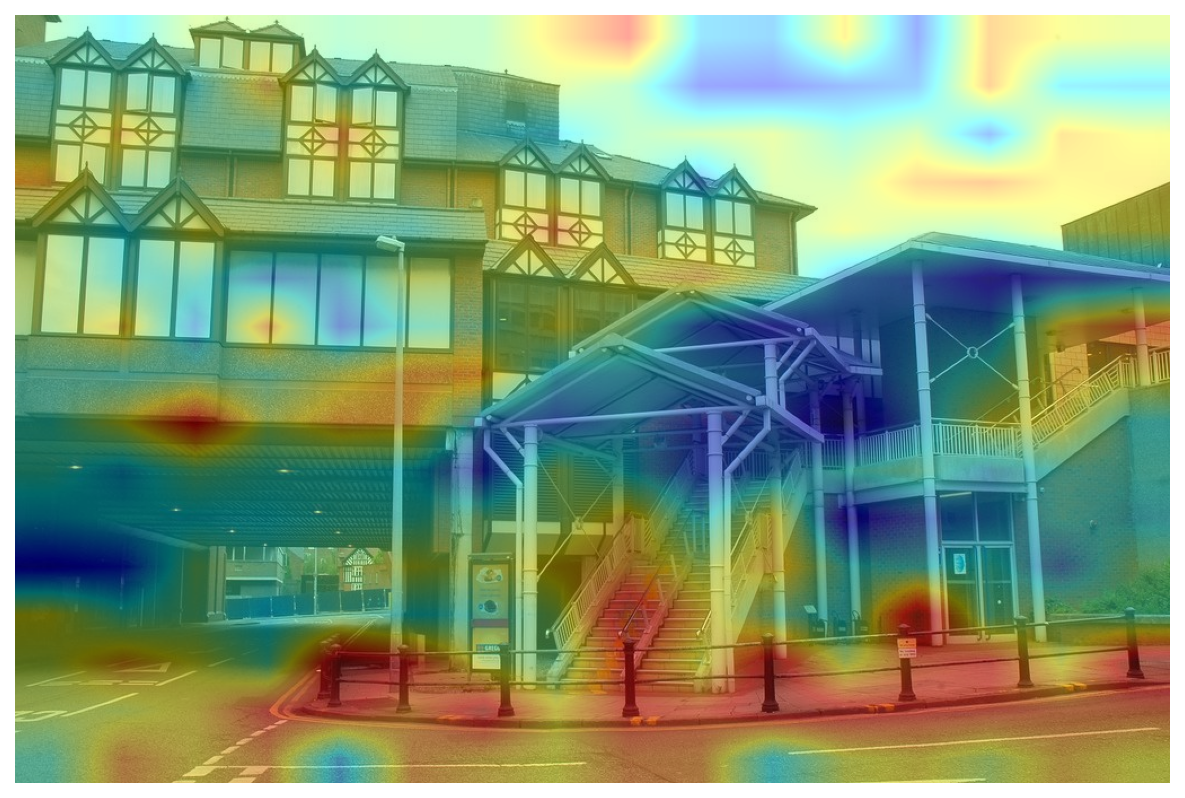} \\[2pt]
        \includegraphics[width=0.22\linewidth, height=0.10\textheight, keepaspectratio=false]{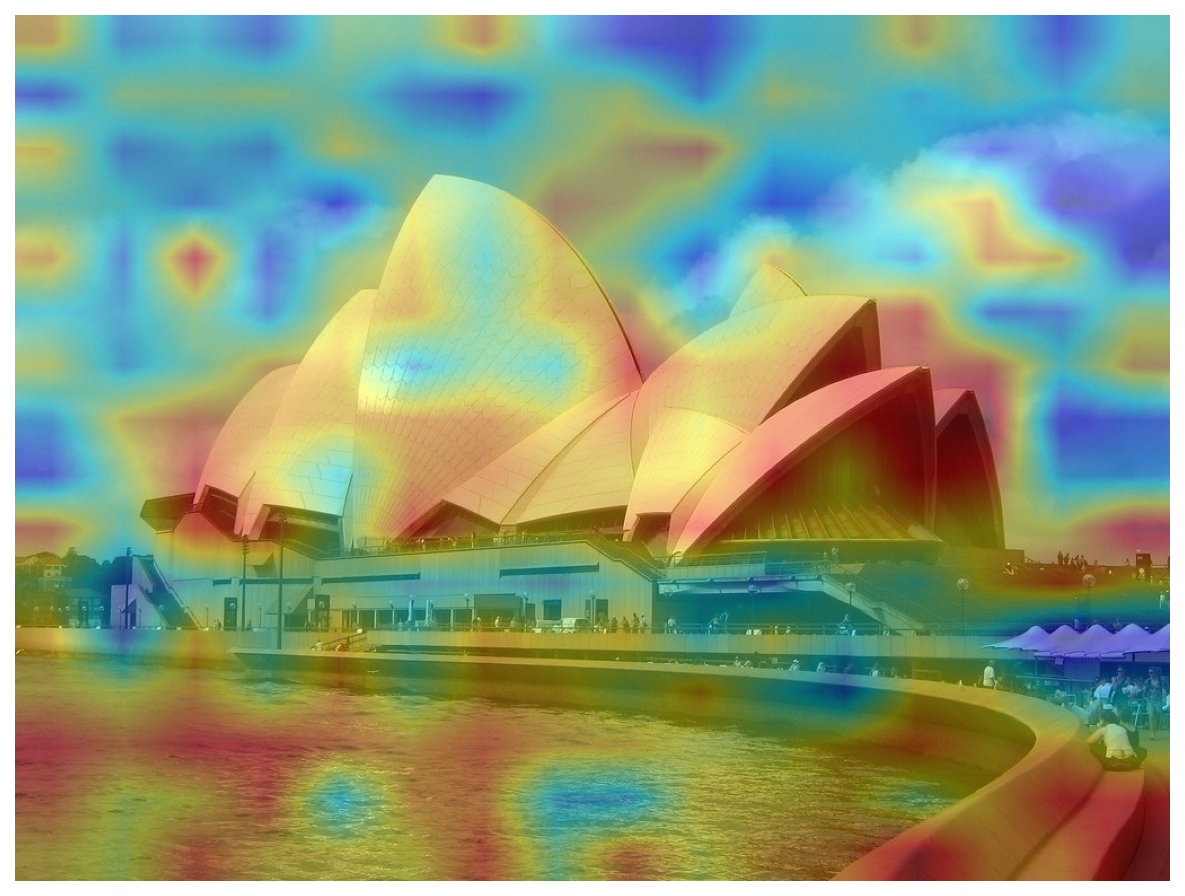} &
        \includegraphics[width=0.22\linewidth, height=0.10\textheight, keepaspectratio=false]{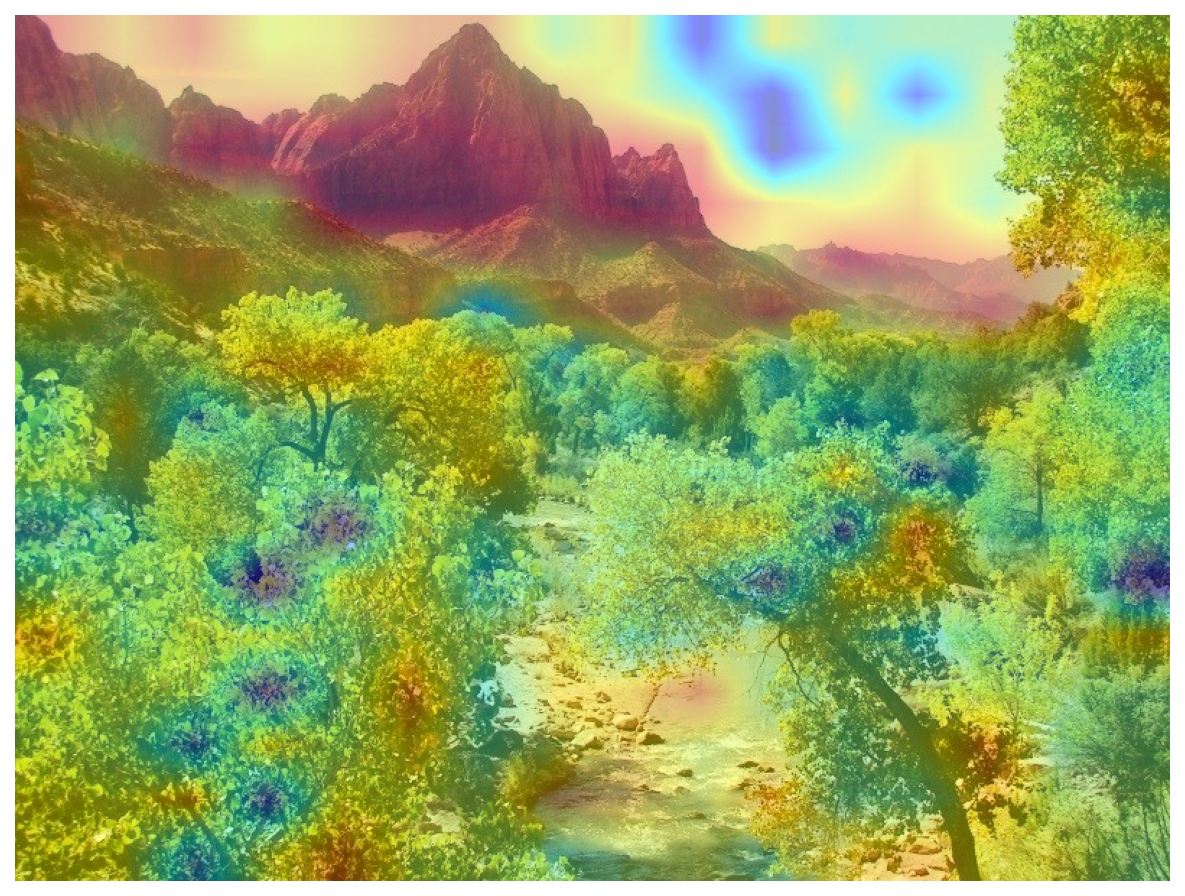} &
        \includegraphics[width=0.22\linewidth, height=0.10\textheight, keepaspectratio=false]{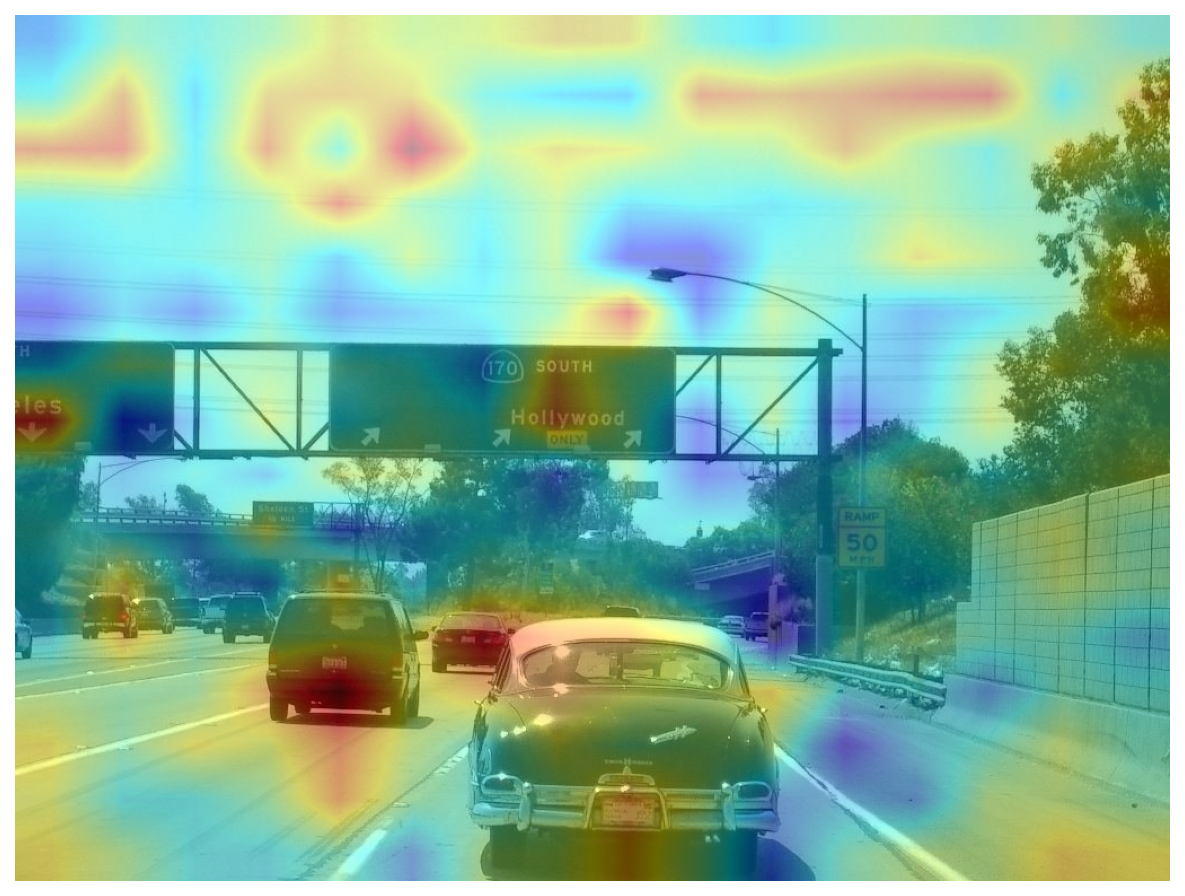} &
        \includegraphics[width=0.22\linewidth, height=0.10\textheight, keepaspectratio=false]{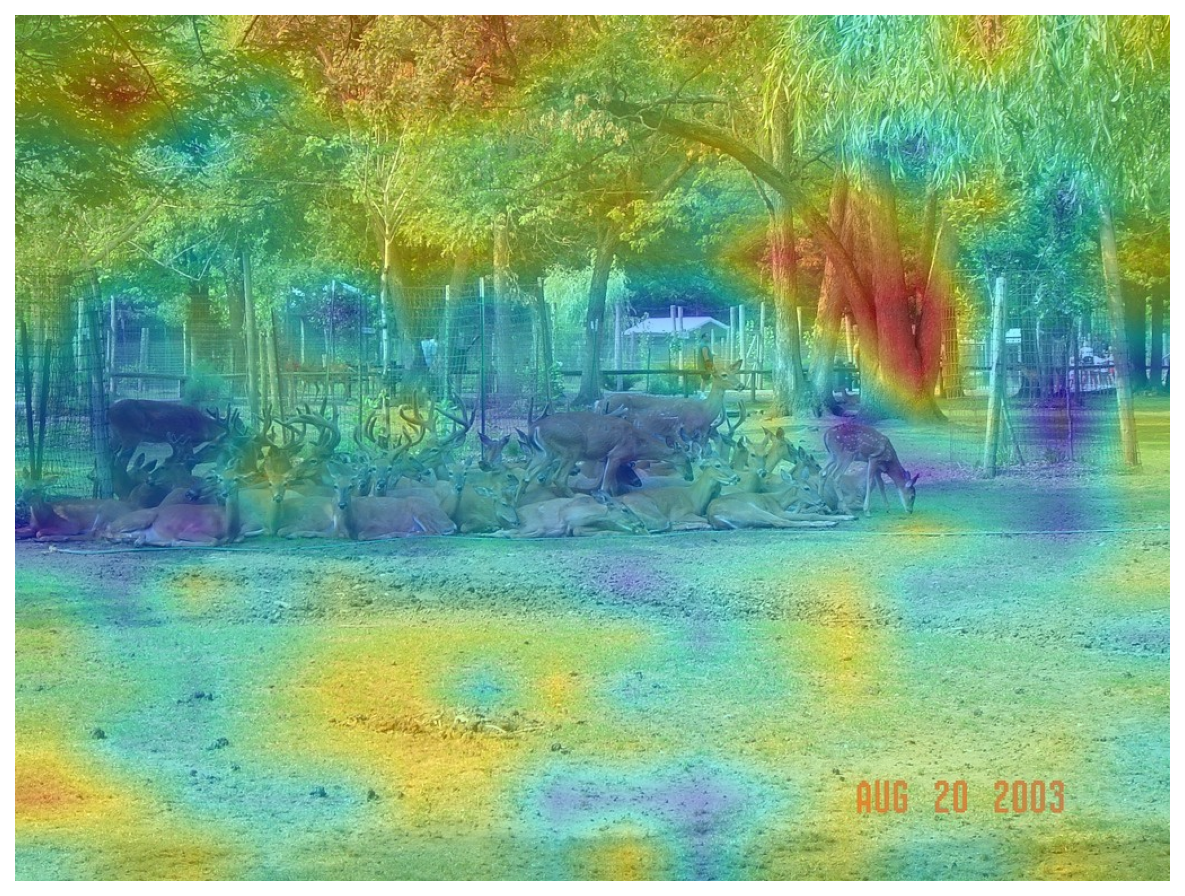} \\[-8pt]
        \multicolumn{1}{c}{\subcaptionbox{Landmarks}[0.22\linewidth]} &
        \multicolumn{1}{c}{\subcaptionbox{Landscapes}[0.22\linewidth]} &
        \multicolumn{1}{c}{\subcaptionbox{Text}[0.22\linewidth]} &
        \multicolumn{1}{c}{\subcaptionbox{Other}[0.22\linewidth]}
    \end{tabular}
    \caption{\textbf{Natural image saliency maps generated using GeoCLIP and CLIP Surgery}, grouped into four categories based on the visual cues the model attends to for geolocation prediction, which include \edited{a) landmarks: recognizable architectural or cultural structures, b) landscapes: natural features such as terrain, vegetation patterns, and sky, c) text: textual cues embedded in the scene, such as signage or place names, and d) other: including street furniture, vegetation, etc.}}
    \label{fig:geoclip_saliency}
\end{figure}

CLIP Surgery requires access to both model weights and source code; therefore we restrict our visualization results to GeoCLIP and SatCLIP, both of which are released under permissive open-source MIT licenses. For GeoCLIP, we applied our method to Im2GPS~\citep{shraman2022translocatordata}, which contains natural images from around the world. We manually grouped the results into four categories---\textit{Landmarks, Landscapes, Text, and Other}---based on the visual cues qualitatively observed to drive the model's attention as shown in Figure~\ref{fig:geoclip_saliency}. 

For SatCLIP, we based our observations on satellite imagery from urban areas\new{, which tend to have greater diversity of landscape elements compared to rural areas}. We selected four cities---\textit{Paris, Amsterdam, New York, and St.\ Louis}---to investigate cross-location patterns, specifically whether European cities exhibit distinct structural elements compared to their American counterparts, and whether the model captures any meaningful similarities or differences between them.
The method for computing SatCLIP saliency maps was identical to that used for GeoCLIP, except that we considered all Sentinel-2 multispectral bands rather than only RGB channels. 

We performed an additional analysis step for the satellite imagery visualizations, since direct maps on satellite imagery are more difficult to interpret than those on natural imagery. We computed saliency maps for 50 images within a city's radius to capture diverse land cover types (urban, suburban, rural, etc.), and for each image we cropped the regions the model attended to. To obtain these crops, we applied a binary mask to the image using the saliency map output scores, then normalized the mask so that each detected region was of roughly equal size. A bounding box was drawn around each region and cropped accordingly. These crops were then automatically grouped by computing their embeddings and applying dimensionality reduction and clustering via t-SNE~\cite{vandermaaten2008tsne} and DBSCAN~\cite{martin1996dbscan} respectively. For the embeddings, we used a ResNet-50~\cite{he2015resnet} pretrained on ImageNet~\cite{deng2009imagenet}, rather than the SatCLIP image encoder, in order to capture purely RGB visual similarities without any bias from satellite-specific training. Figure~\ref{fig:pipeline} visualizes our pipeline to extract and group these visual elements and Figure~\ref{fig:saliency_satclip} shows examples of images taken from each saliency cluster. \new{Additional saliency results for \edited{other} land cover types, including deserts and rainforests, are provided in Appendix~\ref{apx: sec: clipsurgery}.}

\begin{figure}[tbp]
    \centering
    \input{sec/image_space_figures/bboxes_pipeline}
    \captionof{figure}{\new{\textbf{Overview of the visual element extraction pipeline for satellite location embeddings} Starting from a raw image, we compute CLIP-Surgery saliency maps, threshold them into binary masks, normalize and split regions by area, fit bounding boxes around salient regions, and finally cluster all boxes across images within a fixed spatial radius.}}
    \label{fig:pipeline}
    \vspace{0.3cm}
    
    \input{sec/image_space_figures/bboxes_fig}
    \captionof{figure}{\new{\textbf{Extracted visual elements from satellite saliency maps for four cities: Paris, Amsterdam, New York City and St.\ Louis.} Although clusters are derived independently per city, similar clusters were manually aligned across cities for comparison. Original saliency map examples for each city can be found at Appendix Section \ref{apx: sec: clipsurgery}.}}
    \label{fig:saliency_satclip}
\end{figure}

\subsection{Comparing Feature Attributions in Natural and Satellite imagery}

Our initial impressions are that saliency maps on location encoders trained on natural images, such as GeoCLIP, are easier to interpret directly. The location encoder models tend to attend to specific architectural elements in landmarks that are highly distinctive to a particular location, when available. Similarly, for landscapes, the model focuses on a broad range of natural features such as trees, rocks, and sky. Particularly interesting are cases where the model attends to more nuanced image features, such as distinctive lamp posts or tree species,which may serve as implicit geolocation cues (direct image encoders have been shown to pick up on such small differences, like building facades~\citep{doersch2015}). Also noteworthy are cases where textual cues  such as street signs are present.

This motivated our additional analysis step to visualize trends in the saliency maps for satellite imagery. In this modality, the model's attention tends to concentrate on structural boundaries: road patterns (such as roundabouts), land--water edges (rivers, ports), and large urban structures such as bridges and stadiums. Patterns are broadly consistent across European and American cities, yet distinctive elements emerge in each---such as roundabouts prevalent in European cities, and suburban housing clusters and large industrial or storage facilities more typical of American cities.

\section{Discussion}
\label{sec: discussion}

We focus on geographic INRs as an especially challenging case for explainability since, unlike direct embeddings of Earth observation data, they learn continuous representations of multi-scale spatial patterns without directly corresponding to a single input image. We analyze the location embeddings resulting from these geographic INRs through three complementary decomposition strategies: 1) latent concepts, 2) natural language concepts, and 3) visual features. These explainability methods can help to demystify the inner workings of location encoders and support a greater understanding and fidelity in different location embeddings. This transparency can guide users in selecting which location encoder is best suited for a given use case. For model developers, the combination of explainability methods enables debugging and analyzing model performance. For example, SpLiCE can be used to identify concept gaps in the location embeddings, while CLIP Surgery can be used to identify if the model is relying on irrelevant visual features or to identify low-quality data samples and improve the dataset. From a debugging perspective, in Appendix Figure \ref{fig:sae_visual_artifacts}, we illustrate that the sparse encoder analysis can be helpful to identify visual artifacts in the dataset.

\new{The three decomposition methods yield different, but largely consistent interpretations of location embeddings, evidenced by monosemantic neurons representing rainforests and deserts, polysemantic neurons specific to a local area, geospatial text concepts capturing city-level and regional climatic or terrain semantics, and landmarks or distinct visual characteristics of natural and satellite imagery in saliency maps.}
\new{The explainability methods also highlight differences between existing location encoders. Structure in latent and natural language concepts reflect the influence of pretraining data:} \edited{some explanations of location encoder structure highlight fine-grained features such as urban information, while others capture signals of broader geographic attributes like climate zone or biome.} \new{Consistent with these findings,} saliency maps \edited{emphasize key scene elements in pretraining data space---street furniture, trees, and signage---}in natural imagery, while for satellite imagery, they tend to concentrate on structural boundaries such as \edited{roads, urban structures and water}.

\new{Interestingly, the reconstruction results using sparse latent concepts and natural language concepts demonstrate that these decompositions retain significant information content of the original embeddings. For example,} we find that decomposition into latent concepts leads to sparse decompositions that can reconstruct the original location embedding (yielding an $R^2$ score $>0.9$ for location embeddings from SatCLIP and Climplicit). \new{This is consistent with past work that showed the intrinsic dimension of locations embeddings is generally much lower than their ambient dimension \citep{rao2025measuring}, while adding a constructive approach for how one might generate sparse, interpretable concepts either during pretraining or as a post-training step.}

\paragraph{Limitations \& Future Work.} While an important step toward explaining black-box location encoders, our study has several limitations worth noting. Across all three families of methods, interpretability requires some degree of manual effort---for instance, the manual inspection of neuron activations or saliency maps, the steps of aligning the embeddings space of location and text embeddings, or clustering hotspots in satellite image saliency maps. \new{Some of these efforts could be avoided with an interpretable-by-design framework, e.g., a pretrained text-location embedding space would enable direct decomposition into natural concepts via SpLiCE.}
Thus, while we find that level of interpretability varies noticeably across location encoder models for particular methods, this may not be a fundamental limitation of the methodology, so much as the need to very carefully tailor each type of explainability analysis for location encoders. \new{These analyses highlight critical areas for future work, such as using the interpretation of location embeddings for downstream tasks in the presence of polysemantic concepts.} Here, we opted for a broad analysis to understand what reasonable modifications of existing explainability techniques provide when applied to location encoders. We hope our analyses lay a foundation for deeper tailoring of any one of these methods \new{more specifically to geospatial models} (or development of new methods to address current gaps) in future work. 

\section*{Acknowledgements}

This material is based upon work supported by the NSF Graduate Research Fellowship under Grant No. DGE 2040434. We would like to acknowledge use of Jetstream2 at Indiana University through allocation CIS240692 from the Advanced Cyberinfrastructure Coordination Ecosystem: Services \& Support (ACCESS) program, which is supported by National Science Foundation grants \#2138259, \#2138286, \#2138307, \#2137603, and \#2138296. Further, the work of Ivica Obadic is supported by the ML4Earth project of the German Federal Ministry for Economic Affairs and Energy under grant number 50EE2201C.

We also note the use of LLMS to help make our text more concise and generate figure layouts in \verb|tikz|, such as \Cref{fig:method_overview}, and table layouts. However, all results are manually input and verified by the authors.

\bibliographystyle{plainnat}
\bibliography{main}

@inproceedings{klemmer2025satclip,
  title={{SatCLIP}: Global, general-purpose location embeddings with satellite imagery},
  author={Klemmer, Konstantin and Rolf, Esther and Robinson, Caleb and Mackey, Lester and Ru{\ss}wurm, Marc},
  booktitle={Proceedings of the AAAI Conference on Artificial Intelligence},
  volume={39},
  issue={4},
  pages={4347--4355},
  year={2025},
}

@article{clipsurgery,
title = {A closer look at the explainability of Contrastive language-image pre-training},
journal = {Pattern Recognition},
volume = {162},
pages = {111409},
year = {2025},
issn = {0031-3203},
doi = {https://doi.org/10.1016/j.patcog.2025.111409},
url = {https://www.sciencedirect.com/science/article/pii/S003132032500069X},
author = {Yi Li and Hualiang Wang and Yiqun Duan and Jiheng Zhang and Xiaomeng Li}
}

@article{mai2022review,
author = {Gengchen Mai and Krzysztof Janowicz and Yingjie Hu and Song Gao and Bo Yan and Rui Zhu and Ling Cai and Ni Lao},
title = {A review of location encoding for GeoAI: methods and applications},
journal = {International Journal of Geographical Information Science},
volume = {36},
number = {4},
pages = {639--673},
year = {2022},
publisher = {Taylor \& Francis},
doi = {10.1080/13658816.2021.2004602},
URL = { https://doi.org/10.1080/13658816.2021.2004602},
}

@inproceedings{combicam,
  title={Combi-{CAM}: A Novel Multi-Layer Approach for Explainable Image Geolocalization},
  author={Faget, David and Lisani, Jos{\'e} Luis and Colom, Miguel},
  booktitle={21st International Conference on Computer Vision Theory and Applications},
  volume={1},
  pages={275--281},
  year={2026},
  organization={SCITEPRESS-Science and Technology Publications}
}

@article{li2022eclip,
  title={Exploring visual interpretability for contrastive language-image pre-training},
  author={Li, Yi and Wang, Hualiang and Duan, Yiqun and Xu, Hang and Li, Xiaomeng},
  journal={arXiv preprint arXiv:2209.07046},
  year={2022}
}

@article{brown2025alphaearth,
  title={AlphaEarth Foundations: An embedding field model for accurate and efficient global mapping from sparse label data},
  author={Christopher F. Brown and Michal Kazmierski and Valerie J. Pasquarella and William J. Rucklidge and Masha Samsikova and Chenhui Zhang and Evan Shelhamer and Estefania Lahera and Olivia Wiles and Simon Ilyushchenko and Noel Gorelick and Lihui Zhang and Sophia Alj and Emily Schechter and Sean Askay and Oliver Guinan and Rebecca Moore and Alexis Boukouvalas and Pushmeet Kohli},
  journal={arXiv preprint arXiv:2507.22291},
  year={2025}
}

@inproceedings{selvaraju2017grad,
  title={{Grad-CAM}: Visual explanations from deep networks via gradient-based localization},
  author={Selvaraju, Ramprasaath R and Cogswell, Michael and Das, Abhishek and Vedantam, Ramakrishna and Parikh, Devi and Batra, Dhruv},
  booktitle={Proceedings of the IEEE International Conference on Computer Vision},
  pages={618--626},
  year={2017}
}

@inproceedings{chen2020adapting,
  title={Adapting {Grad-CAM} for embedding networks},
  author={Chen, Lei and Chen, Jianhui and Hajimirsadeghi, Hossein and Mori, Greg},
  booktitle={Proceedings of the IEEE/CVF Winter Conference on Applications of Computer Vision},
  pages={2794--2803},
  year={2020}
}

@article{bhalla2024interpreting,
 author = {Bhalla, Usha and Oesterling, Alex and Srinivas, Suraj and Calmon, Flavio P. and Lakkaraju, Himabindu},
 booktitle = {Advances in Neural Information Processing Systems},
 doi = {10.52202/079017-2678},
 pages = {84298--84328},
 title = {Interpreting CLIP with Sparse Linear Concept Embeddings (SpLiCE)},
 url = {https://proceedings.neurips.cc/paper_files/paper/2024/file/996bef37d8a638f37bdfcac2789e835d-Paper-Conference.pdf},
 volume = {37},
 year = {2024}
}

@inproceedings{rao2025measuring,
title={Measuring the Intrinsic Dimension of Earth Representations},
author={Arjun Rao and Marc Ru{\ss}wurm and Konstantin Klemmer and Esther Rolf},
booktitle={The Fourteenth International Conference on Learning Representations},
year={2026},
url={https://openreview.net/forum?id=gQPD83DrGp}
}

@inproceedings{pach2025sparse,
 author = {Pach, Mateusz and Karthik, Shyamgopal and Bouniot, Quentin and Belongie, Serge and Akata, Zeynep},
 booktitle = {Advances in Neural Information Processing Systems},
 pages = {95706--95742},
 title = {Sparse Autoencoders Learn Monosemantic Features in Vision-Language Models},
 url = {https://proceedings.neurips.cc/paper_files/paper/2025/file/89e83382abeee53b932a6df62edbf9cc-Paper-Conference.pdf},
 volume = {38},
 year = {2025}
}

@inproceedings{vivanco2023geoclip,
 author = {Vivanco Cepeda, Vicente and Nayak, Gaurav Kumar and Shah, Mubarak},
 booktitle = {Advances in Neural Information Processing Systems},
 pages = {8690--8701},
 title = {GeoCLIP: Clip-Inspired Alignment between Locations and Images for Effective Worldwide Geo-localization},
 url = {https://proceedings.neurips.cc/paper_files/paper/2023/file/1b57aaddf85ab01a2445a79c9edc1f4b-Paper-Conference.pdf},
 volume = {36},
 year = {2023}
}

@inproceedings{sinr,
  title = 	 {Spatial Implicit Neural Representations for Global-Scale Species Mapping},
  author =       {Cole, Elijah and Horn, Grant Van and Lange, Christian and Shepard, Alexander and Leary, Patrick and Perona, Pietro and Loarie, Scott and Mac Aodha, Oisin},
  booktitle = 	 {Proceedings of the 40th International Conference on Machine Learning},
  pages = 	 {6320--6342},
  year = 	 {2023},
  volume = 	 {202},
  series = 	 {Proceedings of Machine Learning Research},
  month = 	 {23--29 Jul},
  publisher =    {PMLR},
  url = 	 {https://proceedings.mlr.press/v202/cole23a.html},
}

@article{klemmer2025earth,
  title={Earth Embeddings: Towards {AI}-centric Representations of our Planet},
  author={Klemmer, Konstantin and Rolf, Esther and Russwurm, Marc and Camps-Valls, Gustau and Czerkawski, Mikolaj and Ermon, Stefano and Francis, Alistair and Jacobs, Nathan and Kerner, Hannah Rae and Mackey, Lester and Mai, Gengchen and Mac Aodha, Oisin and Reichstein, Markus and  Robinson, Caleb and Rolnick, David and Shelhamer,  Evan and Sitzmann, Vincent and  Tuia, Devis and Zhu, Xiaoxiang },
  year={2025},
  journal={EarthArXiv},
  doi = {10.31223/X5HX9S},
  url = {https://doi.org/10.31223/X5HX9S}
}

@inproceedings{climplicit,
  title={Climplicit: Climatic Implicit Embeddings for Global Ecological Tasks},
  author={Dollinger, Johannes and Robert, Damien and Plekhanova, Elena and Drees, Lukas and Wegner, Jan Dirk},
  booktitle={ICLR 2025 Workshop on Tackling Climate Change with Machine Learning},
  url={https://www.climatechange.ai/papers/iclr2025/44},
  year={2025}
}

@inproceedings{cunningham2023sparse,
title={Sparse Autoencoders Find Highly Interpretable Features in Language Models},
author={Robert Huben and Hoagy Cunningham and Logan Riggs Smith and Aidan Ewart and Lee Sharkey},
booktitle={The Twelfth International Conference on Learning Representations},
year={2024},
url={https://openreview.net/forum?id=F76bwRSLeK}
}

@inproceedings{
bussmann2024batchtopk,
title={BatchTopK Sparse Autoencoders},
author={Bart Bussmann and Patrick Leask and Neel Nanda},
booktitle={NeurIPS 2024 Workshop on Scientific Methods for Understanding Deep Learning},
year={2024},
url={https://openreview.net/forum?id=d4dpOCqybL}
}

@inproceedings{liang2022mind,
 author = {Liang, Victor Weixin and Zhang, Yuhui and Kwon, Yongchan and Yeung, Serena and Zou, James},
 booktitle = {Advances in Neural Information Processing Systems},
 pages = {17612--17625},
 title = {Mind the Gap: Understanding the Modality Gap in Multi-modal Contrastive Representation Learning},
 url = {https://proceedings.neurips.cc/paper_files/paper/2022/file/702f4db7543a7432431df588d57bc7c9-Paper-Conference.pdf},
 volume = {35},
 year = {2022}
}

@article{ivic2019artificial,
author = {Ivi\'c, Majda},
title = {Artificial Intelligence and Geospatial Analysis in Disaster Management},
journal = {The International Archives of the Photogrammetry, Remote Sensing and Spatial Information Sciences},
volume = {XLII-3/W8},
year = {2019},
pages = {161--166},
url = {https://isprs-archives.copernicus.org/articles/XLII-3-W8/161/2019/},
doi = {10.5194/isprs-archives-XLII-3-W8-161-2019}
}

@article{
lam2023learning,
author = {Remi Lam  and Alvaro Sanchez-Gonzalez  and Matthew Willson  and Peter Wirnsberger  and Meire Fortunato  and Ferran Alet  and Suman Ravuri  and Timo Ewalds  and Zach Eaton-Rosen  and Weihua Hu  and Alexander Merose  and Stephan Hoyer  and George Holland  and Oriol Vinyals  and Jacklynn Stott  and Alexander Pritzel  and Shakir Mohamed  and Peter Battaglia },
title = {Learning skillful medium-range global weather forecasting},
journal = {Science},
volume = {382},
number = {6677},
pages = {1416-1421},
year = {2023},
doi = {10.1126/science.adi2336},
URL = {https://www.science.org/doi/abs/10.1126/science.adi2336}}

@article{tolan2024very,
title = {Very high resolution canopy height maps from {RGB} imagery using self-supervised vision transformer and convolutional decoder trained on aerial lidar},
journal = {Remote Sensing of Environment},
volume = {300},
pages = {113888},
year = {2024},
issn = {0034-4257},
doi = {https://doi.org/10.1016/j.rse.2023.113888},
url = {https://www.sciencedirect.com/science/article/pii/S003442572300439X},
author = {Jamie Tolan and Hung-I Yang and Benjamin Nosarzewski and Guillaume Couairon and Huy V. Vo and John Brandt and Justine Spore and Sayantan Majumdar and Daniel Haziza and Janaki Vamaraju and Theo Moutakanni and Piotr Bojanowski and Tracy Johns and Brian White and Tobias Tiecke and Camille Couprie}
}

@inproceedings{aiken2023fairness,
author = {Aiken, Emily and Rolf, Esther and Blumenstock, Joshua},
title = {Fairness and representation in satellite-based poverty maps: evidence of urban-rural disparities and their impacts on downstream policy},
year = {2023},
url = {https://doi.org/10.24963/ijcai.2023/653},
doi = {10.24963/ijcai.2023/653},
booktitle = {Proceedings of the Thirty-Second International Joint Conference on Artificial Intelligence},
articleno = {653},
numpages = {9},
series = {IJCAI '23}
}

@inproceedings{mac2019presence,
  title={Presence-only geographical priors for fine-grained image classification},
  author={Mac Aodha, Oisin and Cole, Elijah and Perona, Pietro},
  booktitle={Proceedings of the IEEE/CVF International Conference on Computer Vision},
  pages={9596--9606},
  year={2019}
}

@inproceedings{russwurm2023geographic,
 author = {Ru{\ss}wurm, Marc and Klemmer, Konstantin and Rolf, Esther and Zbinden, Robin and Tuia, Devis},
 booktitle = {International Conference on Learning Representations},
 pages = {1746--1759},
 title = {Geographic Location Encoding with Spherical Harmonics and Sinusoidal Representation Networks},
 url = {https://proceedings.iclr.cc/paper_files/paper/2024/file/073c8584ef86bee26fe9d639ec648e28-Paper-Conference.pdf},
 volume = {2024},
 year = {2024}
}

@inproceedings{sitzmann2020implicit,
 author = {Sitzmann, Vincent and Martel, Julien and Bergman, Alexander and Lindell, David and Wetzstein, Gordon},
 booktitle = {Advances in Neural Information Processing Systems},
 pages = {7462--7473},
 title = {Implicit Neural Representations with Periodic Activation Functions},
 url = {https://proceedings.neurips.cc/paper_files/paper/2020/file/53c04118df112c13a8c34b38343b9c10-Paper.pdf},
 volume = {33},
 year = {2020}
}

@article{stewart2025torchgeo,
author = {Stewart, Adam J. and Robinson, Caleb and Corley, Isaac A. and Ortiz, Anthony and Lavista Ferres, Juan M. and Banerjee, Arindam},
title = {TorchGeo: Deep Learning With Geospatial Data},
year = {2025},
issue_date = {December 2025},
publisher = {Association for Computing Machinery},
address = {New York, NY, USA},
volume = {11},
number = {4},
issn = {2374-0353},
url = {https://doi.org/10.1145/3707459},
doi = {10.1145/3707459},
journal = {ACM Trans. Spatial Algorithms Syst.},
articleno = {15},
numpages = {28},
}

@inproceedings{hernandez2021natural,
title={Natural Language Descriptions of Deep Features},
author={Evan Hernandez and Sarah Schwettmann and David Bau and Teona Bagashvili and Antonio Torralba and Jacob Andreas},
booktitle={International Conference on Learning Representations},
year={2022},
url={https://openreview.net/forum?id=NudBMY-tzDr}
}

@inproceedings{moayeri2023text,
  title={Text-to-concept (and back) via cross-model alignment},
  author={Moayeri, Mazda and Rezaei, Keivan and Sanjabi, Maziar and Feizi, Soheil},
  booktitle={International Conference on Machine Learning},
  pages={25037--25060},
  year={2023},
  organization={PMLR}
}

@inproceedings{
oikarinen2022clip,
title={{CLIP}-Dissect: Automatic Description of Neuron Representations in Deep Vision Networks},
author={Tuomas Oikarinen and Tsui-Wei Weng},
booktitle={The Eleventh International Conference on Learning Representations },
year={2023},
url={https://openreview.net/forum?id=iPWiwWHc1V}
}

@inproceedings{Radford2021LearningTV,
  title = 	 {Learning Transferable Visual Models From Natural Language Supervision},
  author =       {Radford, Alec and Kim, Jong Wook and Hallacy, Chris and Ramesh, Aditya and Goh, Gabriel and Agarwal, Sandhini and Sastry, Girish and Askell, Amanda and Mishkin, Pamela and Clark, Jack and Krueger, Gretchen and Sutskever, Ilya},
  booktitle = 	 {Proceedings of the 38th International Conference on Machine Learning},
  pages = 	 {8748--8763},
  year = 	 {2021},
  volume = 	 {139},
  series = 	 {Proceedings of Machine Learning Research},
  month = 	 {18--24 Jul},
  publisher =    {PMLR},
  url = 	 {https://proceedings.mlr.press/v139/radford21a.html}
}

@inproceedings{shraman2022translocatordata,
author="Pramanick, Shraman
and Nowara, Ewa M.
and Gleason, Joshua
and Castillo, Carlos D.
and Chellappa, Rama",
title={Where in the World Is This Image? Transformer-Based Geo-localization in the Wild},
  booktitle={European Conference on Computer Vision},
  pages={196--215},
  year={2022},
url={https://doi.org/10.1007/978-3-031-19839-7_12}
}

@article{vandermaaten2008tsne,
  author  = {Laurens van der Maaten and Geoffrey Hinton},
  title   = {Visualizing Data using {t-SNE}},
  journal = {Journal of Machine Learning Research},
  year    = {2008},
  volume  = {9},
  number  = {86},
  pages   = {2579--2605},
  url     = {http://jmlr.org/papers/v9/vandermaaten08a.html}
}

@inproceedings{martin1996dbscan,
author = {Ester, Martin and Kriegel, Hans-Peter and Sander, J\"{o}rg and Xu, Xiaowei},
title = {A density-based algorithm for discovering clusters in large spatial databases with noise},
year = {1996},
publisher = {AAAI Press},
booktitle = {Proceedings of the Second International Conference on Knowledge Discovery and Data Mining},
pages = {226–231},
numpages = {6},
series = {KDD'96}
}

@inproceedings{he2015resnet,
      title={Deep Residual Learning for Image Recognition}, 
      author={He, Kaiming and Zhang, Xiangyu and Ren, Shaoqing and Sun, Jian},
      booktitle={Proceedings of the IEEE conference on computer vision and pattern recognition},
      pages={770--778},
      year={2016}
}

@inproceedings{deng2009imagenet,
  author={Deng, Jia and Dong, Wei and Socher, Richard and Li, Li-Jia and Kai Li and Li Fei-Fei},
  booktitle={2009 IEEE Conference on Computer Vision and Pattern Recognition}, 
  title={ImageNet: A large-scale hierarchical image database}, 
  year={2009},
  volume={},
  number={},
  pages={248-255},
  doi={10.1109/CVPR.2009.5206848}}

@inproceedings{
hu2022lora,
title={Lo{RA}: Low-Rank Adaptation of Large Language Models},
author={Edward J Hu and yelong shen and Phillip Wallis and Zeyuan Allen-Zhu and Yuanzhi Li and Shean Wang and Lu Wang and Weizhu Chen},
booktitle={International Conference on Learning Representations},
year={2022},
url={https://openreview.net/forum?id=nZeVKeeFYf9}
}

@article{Text2Earth,
  author={Liu, Chenyang and Chen, Keyan and Zhao, Rui and Zou, Zhengxia and Shi, Zhenwei},
  journal={IEEE Geoscience and Remote Sensing Magazine}, 
  title={Text2Earth: Unlocking text-driven remote sensing image generation with a global-scale dataset and a foundation model}, 
  year={2025},
  volume={13},
  number={3},
  pages={238-259},
  doi={10.1109/MGRS.2025.3560455}}

@article{achiam2023gpt,
  title={{GPT}-4 technical report},
  author={Achiam, Josh and Adler, Steven and Agarwal, Sandhini and Ahmad, Lama and Akkaya, Ilge and Aleman, Florencia Leoni and Almeida, Diogo and Altenschmidt, Janko and Altman, Sam and Anadkat, Shyamal and others},
  journal={arXiv preprint arXiv:2303.08774},
  year={2023}
}

@article{doersch2015,
author = {Doersch, Carl and Singh, Saurabh and Gupta, Abhinav and Sivic, Josef and Efros, Alexei A.},
title = {What makes {P}aris look like {P}aris?},
year = {2015},
issue_date = {December 2015},
publisher = {Association for Computing Machinery},
volume = {58},
number = {12},
url = {https://doi.org/10.1145/2830541},
doi = {10.1145/2830541},
journal = {Commun. ACM},
month = nov,
pages = {103–110},
numpages = {8}
}

@article{hohl2024survey,
  author={Höhl, Adrian and Obadic, Ivica and Fernández-Torres, Miguel-Ángel and Najjar, Hiba and Oliveira, Dario Augusto Borges and Akata, Zeynep and Dengel, Andreas and Zhu, Xiao Xiang},
  journal={IEEE Geoscience and Remote Sensing Magazine}, 
  title={Opening the Black Box: A systematic review on explainable artificial intelligence in remote sensing}, 
  year={2024},
  volume={12},
  number={4},
  pages={261-304},
  doi={10.1109/MGRS.2024.3467001}}

@article{jia2025towards,
  title={Towards Interpretable Geo-localization: a Concept-Aware Global Image-{GPS} Alignment Framework},
  author={Jia, Furong and Liu, Lanxin and Hou, Ce and Zhang, Fan and Liu, Xinyan and Liu, Yu},
  journal={arXiv preprint arXiv:2509.01910},
  year={2025}
}

@article{benavides2026earth,
  title={What on {Earth} is {AlphaEarth}? {Hierarchical} structure and functional interpretability for global land cover},
  author={Benavides-Martinez, Ivan Felipe and Guthrie, Justin and Arias, Jhon Edwin and Garces-Gomez, Yeison Alberto and Guzman-Alvis, Angela Ines and Portilla-Cabrera, Cristiam Victoriano and Mondal, Somnath and Allyn, Andrew J. and Ganguly, Auroop R.},
  journal={arXiv preprint arXiv:2603.16911},
  year={2026}
}

@inproceedings{shu2025survey,
    title = {A Survey on Sparse Autoencoders: Interpreting the Internal Mechanisms of Large Language Models},
    author = {Shu, Dong  and
      Wu, Xuansheng  and
      Zhao, Haiyan  and
      Rai, Daking  and
      Yao, Ziyu  and
      Liu, Ninghao  and
      Du, Mengnan},
    booktitle = {Findings of the Association for Computational Linguistics: EMNLP 2025},
    year = {2025},
    publisher = {Association for Computational Linguistics},
    url = {https://aclanthology.org/2025.findings-emnlp.89/},
    doi = {10.18653/v1/2025.findings-emnlp.89},
    pages = {1690--1712},
}

@inproceedings{wang2026proxy,
    author    = {Wang, Zhongying and Lane, Kevin and Cai, Levi and Karimzadeh, Morteza and Rolf, Esther},
    title     = {A Proxy Consistency Loss for Grounded Fusion of Earth Observation and Location Encoders},
    booktitle = {Proceedings of the IEEE/CVF Conference on Computer Vision and Pattern Recognition (CVPR) Workshops},
    month     = {June},
    year      = {2026},
    pages     = {8075-8084}
}

@inproceedings{zhai2022lit,
    author    = {Zhai, Xiaohua and Wang, Xiao and Mustafa, Basil and Steiner, Andreas and Keysers, Daniel and Kolesnikov, Alexander and Beyer, Lucas},
    title     = {{LiT}: Zero-Shot Transfer With Locked-Image Text Tuning},
    booktitle = {Proceedings of the IEEE/CVF Conference on Computer Vision and Pattern Recognition (CVPR)},
    month     = {June},
    year      = {2022},
    pages     = {18123-18133}
}

@InProceedings{dubey2021adaptive,
    author    = {Dubey, Abhimanyu and Ramanathan, Vignesh and Pentland, Alex and Mahajan, Dhruv},
    title     = {Adaptive Methods for Real-World Domain Generalization},
    booktitle = {Proceedings of the IEEE/CVF Conference on Computer Vision and Pattern Recognition (CVPR)},
    month     = {June},
    year      = {2021},
    pages     = {14340-14349}
}

@article{zhu2026foundations,
author={Zhu, Xiao Xiang
and Xiong, Zhitong
and Wang, Yi
and Stewart, Adam J.
and Heidler, Konrad
and Wang, Yuanyuan
and Yuan, Zhenghang
and Dujardin, Thomas
and Xu, Qingsong
and Shi, Yilei},
title={On the foundations of Earth foundation models},
journal={Communications Earth {\&} Environment},
year={2026},
month={Jan},
day={08},
volume={7},
number={1},
pages={103},
doi={10.1038/s43247-025-03127-x},
url={https://doi.org/10.1038/s43247-025-03127-x}
}

@article{mcinnes2018umap-software,
  title={{UMAP}: Uniform Manifold Approximation and Projection},
  author={McInnes, Leland and Healy, John and Saul, Nathaniel and Gro{\ss}berger, Lukas},
  journal={The Journal of Open Source Software},
  volume={3},
  number={29},
  pages={861},
  year={2018}
}

@inproceedings{ribeiro2016should,
author = {Ribeiro, Marco Tulio and Singh, Sameer and Guestrin, Carlos},
title = {``{Why} Should {I} Trust You?": Explaining the Predictions of Any Classifier},
year = {2016},
publisher = {Association for Computing Machinery},
address = {New York, NY, USA},
url = {https://doi.org/10.1145/2939672.2939778},
doi = {10.1145/2939672.2939778},
booktitle = {Proceedings of the 22nd ACM SIGKDD International Conference on Knowledge Discovery and Data Mining},
pages = {1135–1144},
numpages = {10},
location = {San Francisco, California, USA},
series = {KDD '16}
}

@inproceedings{lundberg2017unified,
 author = {Lundberg, Scott M. and Lee, Su-In},
 booktitle = {Advances in Neural Information Processing Systems},
 title = {A Unified Approach to Interpreting Model Predictions},
 url = {https://proceedings.neurips.cc/paper_files/paper/2017/file/8a20a8621978632d76c43dfd28b67767-Paper.pdf},
 volume = {30},
 year = {2017}
}

@inproceedings{zhou2016learning,
author = {Zhou, Bolei and Khosla, Aditya and Lapedriza, Agata and Oliva, Aude and Torralba, Antonio},
title = {Learning Deep Features for Discriminative Localization},
booktitle = {Proceedings of the IEEE Conference on Computer Vision and Pattern Recognition (CVPR)},
month = {June},
year = {2016}
}

@inproceedings{bahdanau2015neural,
  author       = {Dzmitry Bahdanau and
                  Kyunghyun Cho and
                  Yoshua Bengio},
  title        = {Neural Machine Translation by Jointly Learning to Align and Translate},
  booktitle    = {Proceedings of the 3rd International Conference on Learning Representations (ICLR)},
  year         = {2015},
  url          = {http://arxiv.org/abs/1409.0473},
}

@inproceedings{vaswani2017attention,
 author = {Vaswani, Ashish and Shazeer, Noam and Parmar, Niki and Uszkoreit, Jakob and Jones, Llion and Gomez, Aidan N and Kaiser, \L ukasz and Polosukhin, Illia},
 booktitle = {Advances in Neural Information Processing Systems},
 pages = {},
 title = {Attention is All you Need},
 url = {https://proceedings.neurips.cc/paper_files/paper/2017/file/3f5ee243547dee91fbd053c1c4a845aa-Paper.pdf},
 volume = {30},
 year = {2017}
}

@ARTICLE{larson2017benchmarking,
  author={Larson, Martha and Soleymani, Mohammad and Gravier, Guillaume and Ionescu, Bogdan and Jones, Gareth J.F.},
  journal={IEEE MultiMedia}, 
  title={The Benchmarking Initiative for Multimedia Evaluation: MediaEval 2016}, 
  year={2017},
  volume={24},
  number={1},
  pages={93-96},
  doi={10.1109/MMUL.2017.9}}

@article{karger2017climatologies,
  title={Climatologies at high resolution for the earth’s land surface areas},
  author={Karger, Dirk Nikolaus and Conrad, Olaf and B{\"o}hner, J{\"u}rgen and Kawohl, Tobias and Kreft, Holger and Soria-Auza, Rodrigo Wilber and Zimmermann, Niklaus E and Linder, H. Peter and Kessler, Michael},
  journal={Scientific data},
  volume={4},
  number={1},
  pages={1--20},
  year={2017},
  publisher={Nature Publishing Group},
  doi = {10.1038/sdata.2017.122},
  URL = {https://doi.org/10.1038/sdata.2017.122}
}

@inproceedings{
mai2020multi,
title={Multi-Scale Representation Learning  for Spatial Feature Distributions using Grid Cells},
author={Gengchen Mai and Krzysztof Janowicz and Bo Yan and Rui Zhu and Ling Cai and Ni Lao},
booktitle={International Conference on Learning Representations},
year={2020},
url={https://openreview.net/forum?id=rJljdh4KDH}
}

@inproceedings{christie2018functional,
  title={Functional map of the world},
  author={Christie, Gordon and Fendley, Neil and Wilson, James and Mukherjee, Ryan},
  booktitle={Proceedings of the IEEE Conference on Computer Vision and Pattern Recognition},
  pages={6172--6180},
  year={2018}
}

@inproceedings{tancik2020fourier,
 author = {Tancik, Matthew and Srinivasan, Pratul and Mildenhall, Ben and Fridovich-Keil, Sara and Raghavan, Nithin and Singhal, Utkarsh and Ramamoorthi, Ravi and Barron, Jonathan and Ng, Ren},
 booktitle = {Advances in Neural Information Processing Systems},
 pages = {7537--7547},
 title = {Fourier Features Let Networks Learn High Frequency Functions in Low Dimensional Domains},
 url = {https://proceedings.neurips.cc/paper_files/paper/2020/file/55053683268957697aa39fba6f231c68-Paper.pdf},
 volume = {33},
 year = {2020}
}

\clearpage
\setcounter{page}{1}
\section{Appendix}
\label{sec:appendix}

\subsection{Reproducibility}\label{sec:reproducibility}

All code necessary to reproduce the experiments in this paper is available at \url{https://github.com/sricke/explainable-earth-embeddings}. All datasets, model code, and model weights are publicly available and can be downloaded from the links in Tables~\ref{tab:datasets}--\ref{tab:weights}. While most datasets and models are available under permissive open-source licenses, many lack any information about licensing. We have reached out to all authors to encourage them to make the license of their data or models clear. Datasets and models under open-source licenses will be contributed to TorchGeo~\citep{stewart2025torchgeo} to further lower the barrier of entry of reproducibility.

\begin{table}[htbp]
    \centering
    \caption{Datasets used in this work.}
    \label{tab:datasets}
    \begin{tabular}{lcl}
        \toprule
        \textbf{Dataset} & \textbf{License} & \textbf{URL} \\
        \midrule
        S2-100k & MIT & \url{https://hf.co/datasets/kklmmr/s2-100k} \\
        Im2GPS & ? & \url{https://www.cis.jhu.edu/~shraman/TransLocator} \\
        Git-10M & CC-BY-NC-ND-4.0 & \url{https://hf.co/datasets/lcybuaa/Git-10M} \\
        \bottomrule
    \end{tabular}
\end{table}

\begin{table}[htbp]
    \centering
    \caption{Model code used in this work.}
    \label{tab:code}
    \begin{tabular}{lcl}
        \toprule
        \textbf{Model} & \textbf{License} & \textbf{URL} \\
        \midrule
        ResNet-50 & BSD-3-Clause & \url{https://github.com/pytorch/vision} \\
        CSP-fMoW & ? & \url{https://github.com/gengchenmai/csp} \\
        
        GeoCLIP & MIT & \url{https://github.com/VicenteVivan/geo-clip} \\
        SatCLIP & MIT & \url{https://github.com/microsoft/satclip} \\
        Climplicit & MIT & \url{https://github.com/ecovision-uzh/climplicit} \\
        \bottomrule
    \end{tabular}
\end{table}

\begin{table}[htbp]
    \centering
    \caption{Model weights used in this work.}
    \label{tab:weights}
    \begin{tabular}{lcl}
        \toprule
        \textbf{Model} & \textbf{License} & \textbf{URL} \\
        \midrule
        ResNet-50 & ? & \url{https://download.pytorch.org/models} \\
        CSP-fMoW & ? & \url{https://gengchenmai.github.io/csp-website} \\
        
        GeoCLIP & MIT & \url{https://github.com/VicenteVivan/geo-clip} \\
        SatCLIP & MIT & \url{https://hf.co/microsoft/SatCLIP-ViT16-L10} \\
        Climplicit & CC-BY-4.0 & \url{https://hf.co/Jobedo/climplicit} \\
        \bottomrule
    \end{tabular}
\end{table}

All SAE experiments were performed on an NVIDIA A40 with 46~GB of RAM and 6 parallel workers for data loading. SatCLIP experiments ran for 2~hours, GeoCLIP experiments ran for 30~minutes, Climplicit experiments ran for 5~minutes, and xAI computation ran for 2~hours per location encoder. Combined, these experiments ran for 275~minutes for 3 different random seeds, resulting in 825~minutes in total.

The majority of SpLiCE and CLIP Surgery experiments were performed on an NVIDIA RTX 8000 with 48~GB of RAM. The location--text alignment of SatCLIP and Climplicit was performed for 55k steps and took approximately 4~days. GeoCLIP location--text alignment was not trained. SpLiCE evaluations took around 2~minutes. Inference of saliency maps takes approximately one second per image for the GeoCLIP and SatCLIP models. The CSP-fMoW experiments were performed on an NVIDIA H100 with 80~GB RAM and took 20~hours in total.

\subsection{Location Encoders and Embeddings}

\begin{table}[t]
\centering
\small
\caption{\textbf{Location encoders analyzed in this work.}}
\label{tab:location_encoders}
\begin{tabular}{lccc}
\toprule
Model & Encoding & Dim. & Pretraining Data \\
\midrule
GeoCLIP \citep{vivanco2023geoclip}  & RFF \citep{tancik2020fourier}  & 512  & MP-16 (Flickr) \citep{larson2017benchmarking} \\
SatCLIP \citep{klemmer2025satclip}  & Spherical Harmonics \citep{russwurm2023geographic}  & 256  & S2-100K (Sentinel-2) \citep{klemmer2025satclip} \\
Climplicit \citep{climplicit} & Sinusoidal   & 1024 & CHELSA (Climate Variables) \citep{karger2017climatologies} \\
CSP-fMoW   & Grid-Cell \citep{mai2020multi} & 512  & fMoW (DigitalGlobe Satellite Imagery) \citep{christie2018functional} \\
\bottomrule
\end{tabular}
\end{table}

In \Cref{tab:location_encoders}, we summarize the location encoders analyzed in this work. We report the pretraining data used to train the location encoder, which is an important consideration when considering the concepts encoded in location embeddings. We additionally report the dimension of each embedding, which can be compared to the number of concepts enforced in the sparsity constraint of SAEs as well as the number of active concepts resulting from SpLiCE (\Cref{tab:splice_quant}).

\subsection{Additional Model Training Details}\label{sec:training-details}

\paragraph{SAE and SpLiCE Evaluation.}

\begin{table}[t]
\centering
\small
\caption{\textbf{Experimental setup for evaluating SAE and SpLiCE Reconstruction.} We randomly sampled 10\% of the locations and their corresponding images from each evaluation dataset.}
\label{tab:sae_ms_evaluation_setup}
\begin{tabular}{lccc}
\toprule
\cmidrule(lr){2-4}
\textbf{Dataset} & \textbf{Eval Dataset} & \textbf{Samples ($N$)} & \textbf{Visual Encoder} \\
\midrule
UAR & S2-100K (Sentinel-2) \citep{klemmer2025satclip} & 9185 & MoCo ViT-S/16 (SatCLIP \citep{klemmer2025satclip}) \\
Human-visited & Geo-YFCC (natural) \citep{dubey2021adaptive} & 8094 & CLIP ViT-L/14 (GeoCLIP \citep{vivanco2023geoclip}) \\
\bottomrule
\end{tabular}
\end{table}

As shown in \Cref{tab:sae_ms_evaluation_setup}, we used two different datasets to evaluate SAE reconstruction (\Cref{tab:sae_reconstruction}) and SpLiCE reconstruction (\Cref{tab:splice_quant}). The first dataset consists of $100,000$ points distributed uniformly-at-random (UAR) across the landmasses. To evaluate the monosemanticity scores of the SAE, we sample 10\% from the samples from the S2-100K dataset, which consists of Sentinel-2 imagery. To compute the monosemanticity scores, we use embeddings from the SatCLIP visual encoder (MoCo ViT-S/16 \citep{klemmer2025satclip}). The second dataset we use comprises $100,000$ points sampled from the Geo-YFCC dataset \citep{dubey2021adaptive}, which contains natural imagery from \textbf{human-visited} locations. We also sample 10\% of the points from this dataset to evaluate the visual monosemanticity, and use the GeoCLIP image encoder (CLIP ViT-L/14 \citep{vivanco2023geoclip}) to obtain the image embeddings.

\paragraph{Location--Text Alignment} To create a shared location--text space, we aligned the OpenCLIP ViT-L text encoder~\citep{Radford2021LearningTV} to each location encoder (SatCLIP, Climplicit, CSP-fMoW). We train the text encoder using a symmetric CLIP loss while the location encoder is kept frozen. A linear projection head maps text embeddings to the shared representation space (with the same dimensionality of the embeddings from the location encoder). The text encoder is adapted using LoRA~\citep{hu2022lora} (rank 8 applied to the final 8 layers). 

Optimization is performed using AdamW with a learning rate of $1 \times 10^{-4}$ and weight decay of $0.01$. Training uses a batch size of 2048 with $4$ gradient accumulation steps for an effective batch size of $8192$. We use a cosine learning rate schedule. The logit temperature is learnable and is initialized at $0.07$. 

We train for 55k steps, validating every 500 steps and applying early stopping based on validation loss (with a patience of 5 validation checks). Training is conducted on NVIDIA RTX 8000s and H100s. 

\begin{figure}[htbp]
    \centering
    \begin{subfigure}{0.48\linewidth}
        \centering
        \includegraphics[width=\linewidth]{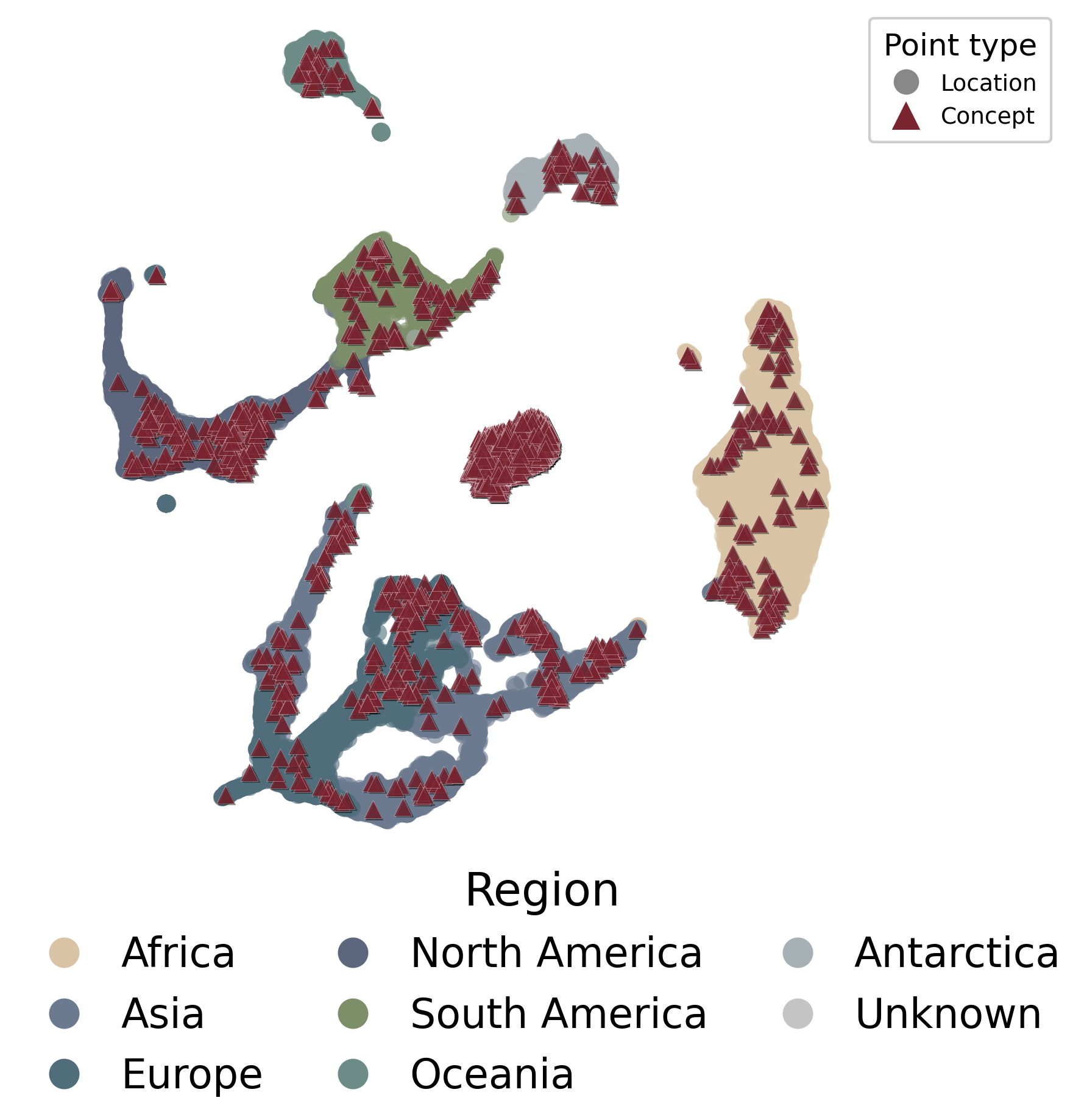}
        \caption{GeoCLIP (Geo-YFCC concepts)}
    \end{subfigure}
    \hfill
    \begin{subfigure}{0.48\linewidth}
        \centering
        \includegraphics[width=\linewidth]{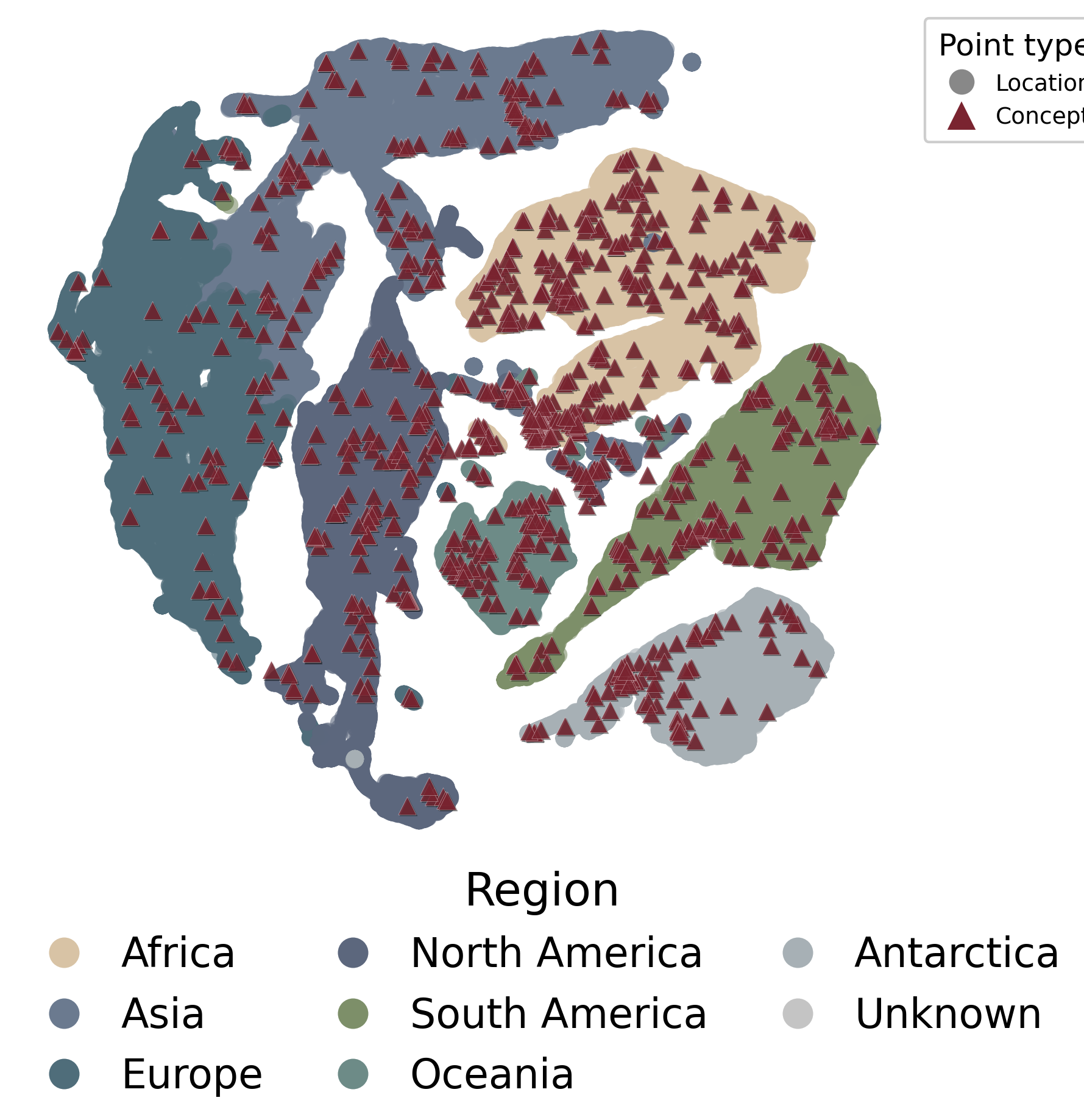}
        \caption{SatCLIP (Git-10M concepts)}
    \end{subfigure}
    \caption{\textbf{UMAP visualizations of GeoCLIP and SatCLIP embeddings} (with respective concept sets). Location embeddings are generated from a dense grid of $100,000$ points sampled uniformly at random over the landmasses. 
    }
    \label{fig:umap_comparison}
\end{figure}

\paragraph{Assessing and Visualizing Location--Text Alignment.} We show visualizations of alignment in \Cref{fig:umap_comparison} for GeoCLIP and SatCLIP to assess the success of the location--text alignment training for SatCLIP. UMAPs are created by first applying PCA to project down to 30 dimensions (to reduce noise), and then applying the UMAP algorithm~\citep{mcinnes2018umap-software} with 200 neighbors and minimum distance $0.1$ to capture global alignment structure. 

For GeoCLIP, we display the UMAP results using the Geo-YFCC concept set, which is generated by taking the 10,000 most common concepts from the Geo-YFCC dataset~\citep{dubey2021adaptive}. We visualize the GeoCLIP aligned space with this concept set, as the OpenCLIP ViT-L text encoder was trained on scraped image-text pairs, more closely mirroring the captions in Geo-YFCC. This concept set is overcomplete, containing non-geospatial concepts. For SpLiCE decompositions, we default to using a geospatial concept set (described in the main text). 

In these UMAP visualizations, we can see overlap between concepts and certain locations. Manual inspection of these figures can qualitatively show how aligned these spaces are. For example, for the GeoCLIP UMAP with the Geo-YFCC concepts, locations in Western North America have overlap with concepts such as ``rocky mountains'' and specific states in the Western US, whereas locations in South America correspond to concepts such as ``andes'' for the mountain range. On the other hand, the Git-10M concepts contain less location-specific terms, leading to overlap in concepts and locations that's less immediately interpretable in the SatCLIP UMAP. Additionally, in the GeoCLIP UMAP visualization, we see a cluster of solely concepts in the center, which tend to be non-geospatial terms. We don't see this for SatCLIP as we restrict ourselves to exclusively geospatial terms.

\new{\paragraph{Evaluation Regions for SpLiCE Concepts.}
We define four geographic regions using publicly available shapefiles.
The Russia and Indonesia shapefiles are sourced from
GADM~(\url{https://gadm.org/download\_country.html}).
Siberia is constructed by dissolving 12 level-1 Russian administrative regions
(\texttt{gadm41\_RUS\_1}), including Krasnoyarsk, Irkutsk, Buryat, Altay, and Tomsk, among others.
Bali is extracted from the level-1 Indonesian province shapefile (\texttt{gadm41\_IDN\_1}),
selecting the province named ``Bali.''
The Sahara shapefile is sourced from GISCarta~(\url{https://map.giscarta.com}) and reprojected
to WGS84 (EPSG:4326).
The Paris shapefile is derived from the French commune boundaries dataset on
data.gouv.fr~(\url{https://www.data.gouv.fr/datasets/contours-administratifs}),
selecting the commune named ``Paris.''}

\subsection{Method Details}\label{sec:method-details}

\subsubsection{Sparse AutoEncoders for Reconstructing Earth Embeddings}

Let $\vec{x} \in \mathbb{R}^{d_{ee}}$ denote an location embedding with dimension $d_{ee}$ ($ee$ stands for ``Earth embedding''). The SAE first maps $\vec{x}$ into a higher-dimensional latent space of dimension $d_l$ via an encoder-decoder architecture. The encoder computes a sparse latent representation $\vec{z} = \enc(\vec{x}) := \sigma(\mat{W}_\text{enc}^\top (\vec{x} - \vec{b}))$, where $\mat{W}_\text{enc} \in \mathbb{R}^{d_{ee} \times d_l}$ and $\sigma$ is applied elementwise. The decoder reconstructs the embedding as $\hat{\vec{x}} = \dec(\vec{z}) := \mat{W}_\text{dec}^\top\vec{z} + \vec{b}$.
The columns of $\mat{W}_\text{dec}^\top$ define a learned dictionary of feature directions $\{ \vec{c}_i \}_{i=1}^{d_l}$ in embedding space such that the reconstruction can be expressed as a sparse linear combination $\hat{\vec{x}} = \sum_{i=1}^{d_l} z_i \vec{c}_i + \vec{b}$. In this formulation, each location embedding is represented using only a small subset of dictionary elements, with sparsity enforced in the latent embedding $\vec{z}$.

The full SAE (encoder and decoder) is trained using a reconstruction loss combined with a sparsity penalty, $\mathcal{L}(\vec{x}) := \mathcal{R}(\vec{x}) + \lambda \mathcal{S}(\vec{x})$, where $\lambda$ controls the tradeoff between sparsity and reconstruction accuracy. After training, the decoder matrix serves as a learned dictionary of features: each column corresponds to a direction in embedding space that captures a distinct, approximately monosemantic \cite{pach2025sparse} feature, while the sparse latent representation $\vec{z} = \enc(\vec{x})$ determines which features are active for a given input.

\subsubsection{Sparse Linear Concept Embeddings (SpLiCE)}

To apply SpLiCE, we use a \textit{predefined} concept dictionary, $\mat{C} \in \mathbb{R}^{n_c \times d_{ee}}$, where $n_c$ is the number of concepts. Each row in $\mat{C}$ corresponds to an embedding in the same space as the location embedding $\vec{x} \in \mathbb{R}^{d_{ee}}$. This method uses a centered and normalized concept dictionary, given by $\mat{C} = [\sigma(f_\text{text}(\vec{x}_1^{\text{con}}) - \mu_\text{text}), \ldots, \sigma(f_\text{text}(\vec{x}_{n_c}^{\text{con}})-\mu_\text{text})]$, where $f_\text{text}$ denotes the text encoder, $\vec{x}_i^\text{con}$ a text concept, $\mu_\text{text}$ the (estimated) mean of text embeddings, and $\sigma(\cdot)$ denotes normalization. Given an input embedding $\vec{x}$, SpLiCE presents the following objective
\begin{equation}
    \min_{\vec{w} \in \mathbb{R}_+^{n_c}} \| \vec{w}\|_0 \, \text{s.t.} \, \langle \vec{x}, \sigma(\mat{C}\vec{w})\rangle \geq 1- \varepsilon.
\end{equation}
In practice, SpLiCE computes a sparse concept representation $\vec{w} \in \mathbb{R}^{n_c}$ by solving the following relaxation:
\begin{equation}\label{eqn: splice}
    \min_{\vec{w} \in \mathbb{R}^d} \| \mat{C}^\top \vec{w} - \vec{x} \|_2^2 + \lambda \| \vec{w} \|_1
\end{equation}
The reconstructed embedding is then given by $\hat{\vec{x}} = \sigma(\mat{C}\vec{w}^* + \mu_{\text{loc}})$.

\subsubsection{CLIP Surgery for Feature Attribution}

\edited{CLIP Surgery changes the attention schema to what the authors call \textit{consistent self-attention}, which replaces $Q$ and $K$ with $V$:
\[
    \text{Attn}(V,V,V) = \text{softmax}\!\left(\frac{VV^\top}{\sqrt{d}}\right)V
\]
Only features from this partial consistent self-attention are used, without the feed forward network (FFN), arguing that FFNs push features toward negatives when identifying positives, thus leading to opposite saliency maps. They implement these changes only to the last 6 layers of the vision model leaving other layers unchanged.}

\edited{After these changes to the vision model inference}, given an input image $\mathcal{I} \in \mathbb{R}^{H \times W \times C}$, the normalized output of the image encoder $\mathcal{F}_{i} \in \mathbb{R}^{N_{t} \times N_{d}}$, where $N_{t}$ corresponds to the number of output tokens (excluding the class token) and $N_{d}$ is the dimensionality of the aligned CLIP space.

\edited{Given} the normalized location encoder output $\mathcal{F}_{l} \in \mathbb{R}^{N_{d}}$, \edited{a noise reduction step is applied in which a mean noise estimate $\mathcal{F}_{n}$ is subtracted from the location encoder output}

\[
    \mathcal{F}_{l}' = \mathcal{F}_{l} - \mathcal{F}_{n}, \qquad \mathcal{F}_{l}' \in \mathbb{R}^{N_{d}}
\]

\edited{$\mathcal{F}_{n}$ can be obtained in one of two ways: (1) compute the mean over different classes or (2) proxied by the embedding of a ``null'' class (which in the case of text is an empty string).}

\edited{After this, }per-patch similarity scores $\mathcal{S}_{n}$ are obtained between the image and location representations:
\[
    \mathcal{S}_{n} = \mathcal{F}_{i} \cdot \mathcal{F}_{l}', \qquad \mathcal{S}_{n} \in \mathbb{R}^{N_{t}}
\]

Finally, function $\mathcal{U}$ is used to (1) reshape $\mathcal{S}_{n}$ to a square $ \in\mathbb{R}^{\sqrt{N_{t}} \times \sqrt{N_{t}}}$; (2) expand to original image shape $\mathbb{R}^{H \times W}$ through bilinear interpolation; and (3) min--max normalize values. 
\[
    \mathcal{S}_{f} = \mathcal{U}(\mathcal{S}_{n})
\]
where the final saliency map $\mathcal{S}_{f} \in \mathbb{R}^{H \times W}$ can be \edited{overlaid on the} original the image.

\edited{Inspired by} \citet{combicam}, we compute a per-layer saliency map $\mathcal{S}_{l}$ for each layer \edited{$l \in \mathcal{L}'$}, where \edited{$\mathcal{L}'$} denotes the same 6 layers to which CLIP Surgery is applied. The per-layer maps are then summed and min-max normalized to obtain the final saliency map \edited{$\tilde{\mathcal{S}}_{f}$}:
\edited{
\[
    \mathcal{S}_{\text{sum}} = \sum_{l \in \mathcal{L}'} \mathcal{S}_{l}
\]
}
\edited{
\[
    \tilde{\mathcal{S}}_{f} = \frac{\mathcal{S}_{\text{sum}} - \min(\mathcal{S}_{\text{sum}})}{\max(\mathcal{S}_{\text{sum}}) - \min(\mathcal{S}_{\text{sum}})}
\]
}

\subsection{Additional Results}

\subsubsection{Sparse Autoencoder Explanations}

\begin{figure}[!htbp]
    \centering
   \includegraphics[width=0.9\linewidth]{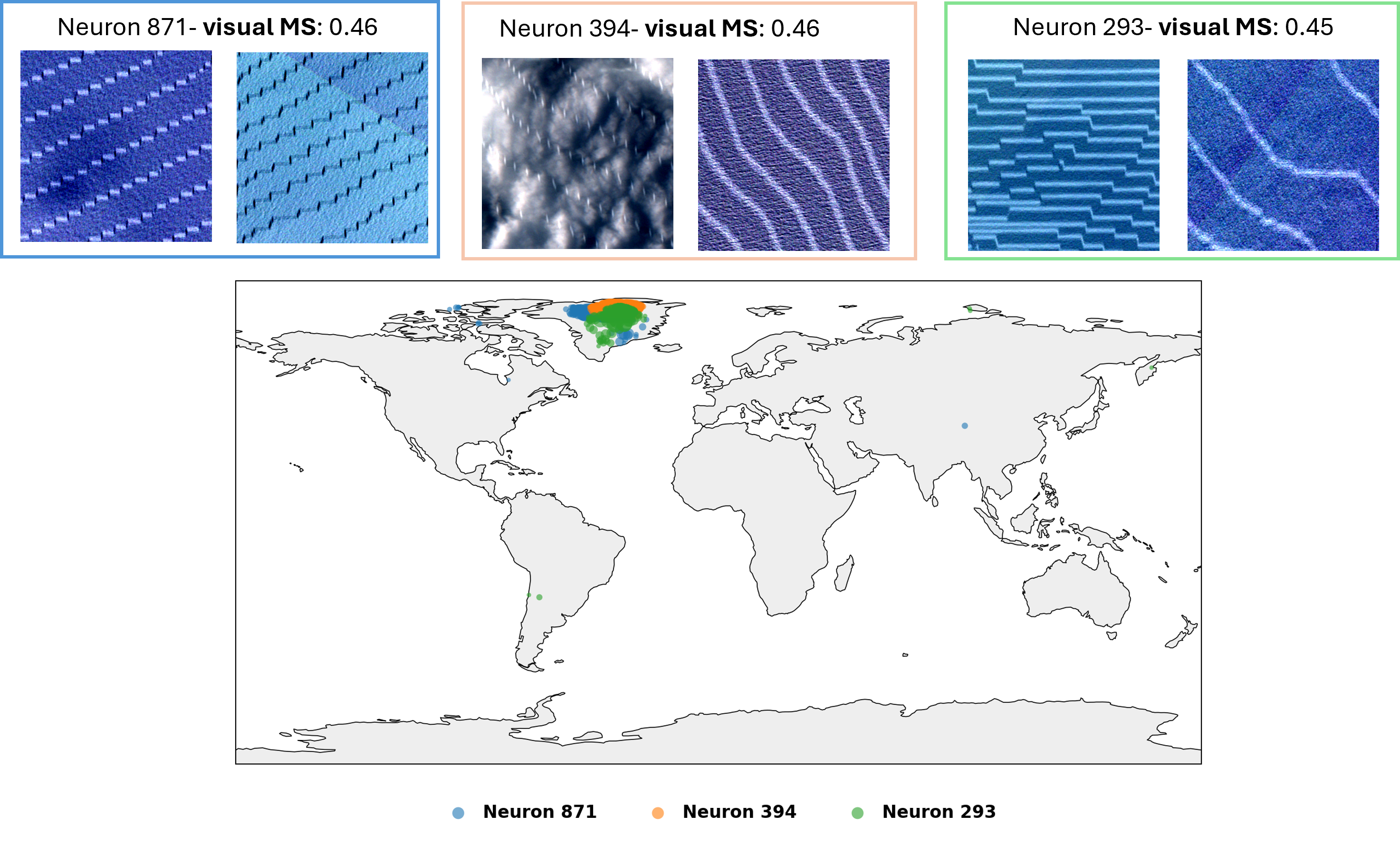}
    \caption{\textbf{Visual artifacts from the S2-100k dataset} around the Arctic region identified by analyzing the images for the strongest activating samples for neurons ranked among the top-10 visual monosemanticity values.}
    \label{fig:sae_visual_artifacts}
\end{figure}

We provide additional visualizations for the SAE method, showing that this method can be used to identify visual artifacts (\Cref{fig:sae_visual_artifacts}). 

\subsubsection{SpLiCE Natural Language Explanations}
\label{apx: sec: splice}

\new{\begin{figure}[thbp]
    \centering
    \setlength{\tabcolsep}{2pt}
    \renewcommand{\arraystretch}{0.95}
    \newcommand{\spliceimg}[1]{\raisebox{-0.5\height}{\includegraphics[width=0.22\linewidth]{#1}}}
    \newcommand{\spliceimgbig}[1]{\raisebox{-0.5\height}{\includegraphics[width=0.25\linewidth]{#1}}}
    \newcommand{\spliceimgxbig}[1]{\raisebox{-0.5\height}{\includegraphics[width=0.275\linewidth]{#1}}}
    \newcommand{\spliceimgsmall}[1]{\raisebox{-0.5\height}{\includegraphics[width=0.20\linewidth]{#1}}}
    \newcommand{\splicetag}[1]{
        \refstepcounter{subfigure}\label{fig:splice_#1}
        \small(\alph{subfigure})
    }
    \setcounter{subfigure}{0}
    \begin{tabular}{@{}l@{\hspace{3pt}}c@{\hspace{-3mm}}cc@{\hspace{-3mm}}c@{}}
        & \small Paris & \, \, \, \small Sahara & \, \, \, \small Siberia & \, \, \, \small Bali \\
        \raisebox{-0.5cm}{\rotatebox{90}{\small \textbf{Climplicit}}}  &
            \spliceimgbig{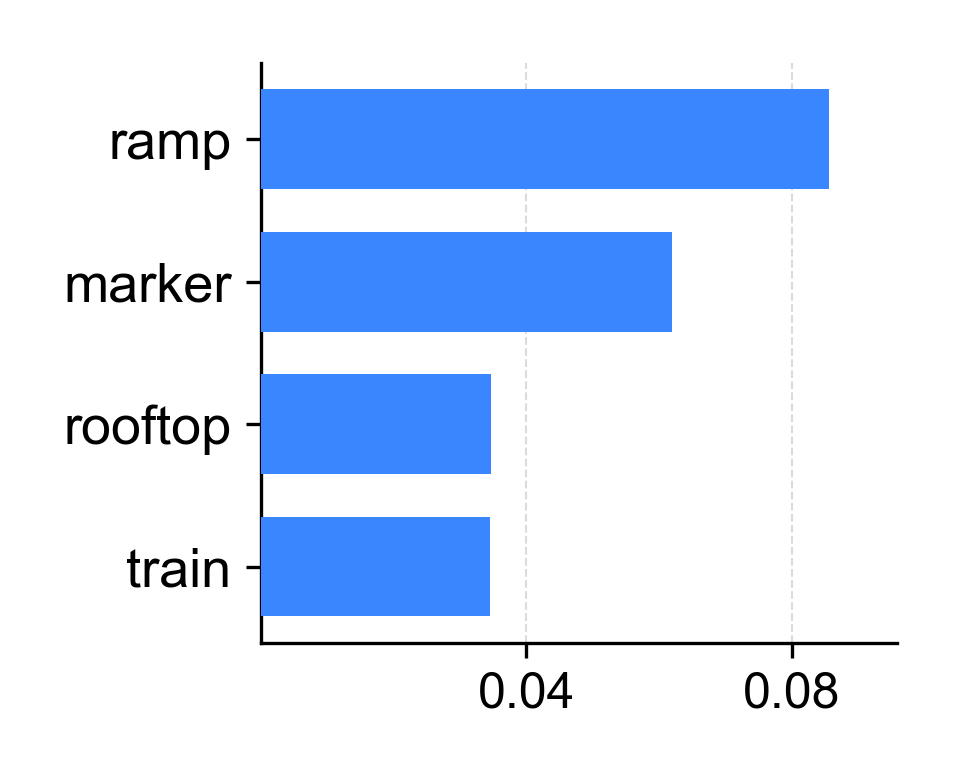} &
            \spliceimgbig{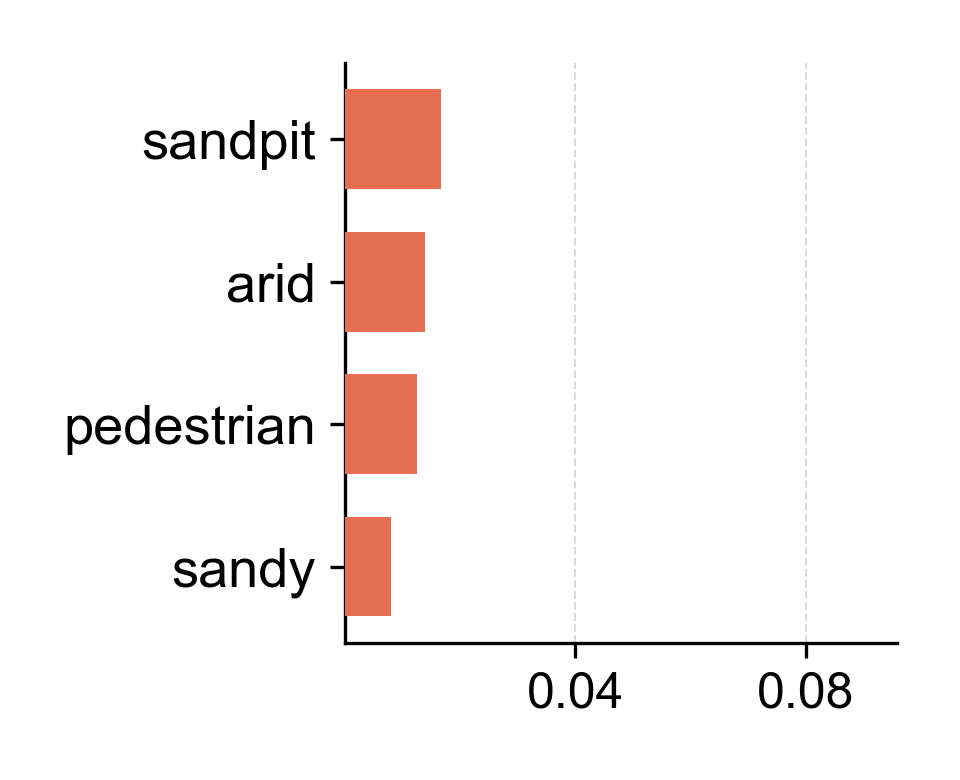} &
            \spliceimgbig{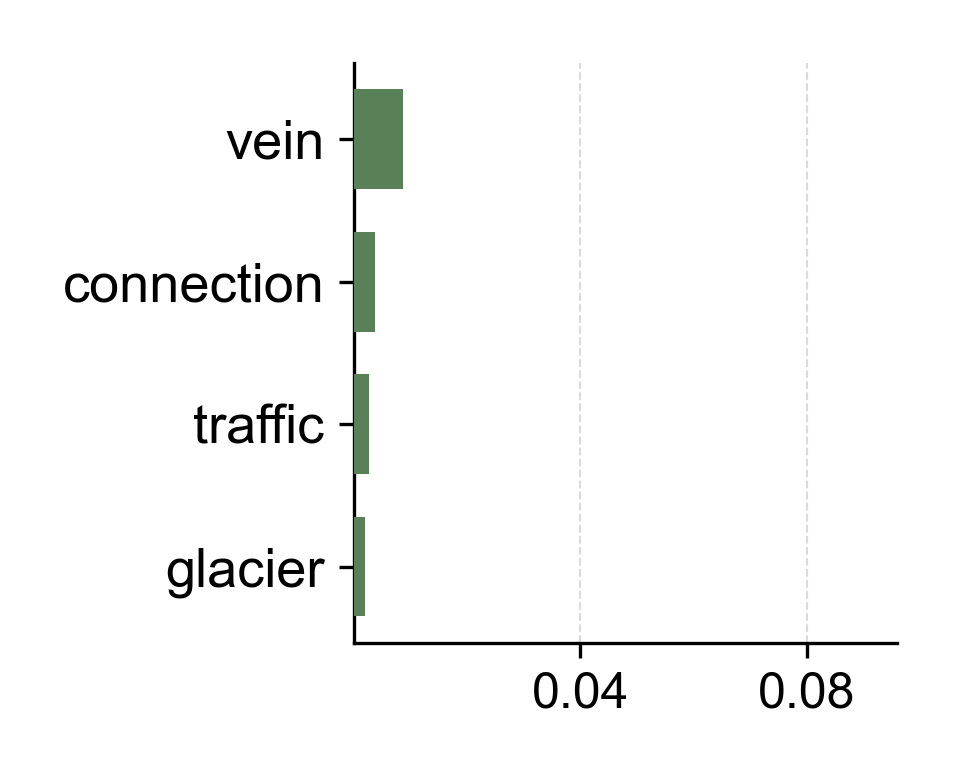} &
            \spliceimgbig{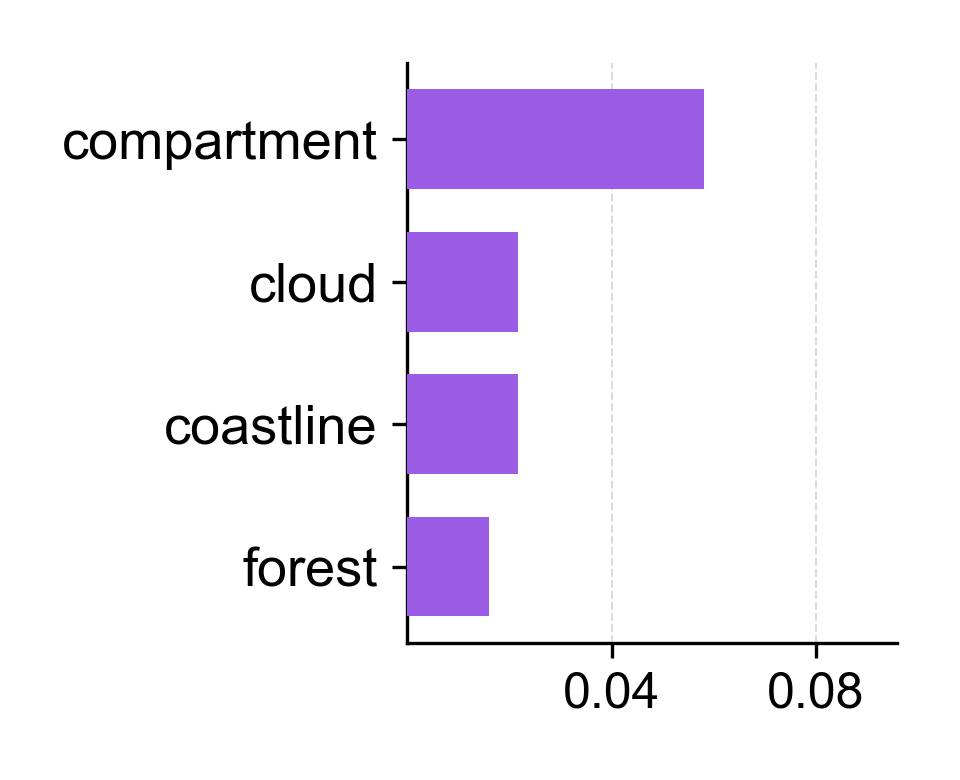} \\
        \raisebox{-0.5cm}{\rotatebox{90}{\small \textbf{CSP-fMoW}}}  &
            \spliceimgxbig{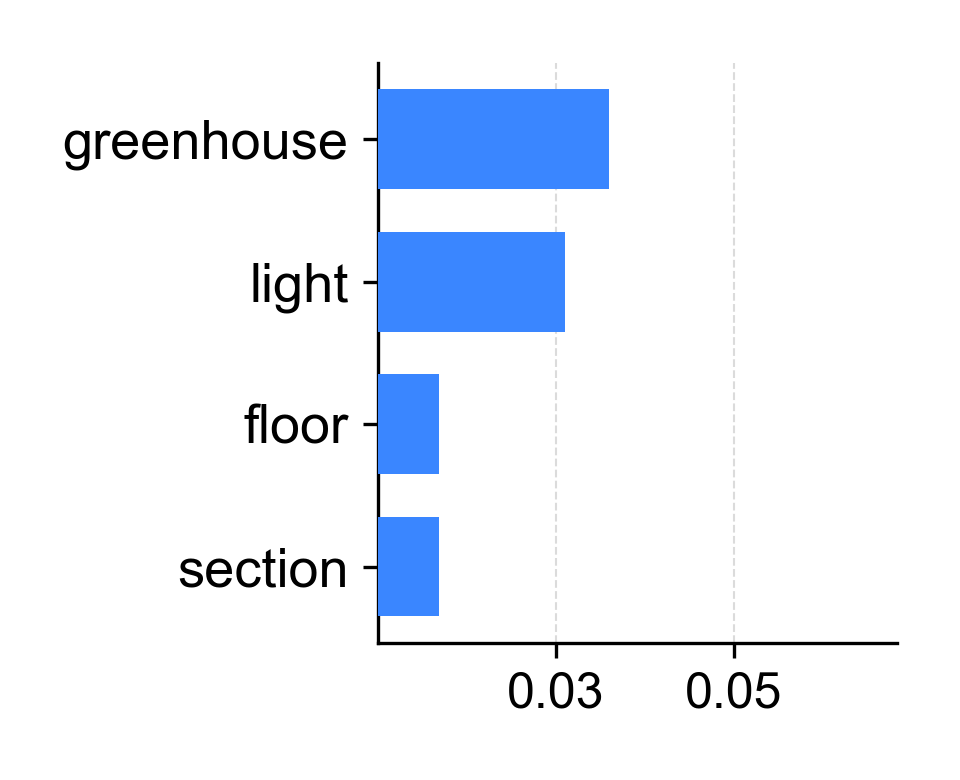} &
            \spliceimgbig{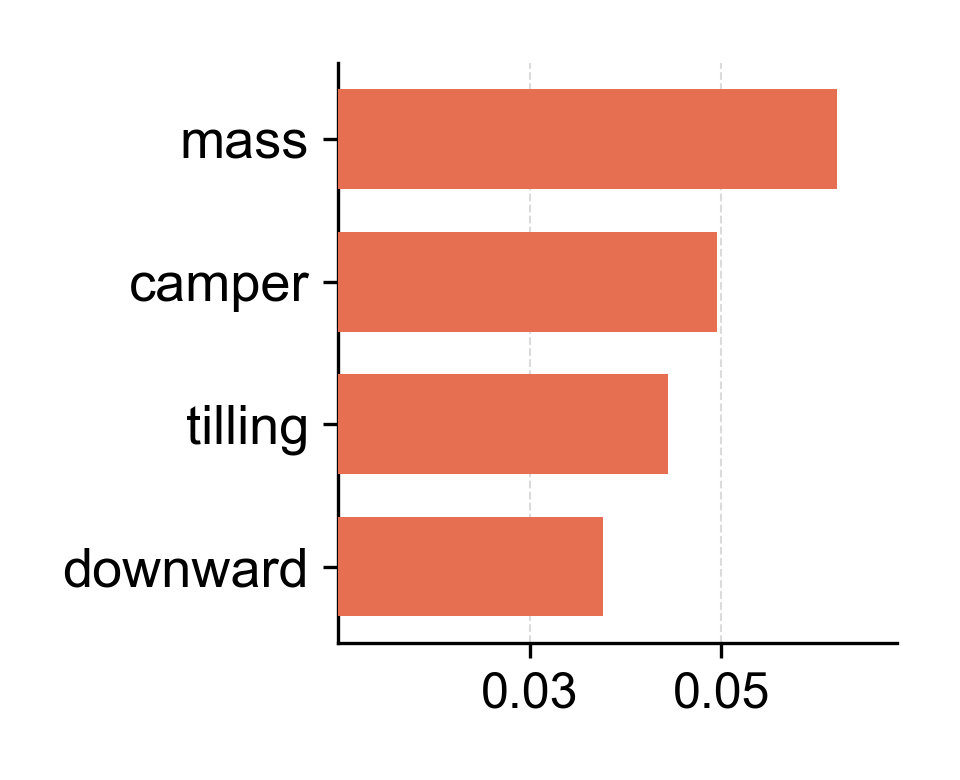} &
            \spliceimgbig{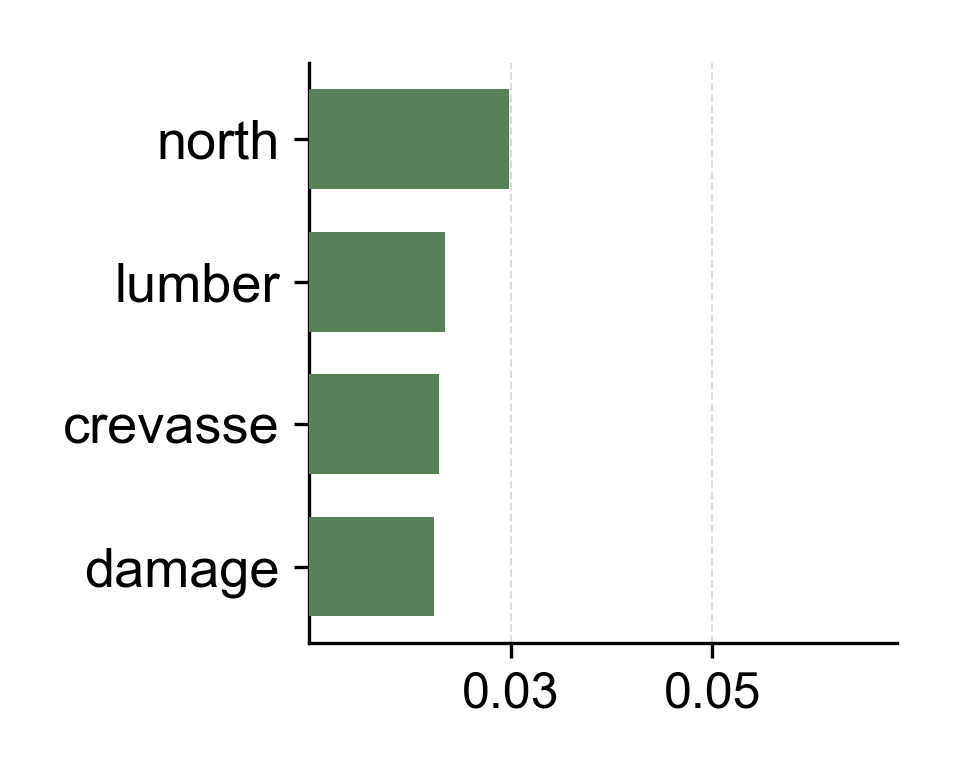} &
            \spliceimgbig{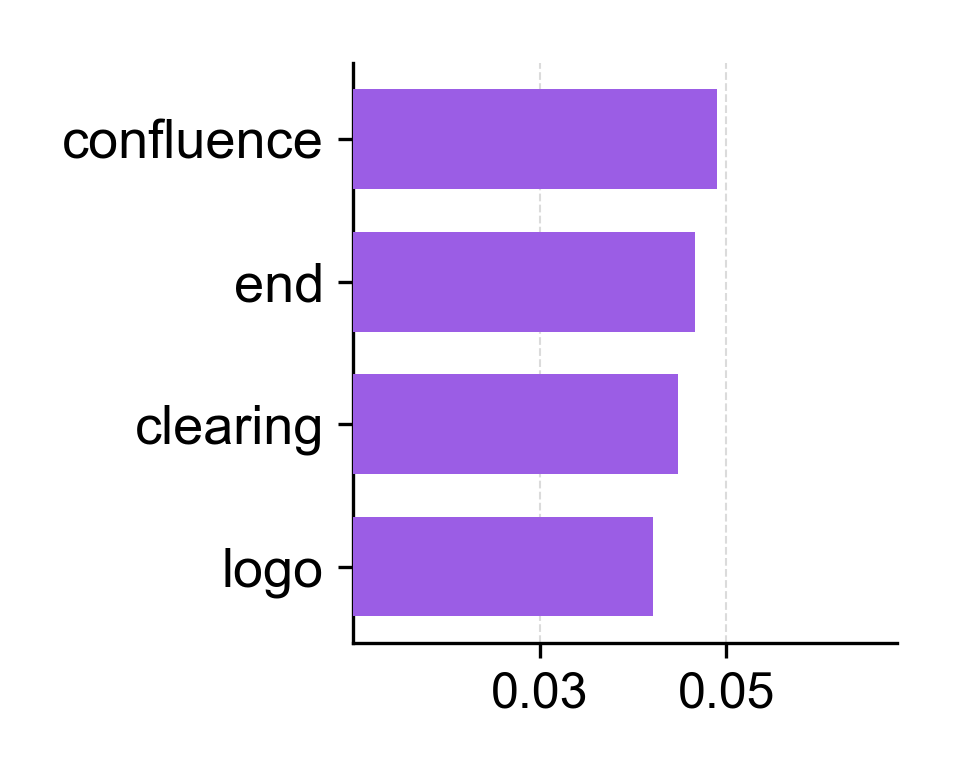} \\
    \end{tabular}
    \caption{\textbf{SpLiCE decompositions for Climplicit and CSP-fMoW.} Top-4 concepts in SpLiCE decompositions for Climplicit~(top) and CSP-fMoW~(bottom) across four regions (chosen to represent different land cover types). All decompositions use a shared concept basis derived from Git-10M.}
    \label{fig:splice_decompositions_appendix}
\end{figure}}

We first provide SpLiCE decomposition results for additional location encoders Climplicit and CSP-fMoW in \Cref{fig:splice_decompositions_appendix}. These SpLiCE results show that some of the concepts for these location embeddings seem less clearly aligned with underlying geographic semantics.

\begin{table}[tbp]
\centering

\caption{\textbf{Sparse concept decompositions preserve structure in location embedding space even on a small geospatial concept set.} Here, we use a subset of about 200 geospatial concepts from the Git-10M dataset. To compute the reconstruction metrics, we use the same methods as in \Cref{tab:splice_quant}. We use $\lambda = 0.175$ for GeoCLIP and SatCLIP and $\lambda=0.125$ for Climplicit and CSP-fMoW.}

\new{\setlength{\tabcolsep}{4pt}
\resizebox{\linewidth}{!}{%
\begin{tabular}{l ccc ccc}
\toprule
 & \multicolumn{3}{c}{\textbf{UAR}} & \multicolumn{3}{c}{\textbf{Human-visited}} \\
\cmidrule(lr){2-4} \cmidrule(lr){5-7}
\textbf{Model} &
\textbf{MSE} $\downarrow$ &
\textbf{Cos. Sim.} $\uparrow$ &
\textbf{\# Concepts} &
\textbf{MSE} $\downarrow$ &
\textbf{Cos. Sim.} $\uparrow$ &
\textbf{\# Concepts} \\
\midrule
GeoCLIP & 0.002 $\pm$ 0.000 & 0.451 $\pm$ 0.061 & 11.4 $\pm$ 3.1 & 0.002 $\pm$ 0.000 & 0.384 $\pm$ 0.056 & 10.4 $\pm$ 3.0 \\
SatCLIP & 0.005 $\pm$ 0.001 & 0.322 $\pm$ 0.074 & 9.8 $\pm$ 3.2 & 0.004 $\pm$ 0.001 & 0.433 $\pm$ 0.088 & 10.0 $\pm$ 3.1 \\
Climplicit & 0.001 $\pm$ 0.000 & 0.282 $\pm$ 0.081 & 10.2 $\pm$ 3.3 & 0.001 $\pm$ 0.000 & 0.341 $\pm$ 0.083 & 11.8 $\pm$ 4.0 \\
CSP-fMoW & 0.005 $\pm$ 0.001 & 0.407 $\pm$ 0.183 & 13.9 $\pm$ 4.1 & 0.006 $\pm$ 0.001 & 0.290 $\pm$ 0.116 & 13.4 $\pm$ 5.8 \\
\bottomrule
\end{tabular}
}}

\label{tab:splice_quant_small}
\end{table}

\begin{figure}[thbp]
    \centering
    \setlength{\tabcolsep}{2pt}
    \renewcommand{\arraystretch}{0.95}
    \newcommand{\spliceimg}[1]{\raisebox{-0.5\height}{\includegraphics[width=0.22\linewidth]{#1}}}
    \newcommand{\spliceimgbig}[1]{\raisebox{-0.5\height}{\includegraphics[width=0.25\linewidth]{#1}}}
    \newcommand{\spliceimgxbig}[1]{\raisebox{-0.5\height}{\includegraphics[width=0.275\linewidth]{#1}}}
    \newcommand{\spliceimgsmall}[1]{\raisebox{-0.5\height}{\includegraphics[width=0.20\linewidth]{#1}}}
    \newcommand{\splicetag}[1]{
        \refstepcounter{subfigure}\label{fig:splice_#1}
        \small(\alph{subfigure})
    }
    \setcounter{subfigure}{0}
    \begin{tabular}{@{}l@{\hspace{3pt}}c@{\hspace{-3mm}}cc@{\hspace{-3mm}}c@{}}
        & \small Paris & \, \, \, \small Sahara & \, \, \, \small Siberia & \, \, \, \small Bali \\
        \raisebox{-0.5cm}{\rotatebox{90}{\small \textbf{GeoCLIP}}}  &
            \spliceimgbig{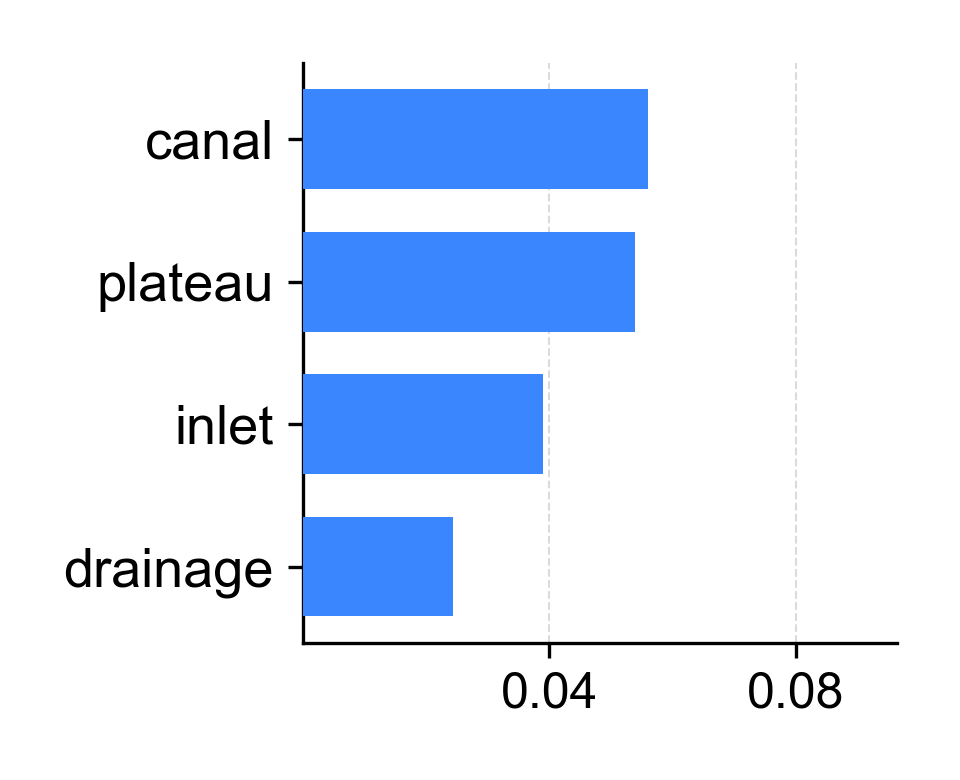} &
            \spliceimgbig{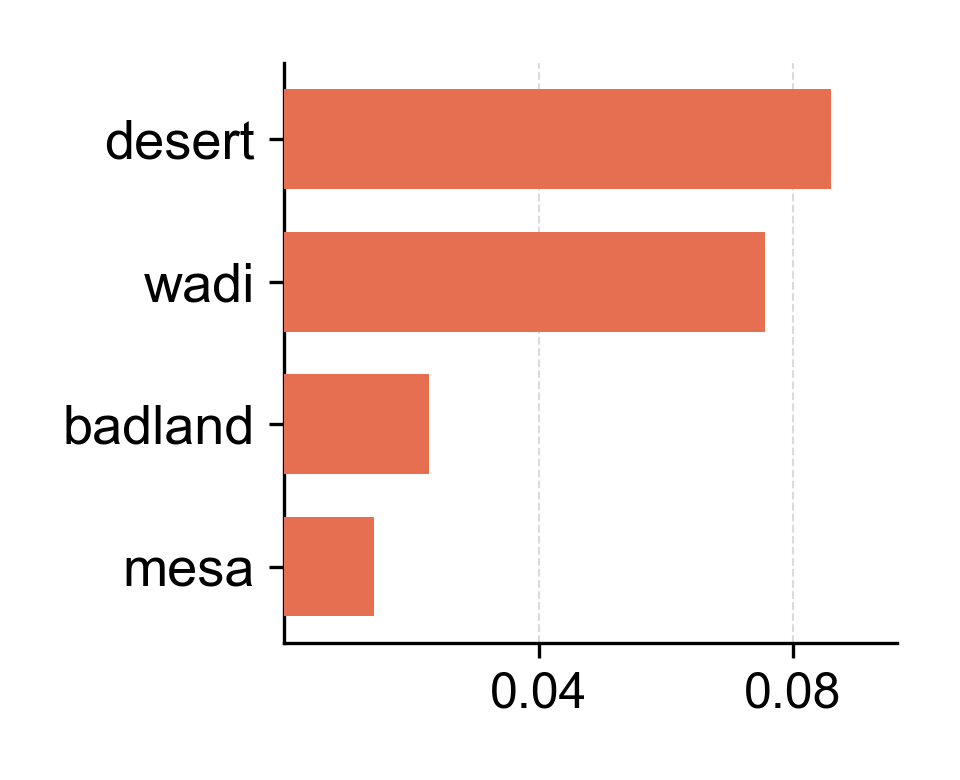} &
            \spliceimgbig{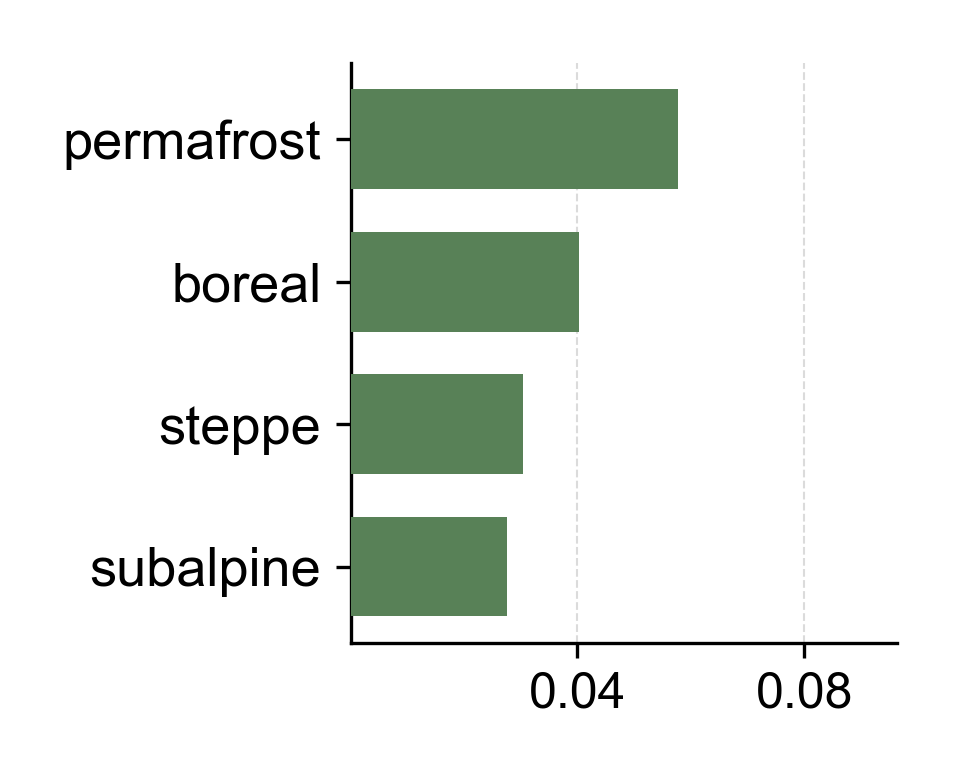} &
            \spliceimgbig{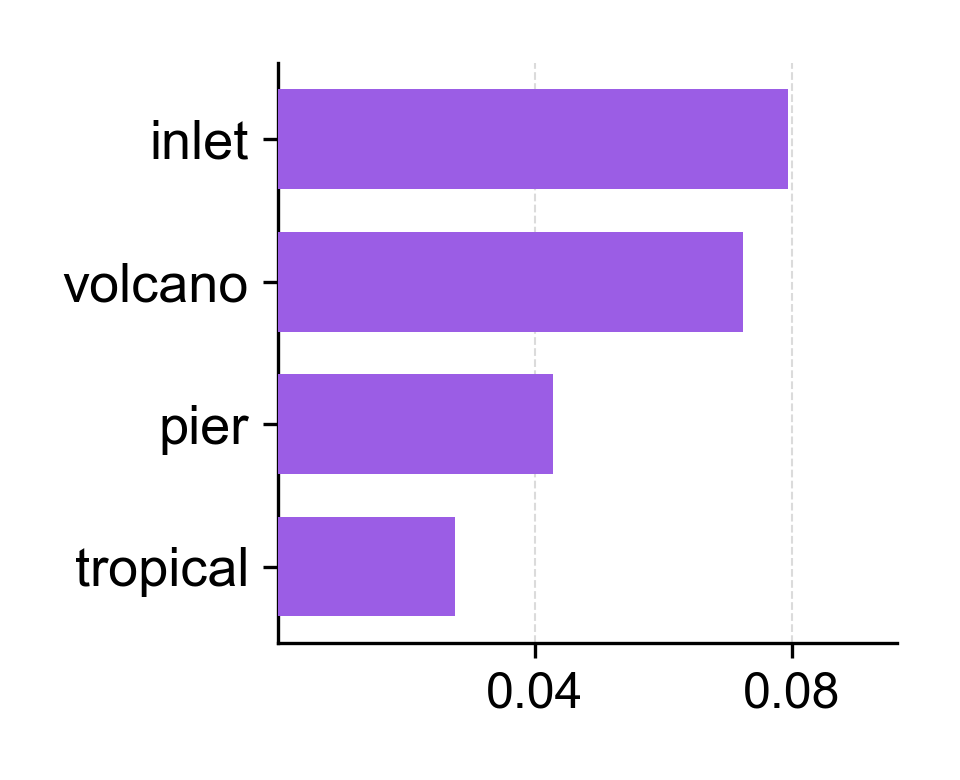} \\
       \raisebox{-0.5cm}{\rotatebox{90}{\small \textbf{SatCLIP}}} &
            \spliceimgxbig{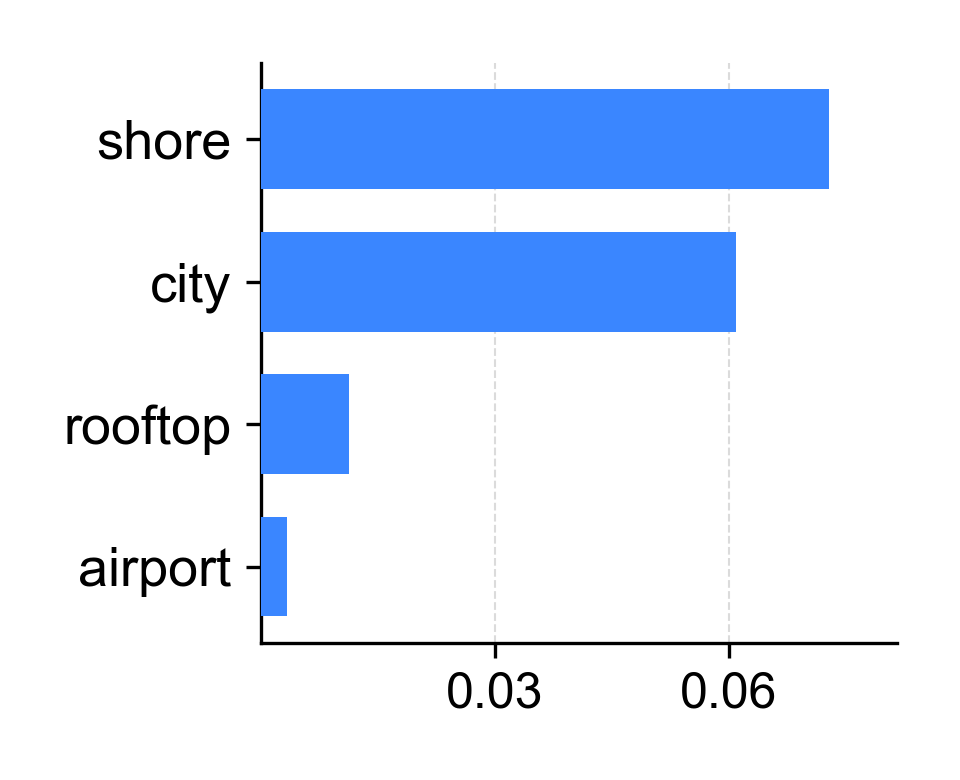} &
            \spliceimgxbig{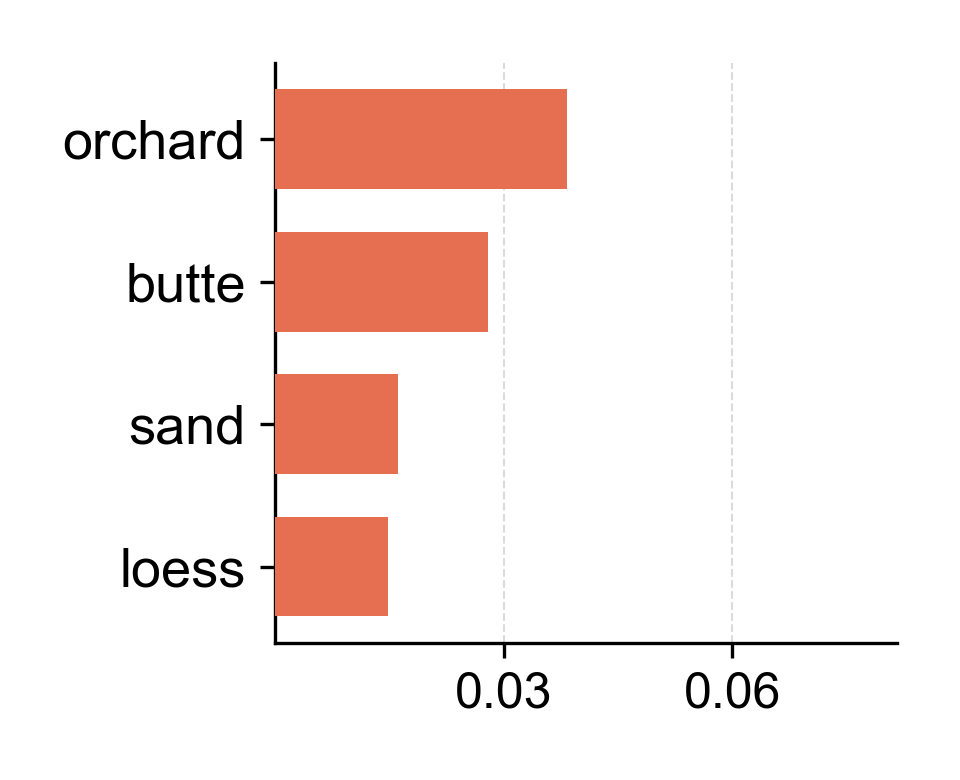} &
            \spliceimgxbig{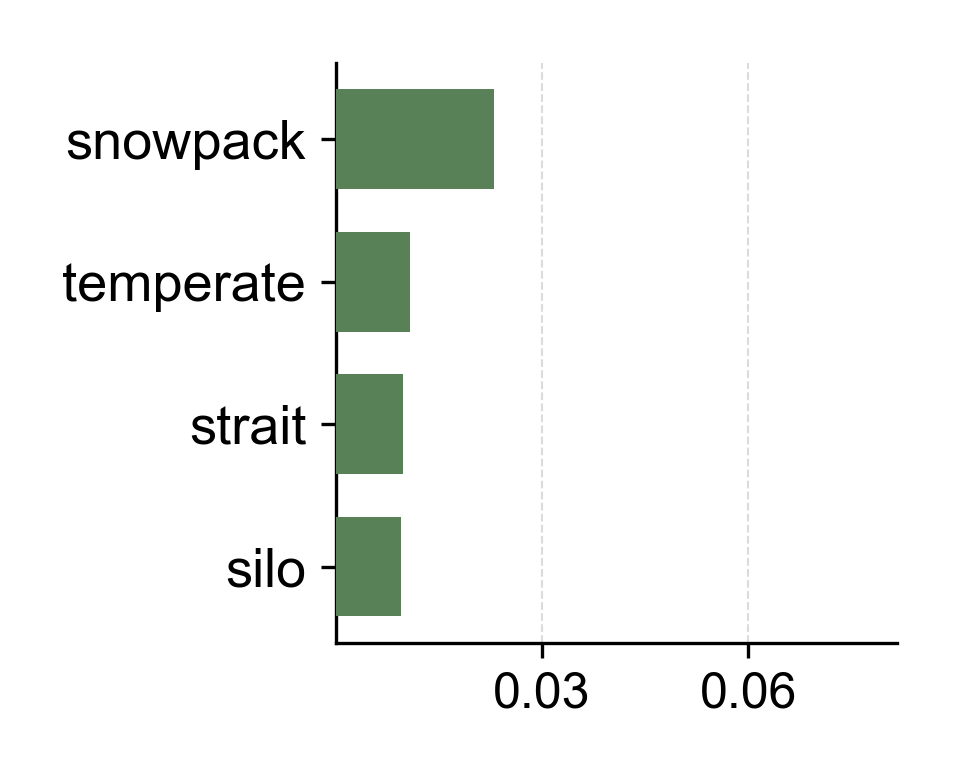} &
            \spliceimgxbig{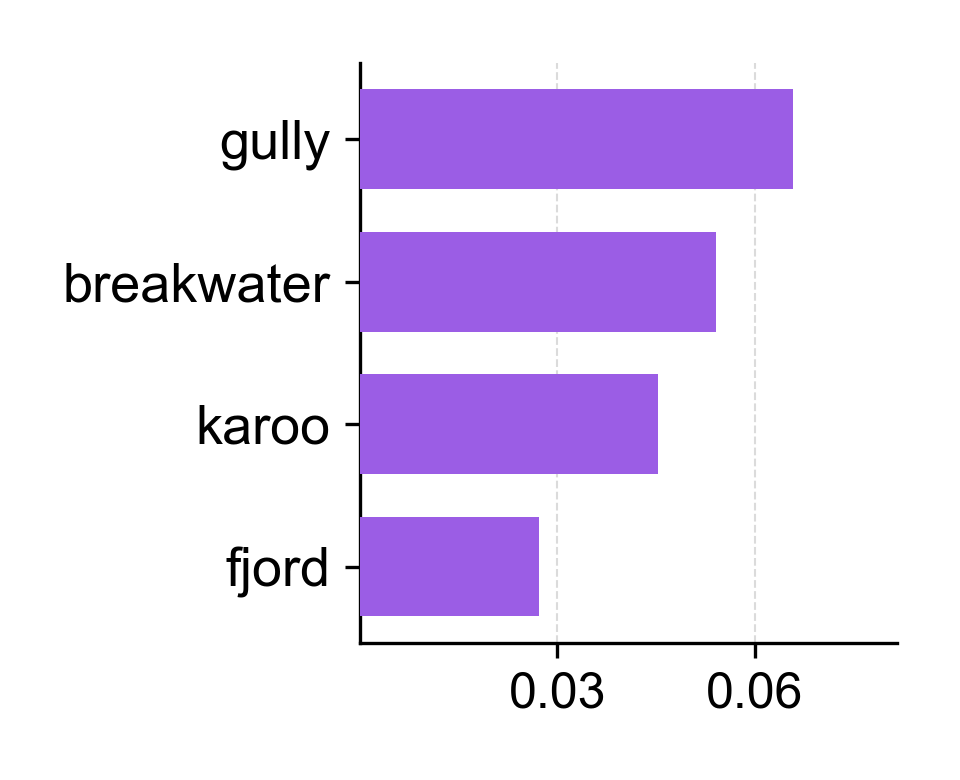}\\
        \raisebox{-0.5cm}{\rotatebox{90}{\small \textbf{Climplicit}}}  &
            \spliceimgbig{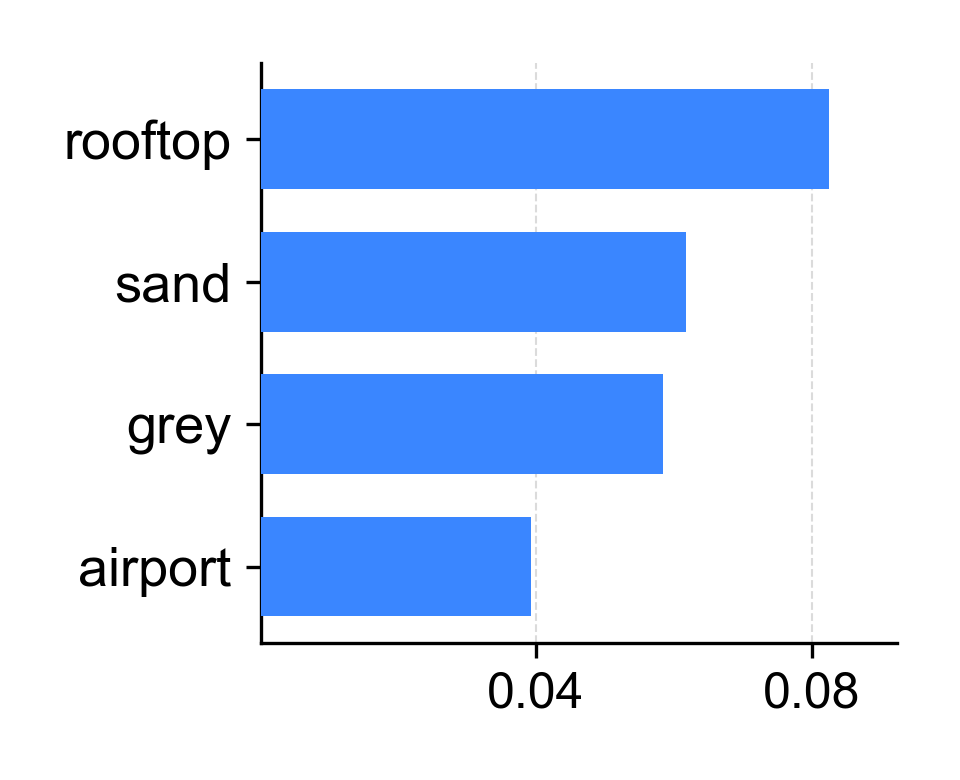} &
            \spliceimgbig{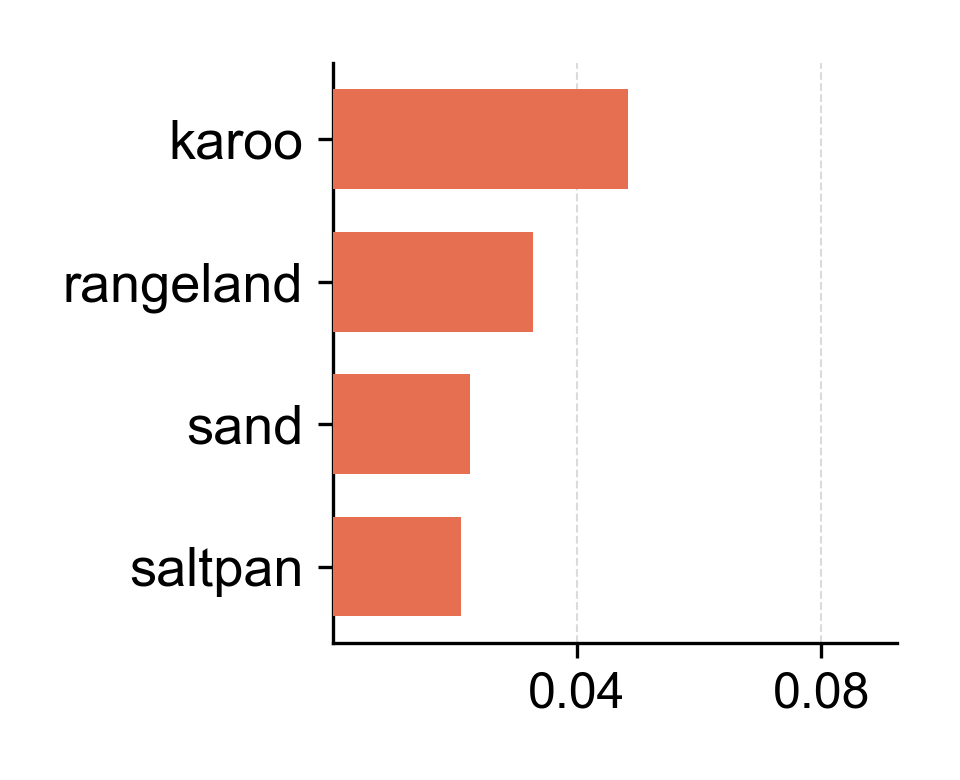} &
            \spliceimgbig{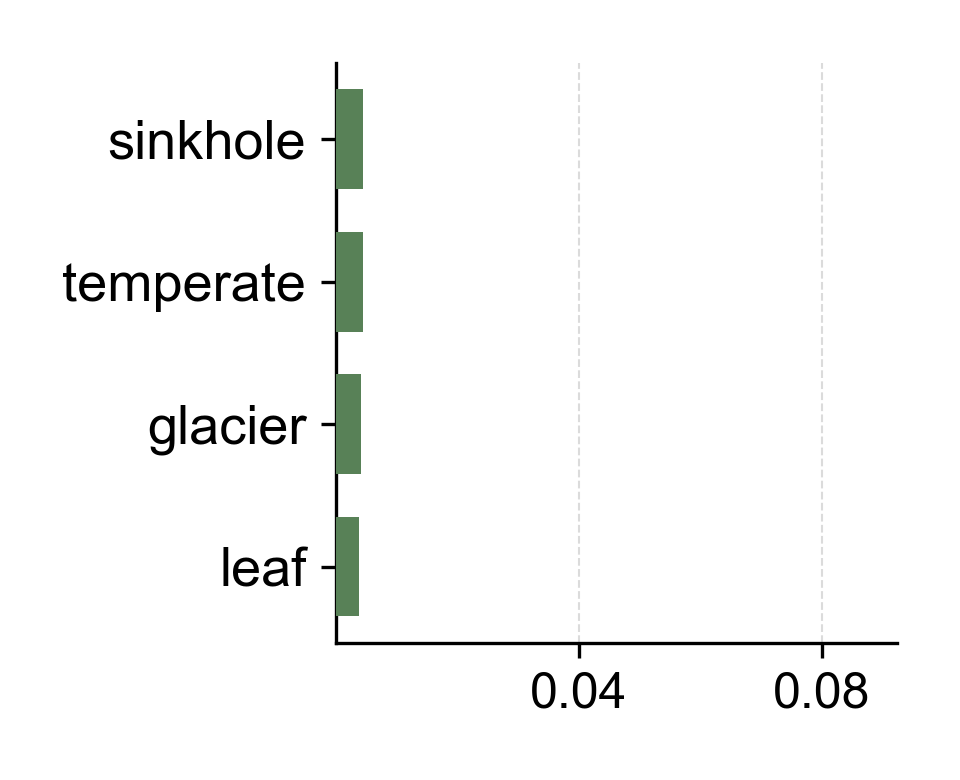} &
            \spliceimgbig{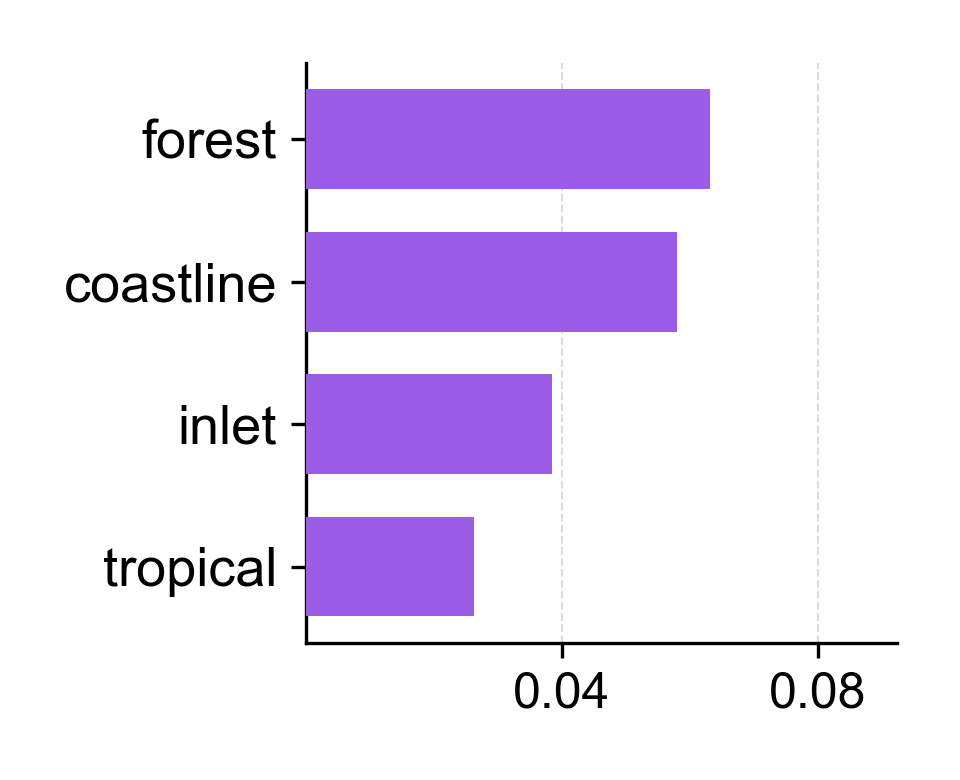} \\
        \raisebox{-0.5cm}{\rotatebox{90}{\small \textbf{CSP-fMoW}}}  &
            \spliceimgxbig{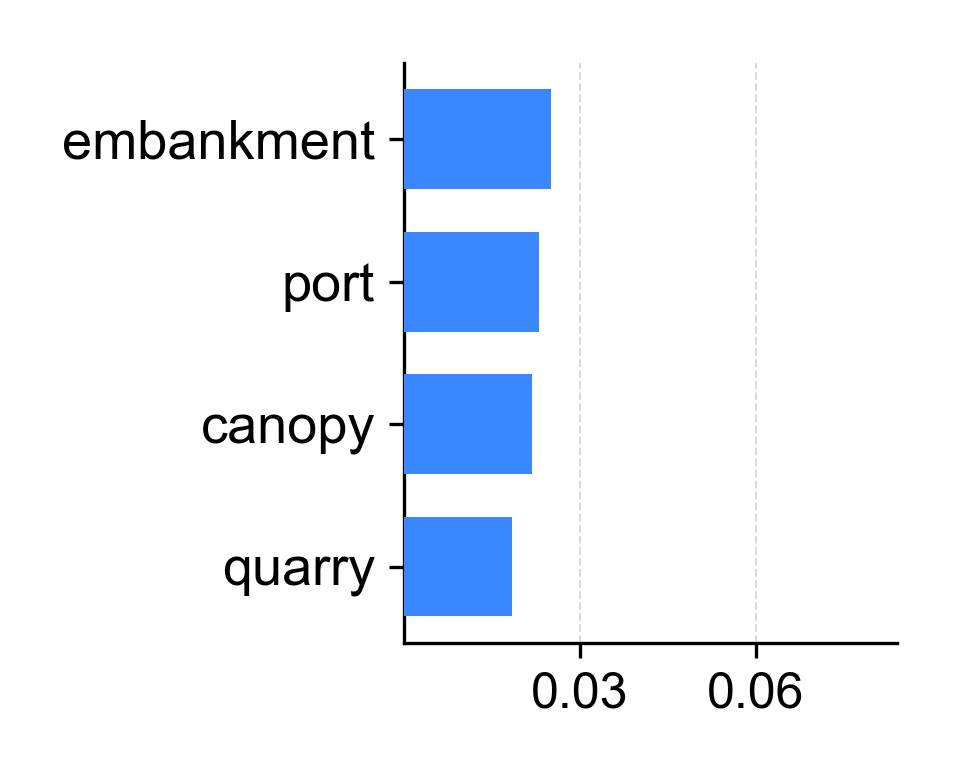} &
            \spliceimgbig{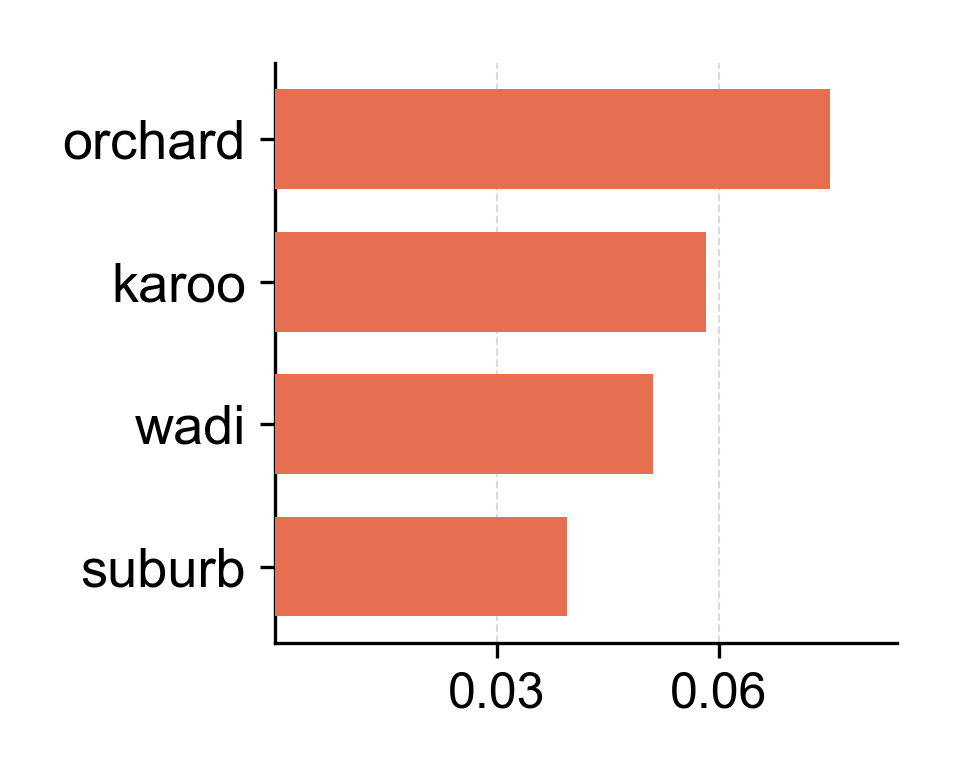} &
            \spliceimgbig{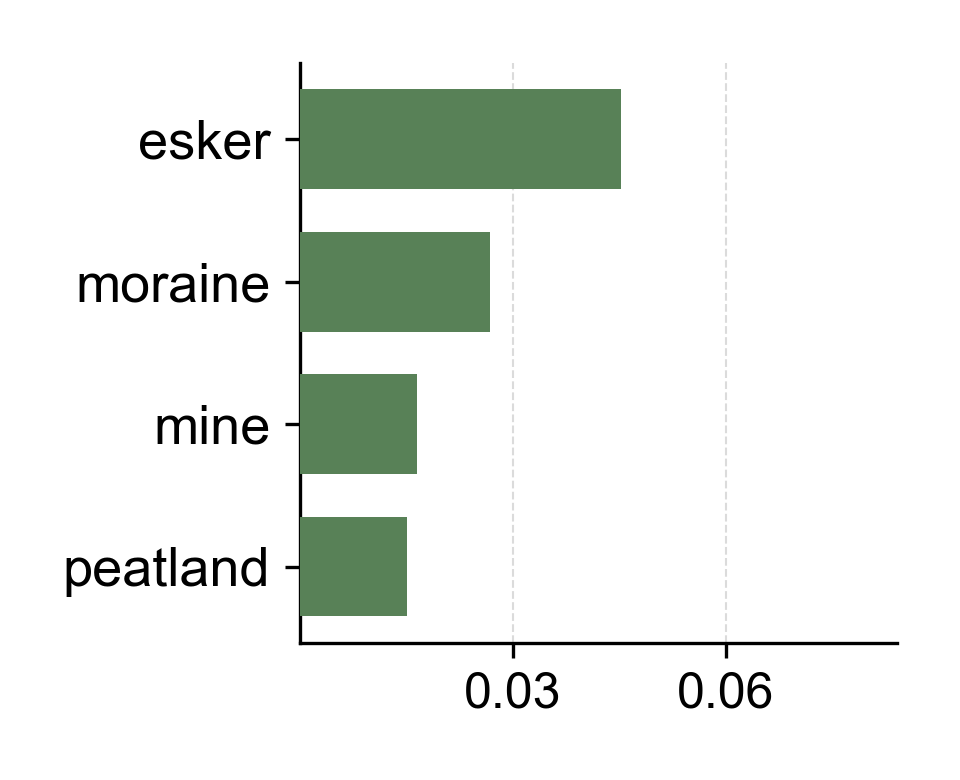} &
            \spliceimgbig{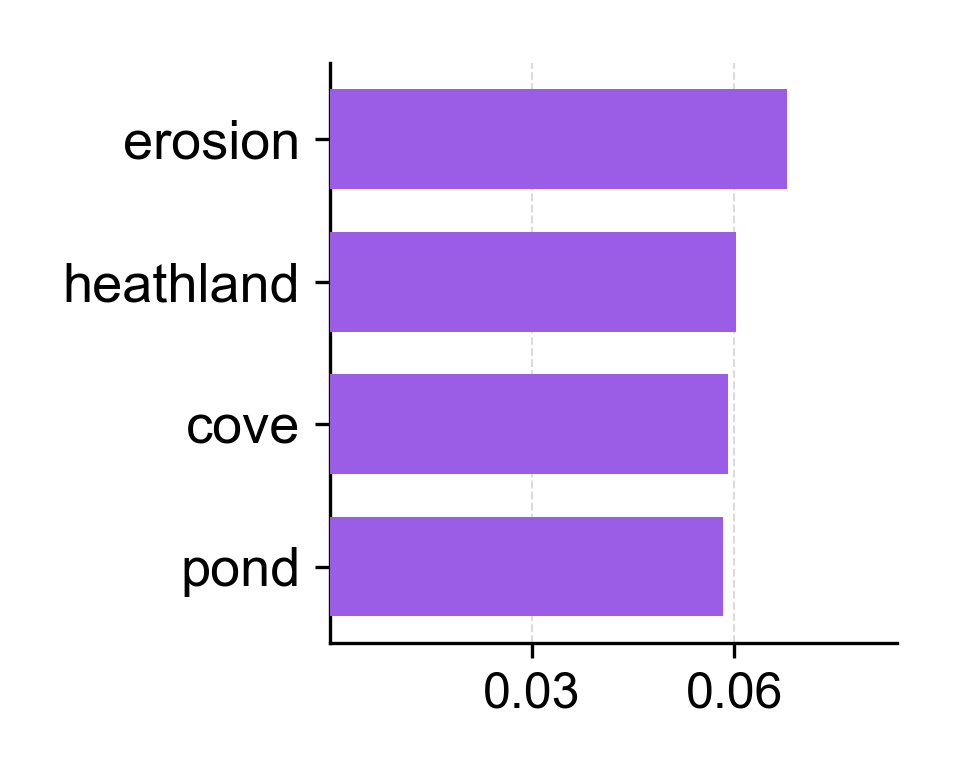} \\
    \end{tabular}

    \caption{\textbf{SpLiCE results on a smaller geospatial concept set, containing roughly 200 concepts,} on GeoCLIP, SatCLIP, Climplicit, and CSP-fMoW Using a smaller geospatial concept set biases the natural language decompositions to contain solely the geographic information, with tradeoffs for how much fine-grained information can be extracted.}
    \label{fig:splice_decompositions_small}
\end{figure}

Next, we provide additional SpLiCE results on a small geospatial concept set consisting of 200 geospatial concepts derived from Git-10M concepts. \Cref{tab:splice_quant_small} shows that even with a small geospatial concept set, MSE and average cosine similarity show that sparse embeddings can still be used to reconstruct the original embeddings. A smaller geospatial concept, however, introduces more inductive bias by further limiting the potential concepts that can be revealed by SpLiCE. However, we do see that the SpLiCE decompositions in \Cref{fig:splice_decompositions_small} align better with geographic attributes. Interestingly, between location encoders, different geographic text concepts are expressed for the same location. \edited{For example, the GeoCLIP location embeddings in the Sahara decompose with concepts such as ``desert'' and ``wadi'', while SatCLIP decomposes with ``orchard'' and ``butte''.}

\subsubsection{CLIP Surgery Saliency Maps}\label{apx: sec: clipsurgery}

\begin{figure}[htbp]
  \centering

  \begin{subfigure}[b]{0.45\textwidth}
    \includegraphics[width=0.49\linewidth]{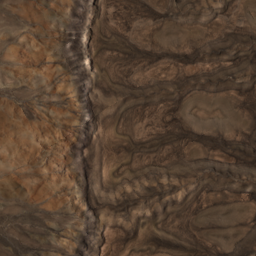}\hfill
    \includegraphics[width=0.49\linewidth]{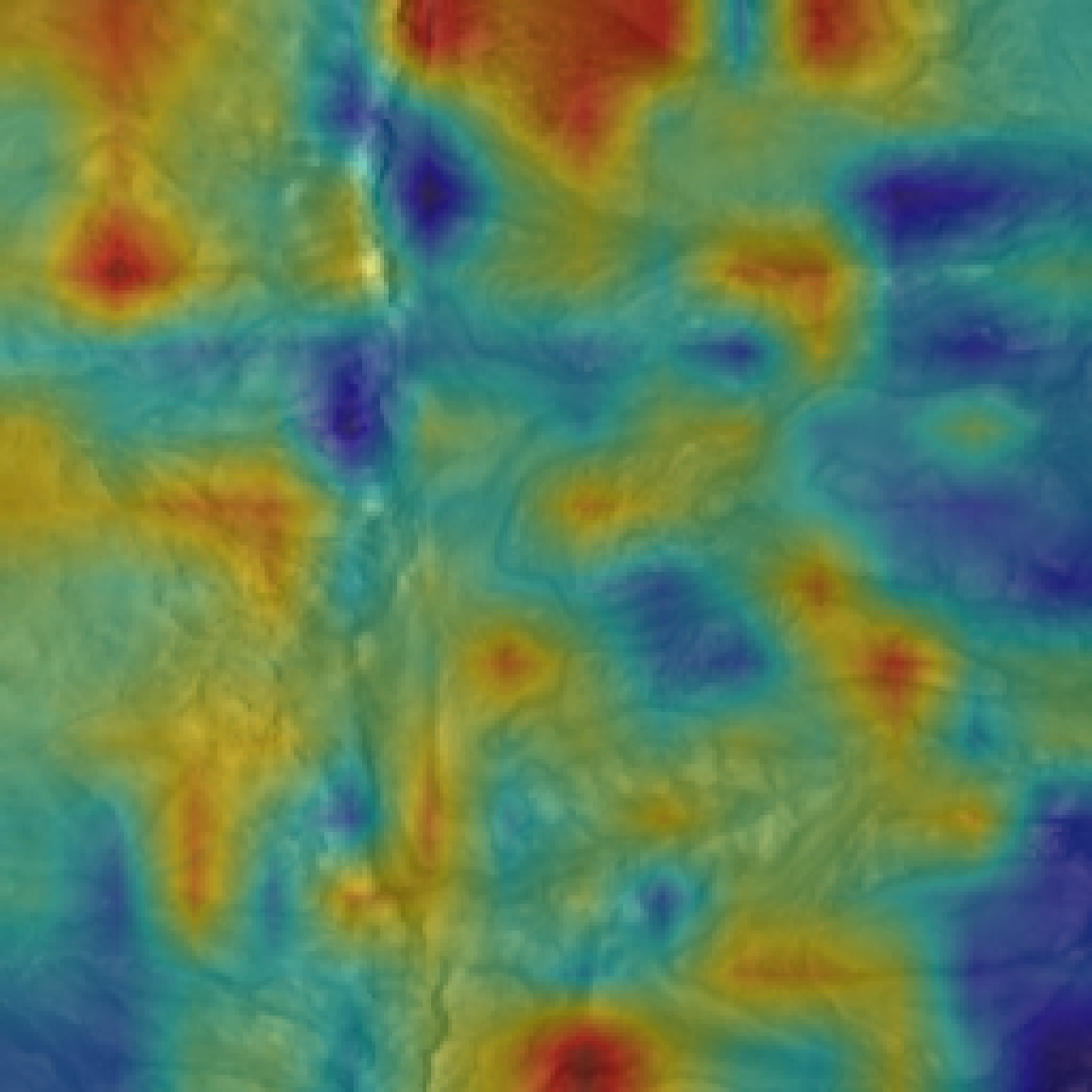}
    \caption{Africa}
    \label{fig:africa-saliency}
  \end{subfigure}
  \hfill
  \begin{subfigure}[b]{0.45\textwidth}
    \includegraphics[width=0.49\linewidth]{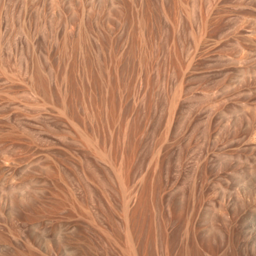}\hfill
    \includegraphics[width=0.49\linewidth]{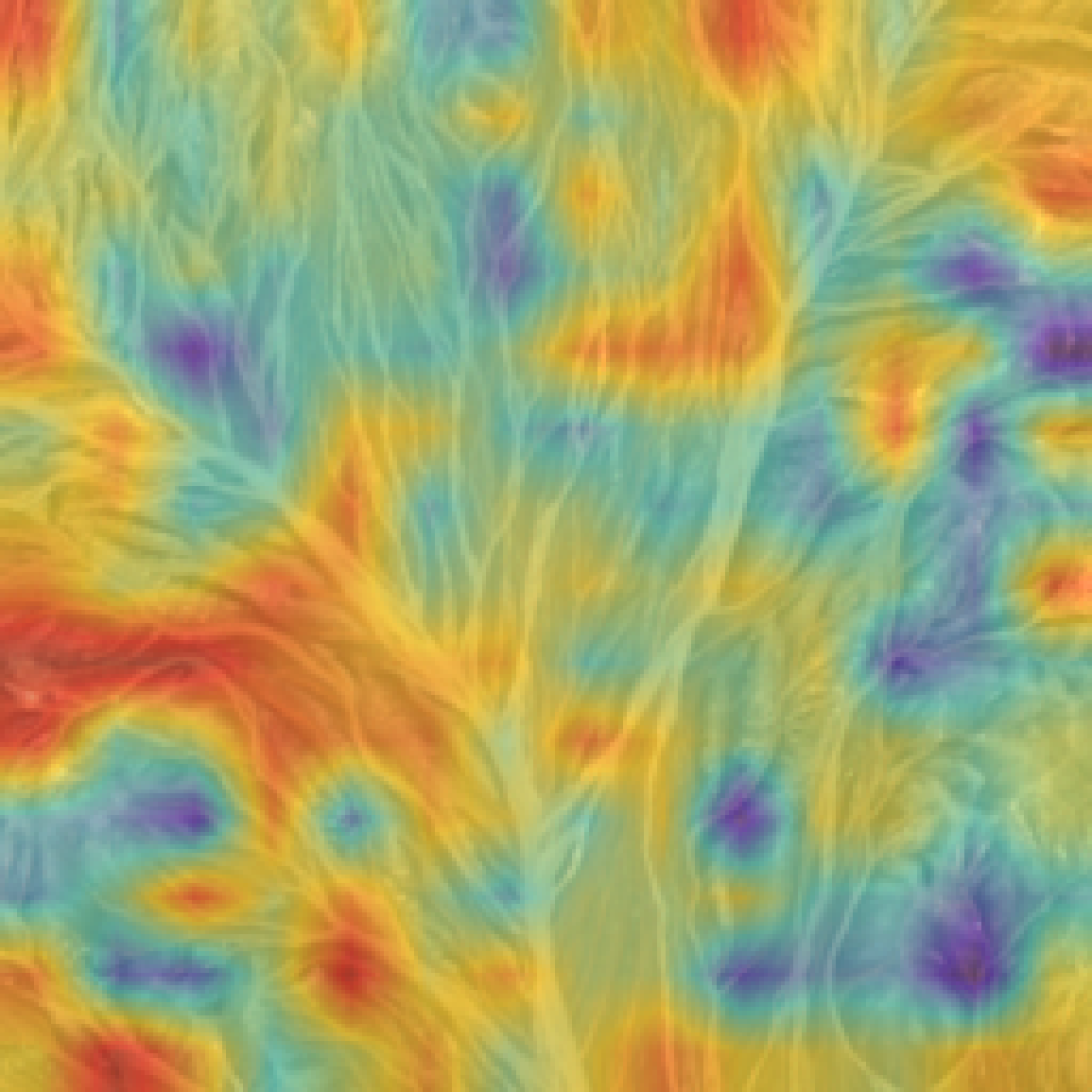}
    \caption{Atacama}
    \label{fig:atacama-saliency}
  \end{subfigure}

  \vspace{0.8em}


  \begin{subfigure}[b]{0.45\textwidth}
    \includegraphics[width=0.49\linewidth]{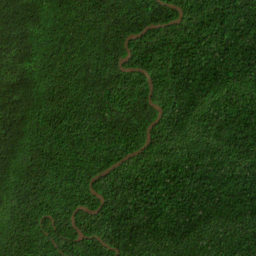}\hfill
    \includegraphics[width=0.49\linewidth]{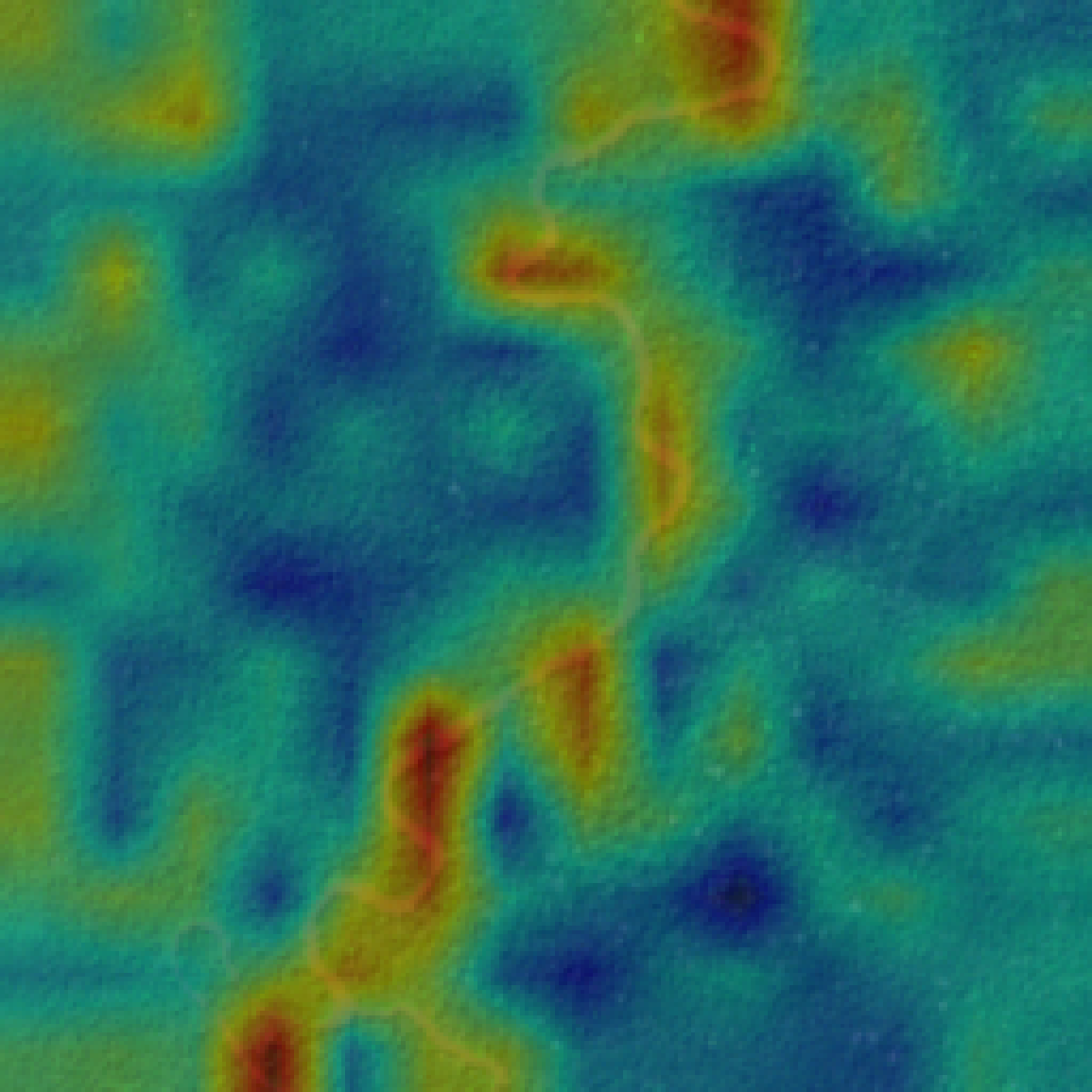}
    \caption{Congo}
    \label{fig:congo-saliency}
  \end{subfigure}
  \hfill
  \begin{subfigure}[b]{0.45\textwidth}
    \includegraphics[width=0.49\linewidth]{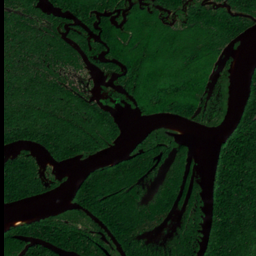}\hfill
    \includegraphics[width=0.49\linewidth]{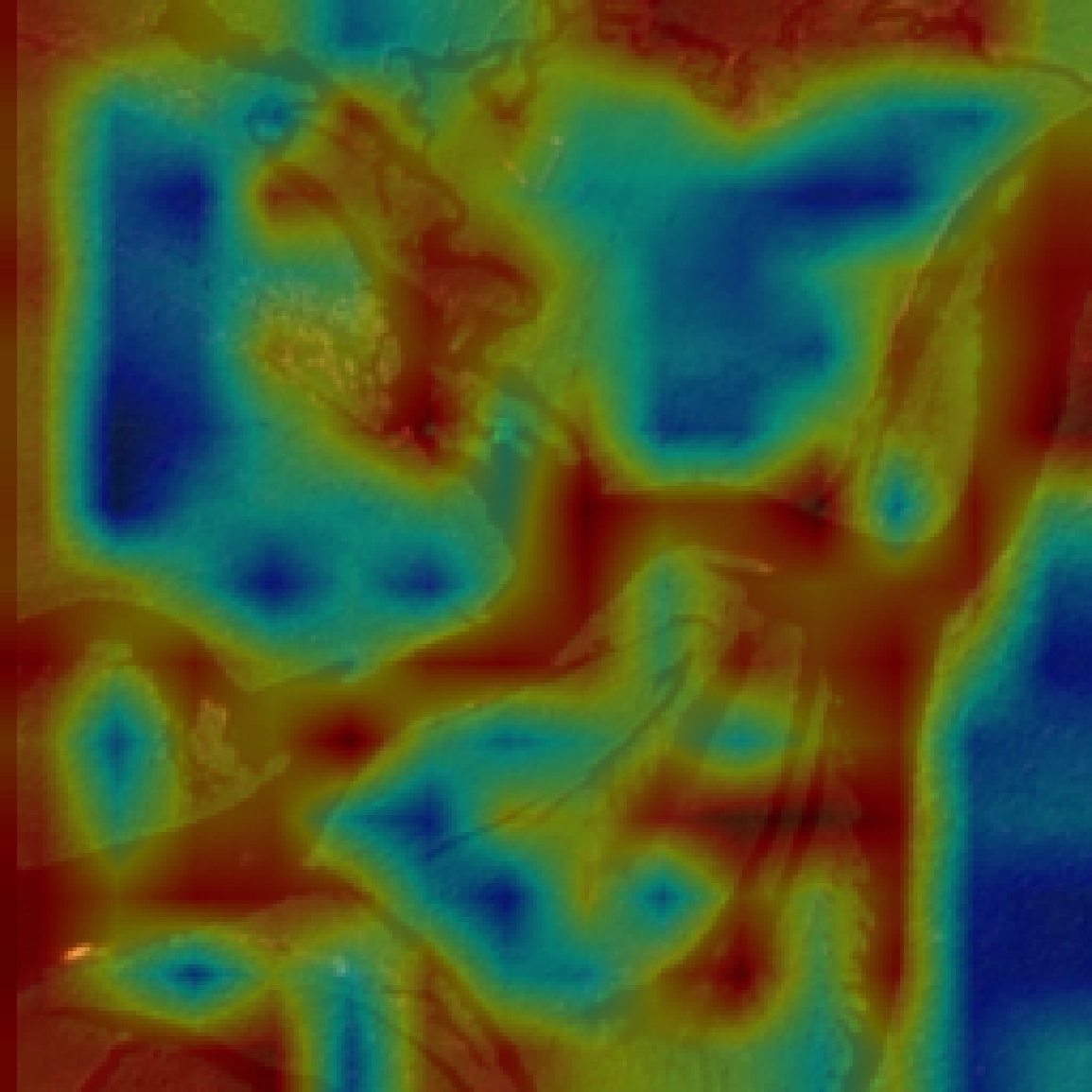}
    \caption{Amazon}
    \label{fig:amazon-saliency}
  \end{subfigure}

  \caption{\textbf{CLIP Surgery saliency maps generated from Sentinel-2 (S2) imagery using the SatCLIP location and image encoders}. The first row shows desert-like land cover from geographically distinct yet visually similar regions — the Atacama and African Deserts — while the second row shows tropical rainforest-like land cover from the Congo Basin and Amazon. Though attention to water edges and rivers emerges in rainforest areas, deserts are much more difficult to interpret. 
  This suggests that saliency maps are less reliable for satellite imagery over spatially homogeneous land cover.}
  \label{fig:landcover-saliency}
\end{figure}

\begin{figure}[htbp]
  \centering

  \begin{subfigure}[b]{0.45\textwidth}
    \includegraphics[width=0.49\linewidth]{sec/figures/bboxes-NP/ImageNet/Paris/saliency/1.png}\hfill
    \includegraphics[width=0.49\linewidth]{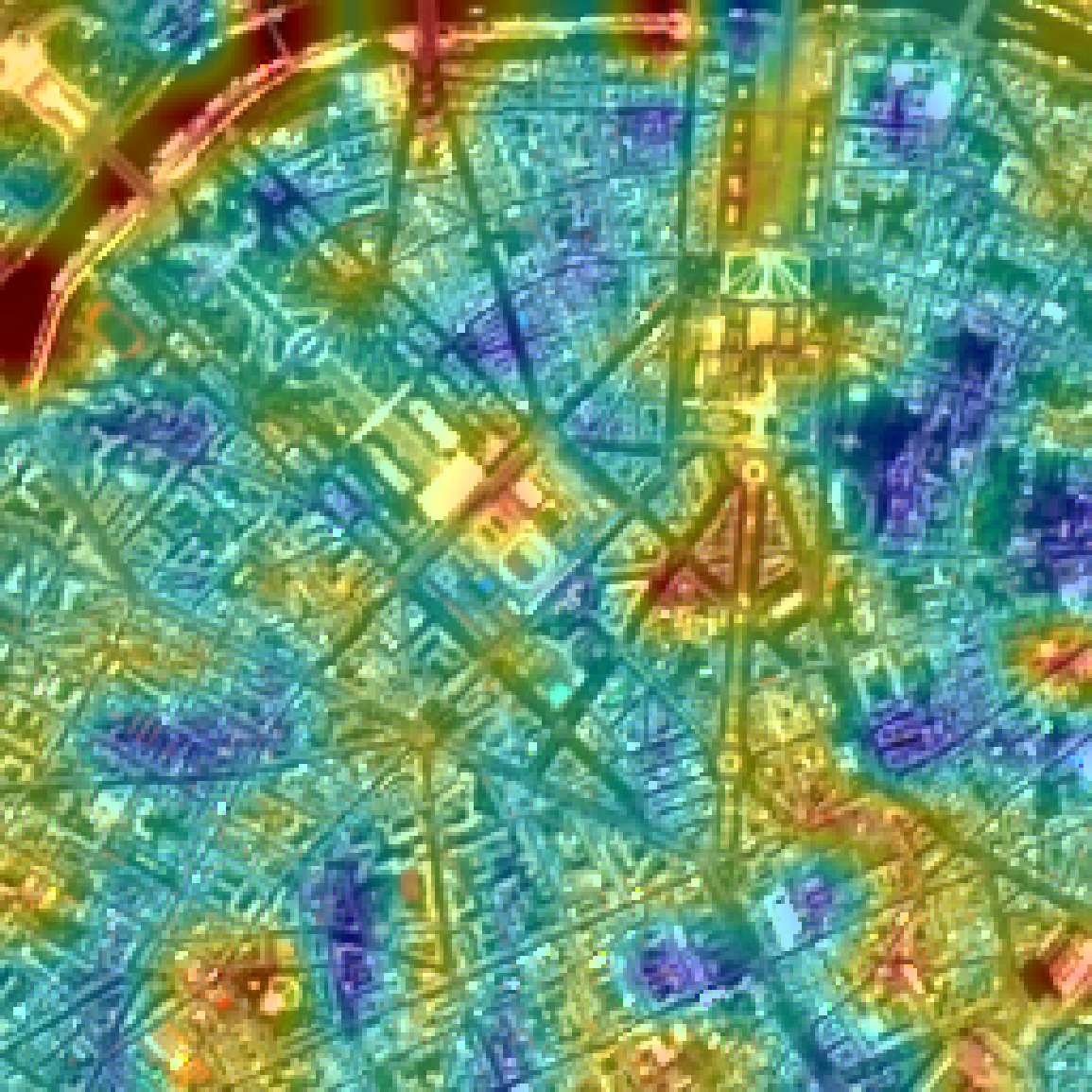}
    \caption{Paris}
    \label{fig:paris-saliency}
  \end{subfigure}
  \hfill
  \begin{subfigure}[b]{0.45\textwidth}
    \includegraphics[width=0.49\linewidth]{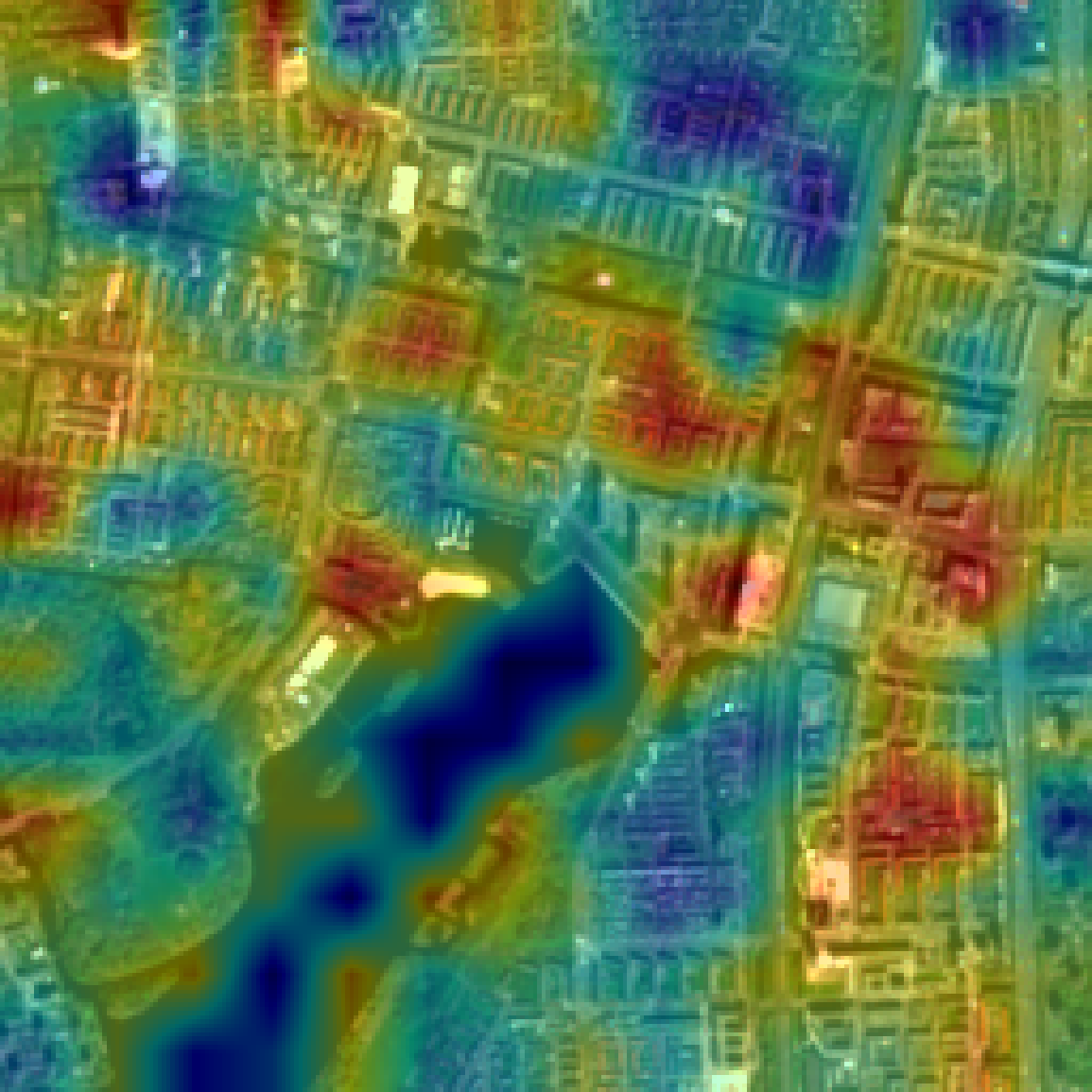}\hfill
    \includegraphics[width=0.49\linewidth]{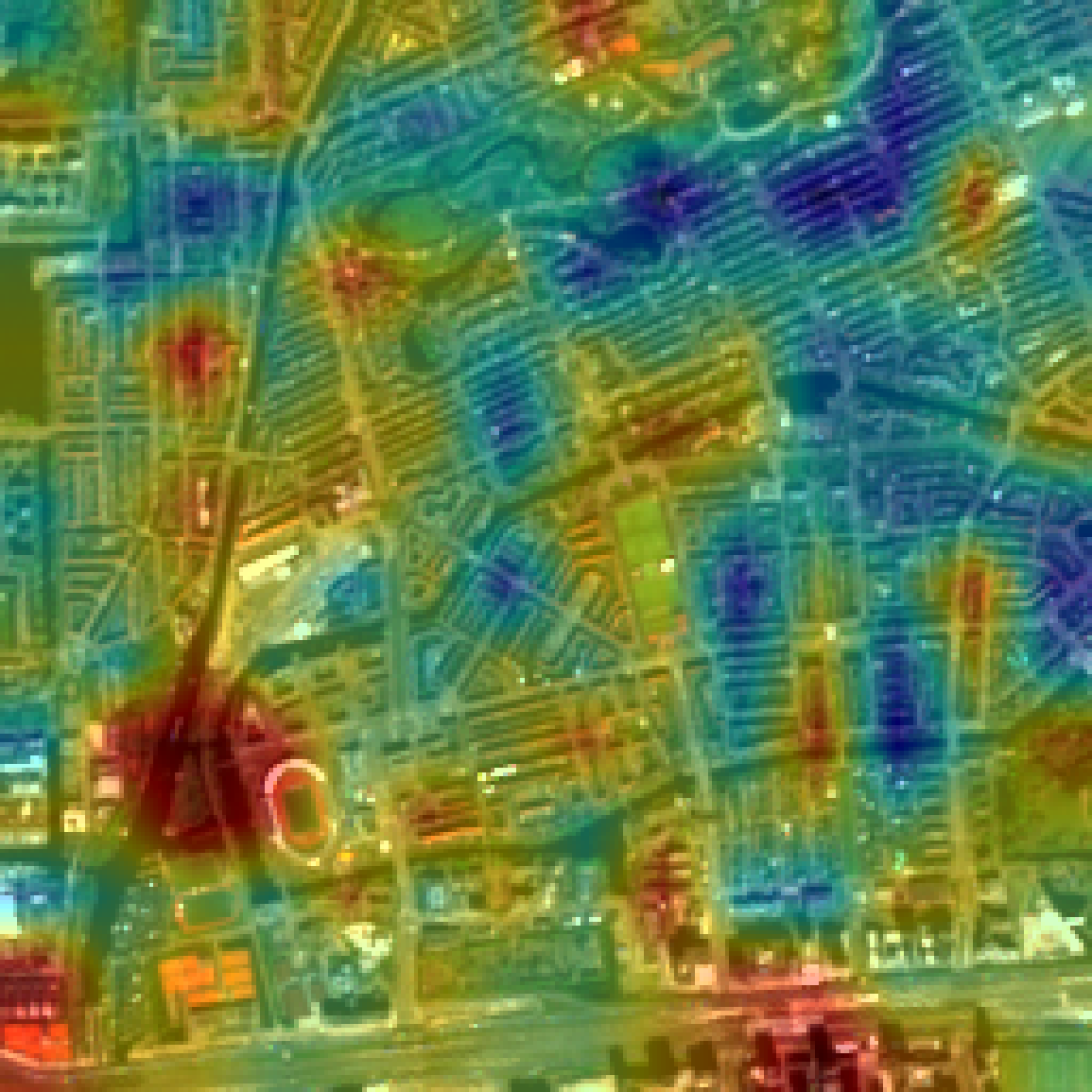}
    \caption{Amsterdam}
    \label{fig:ansterdam-saliency}
  \end{subfigure}

  \vspace{0.8em}


  \begin{subfigure}[b]{0.45\textwidth}
    \includegraphics[width=0.49\linewidth]{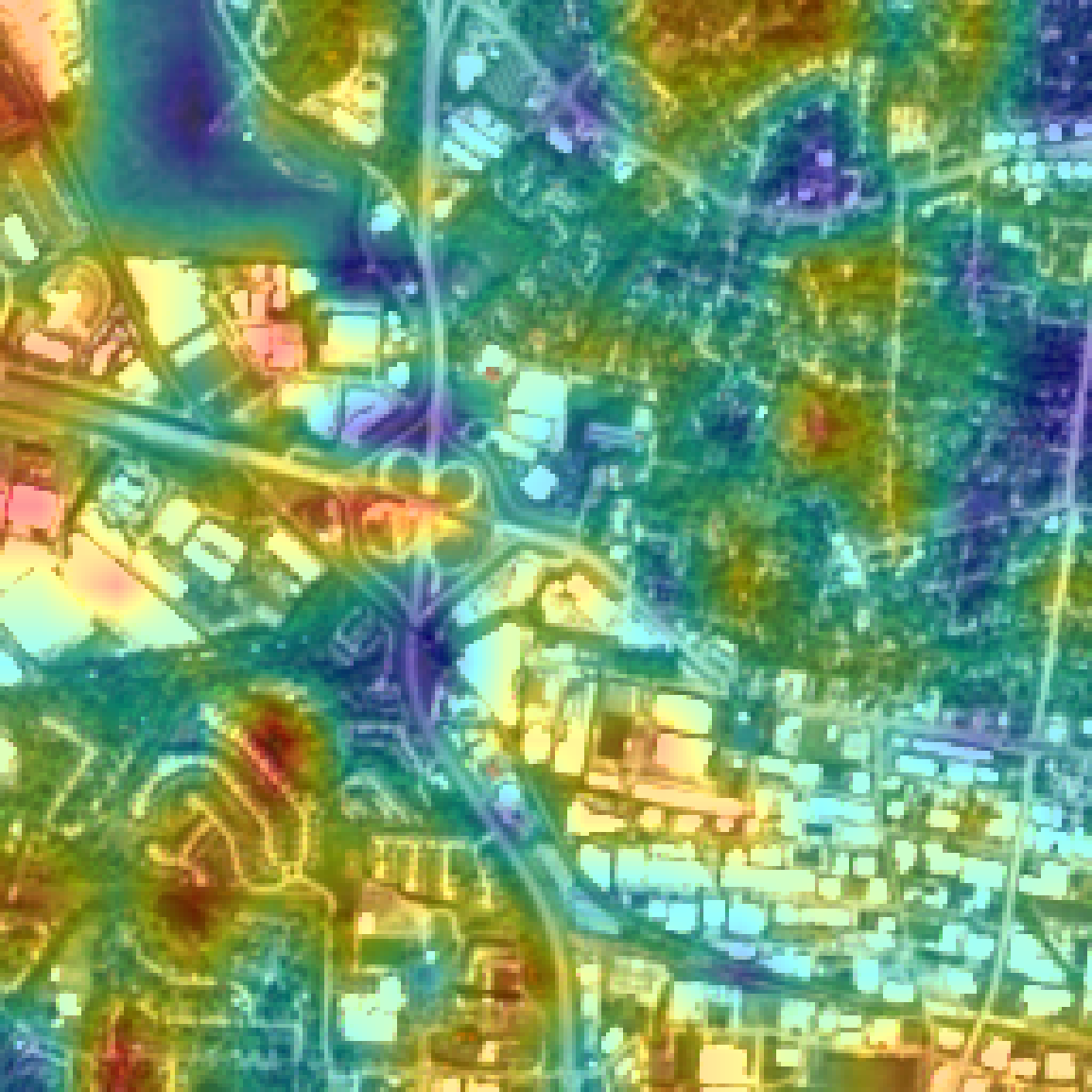}\hfill
    \includegraphics[width=0.49\linewidth]{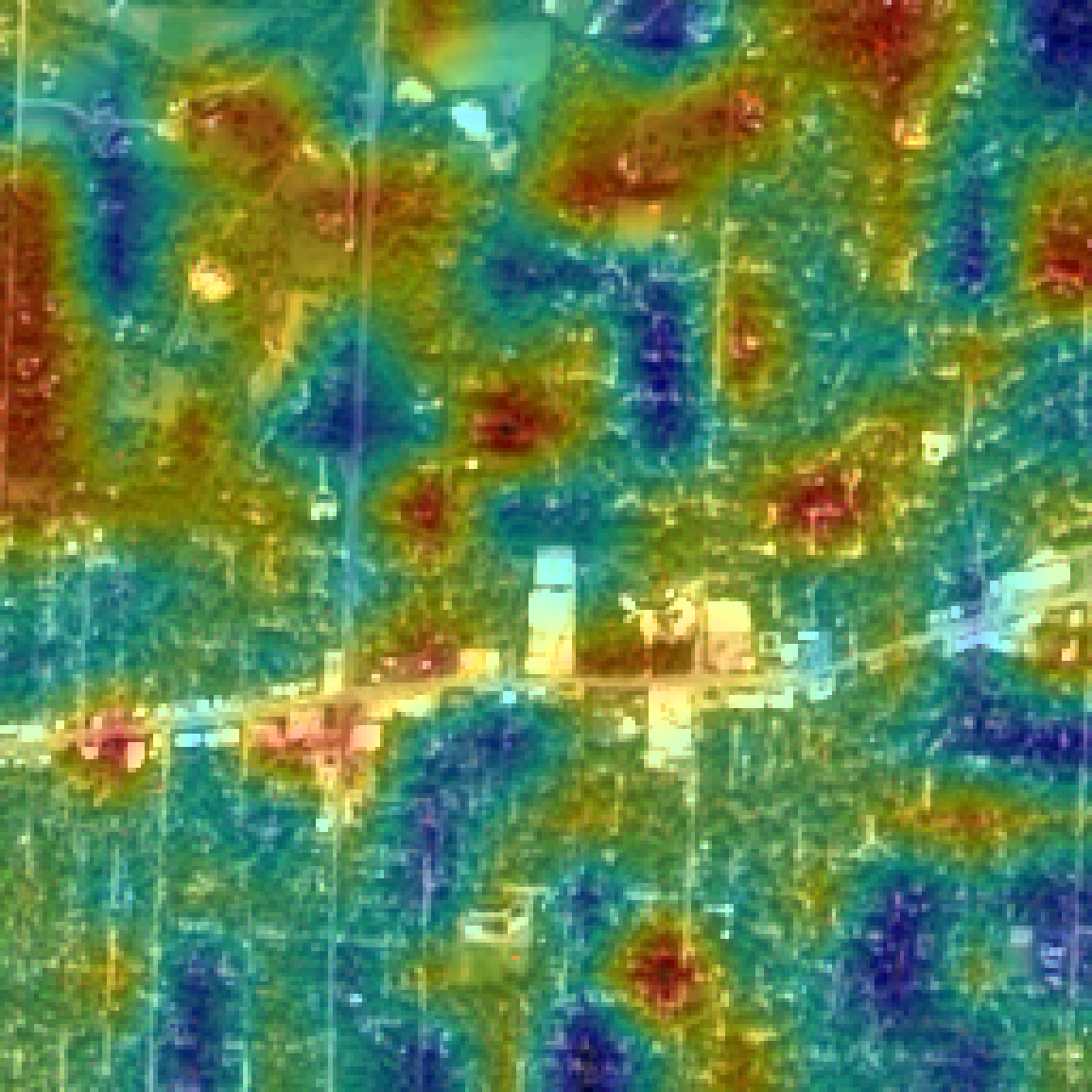}
    \caption{St. Louis}
    \label{fig:stlouis-saliency}
  \end{subfigure}
  \hfill
  \begin{subfigure}[b]{0.45\textwidth}
    \includegraphics[width=0.49\linewidth]{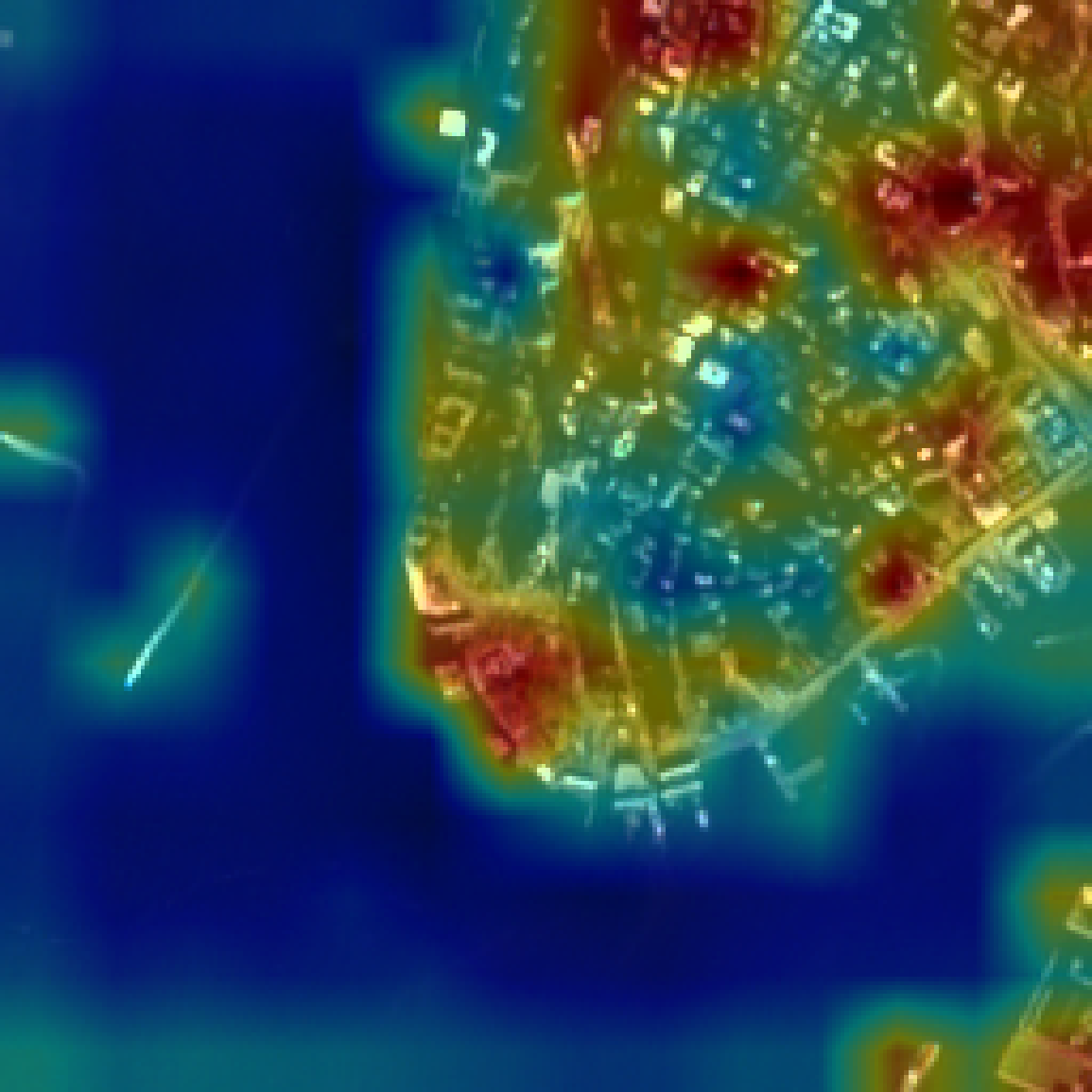}\hfill
    \includegraphics[width=0.49\linewidth]{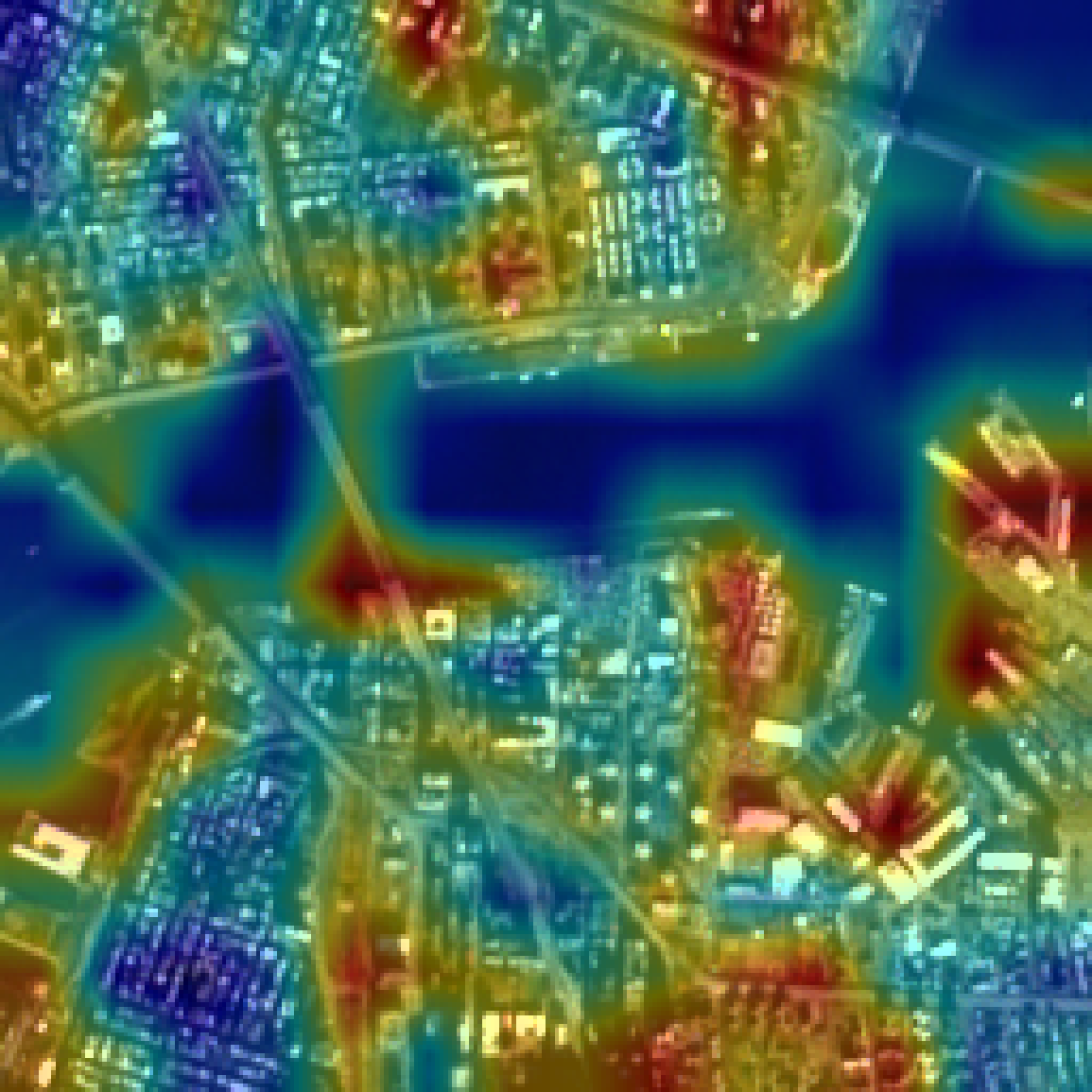}
    \caption{New York City}
    \label{fig:nyc-saliency}
  \end{subfigure}

  \caption{\textbf{Satellite image saliency maps for four cities: Paris, Amsterdam \edited{, St. Louis and New York City}}. While the model's attention tends to highlight structural boundaries such as roads and waterways, the maps are difficult to interpret directly due to their scale and resolution. We therefore introduce an additional step of cropping and clustering to extract more interpretable visual patterns.}
  \label{fig:city-saliency}
\end{figure}

\begin{figure}[tbh]
    \centering
    \setlength{\tabcolsep}{2pt}
    \begin{tabular}{@{}cccc@{}}
        \includegraphics[width=0.24\linewidth, height=0.15\textheight, keepaspectratio=false]{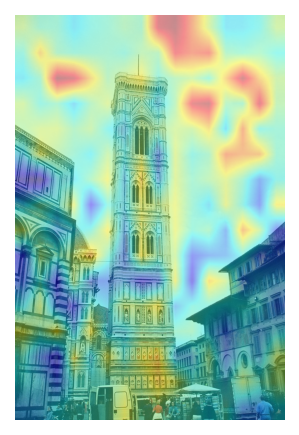} &
        \includegraphics[width=0.24\linewidth, height=0.15\textheight, keepaspectratio=false]{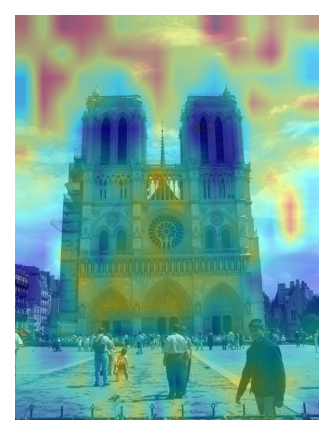} &
        \includegraphics[width=0.24\linewidth, height=0.15\textheight, keepaspectratio=false]{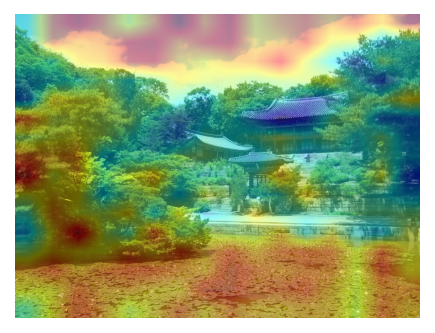} &
        \includegraphics[width=0.24\linewidth, height=0.15\textheight, keepaspectratio=false]{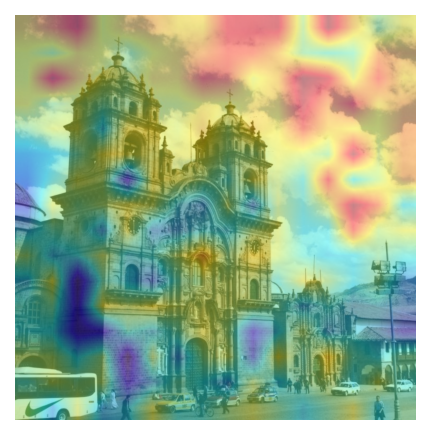} \\[2pt]
        \includegraphics[width=0.24\linewidth, height=0.15\textheight, keepaspectratio=false]{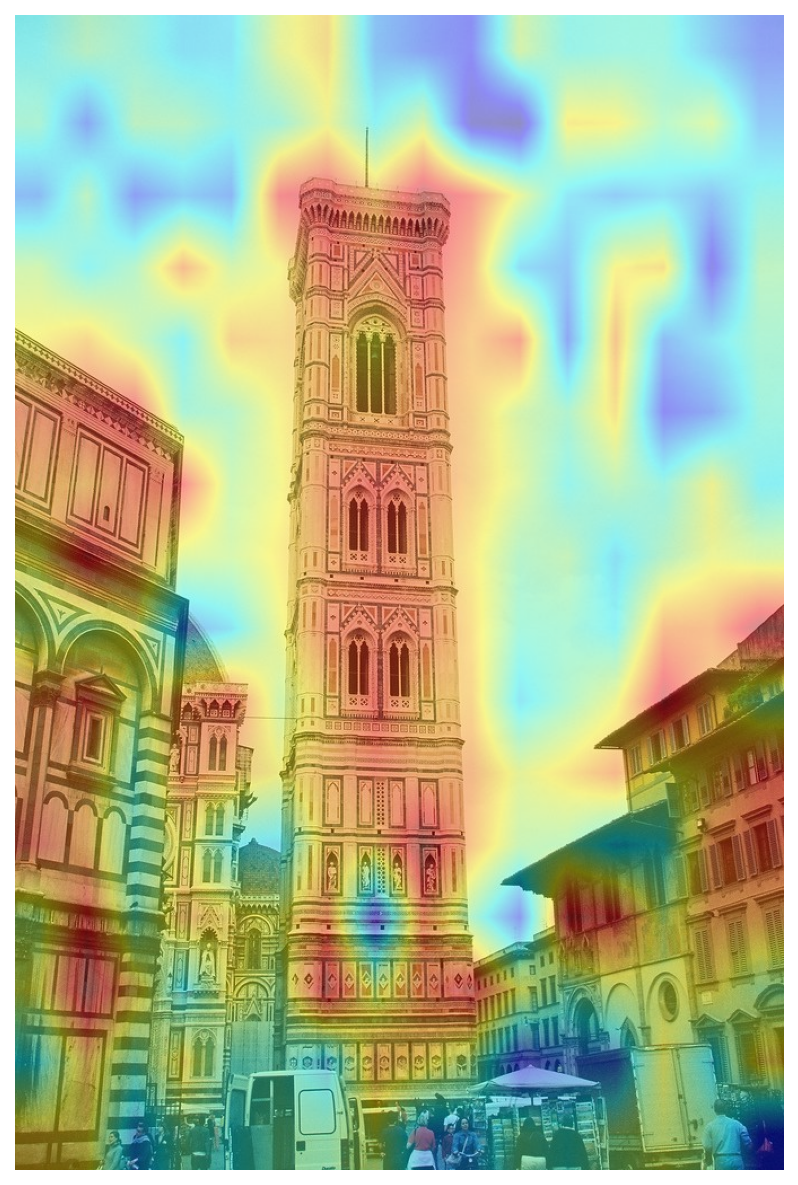} &
        \includegraphics[width=0.24\linewidth, height=0.15\textheight, keepaspectratio=false]{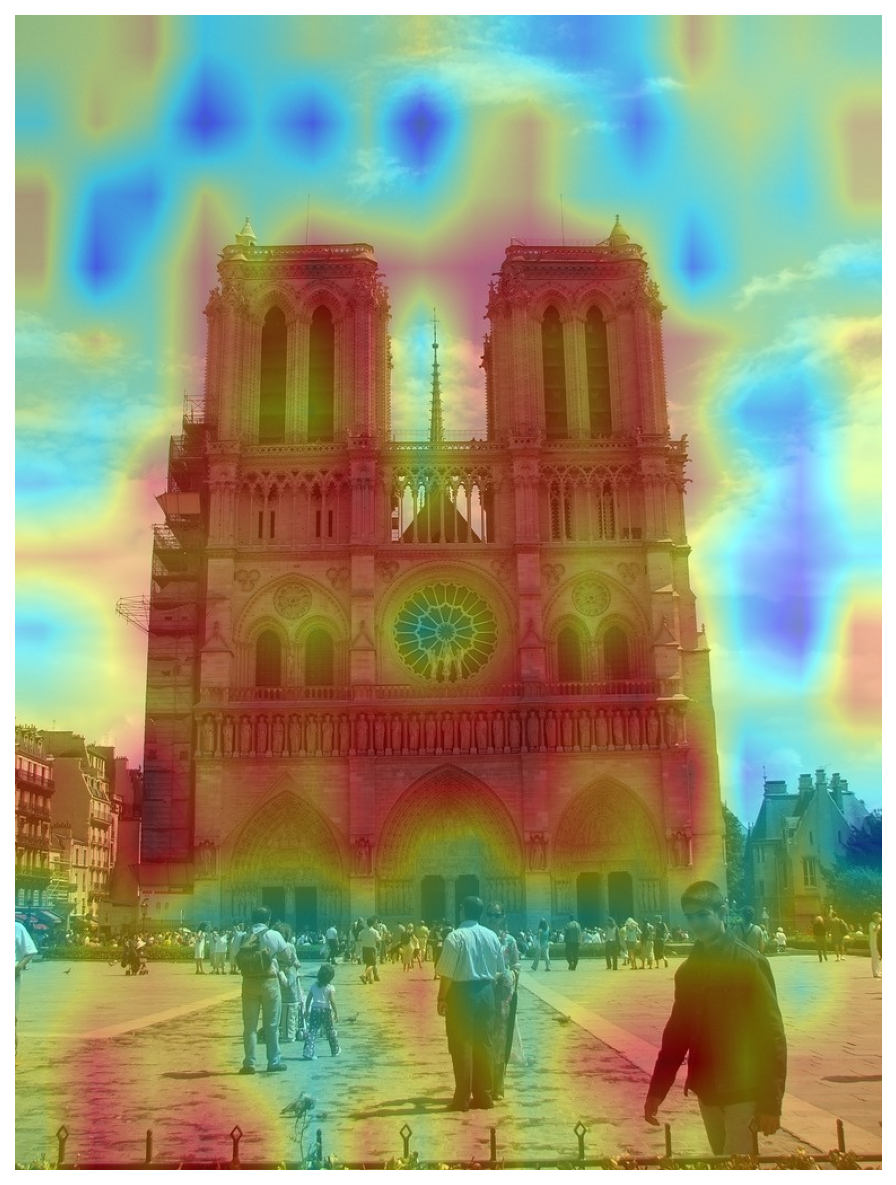} &
        \includegraphics[width=0.24\linewidth, height=0.15\textheight, keepaspectratio=false]{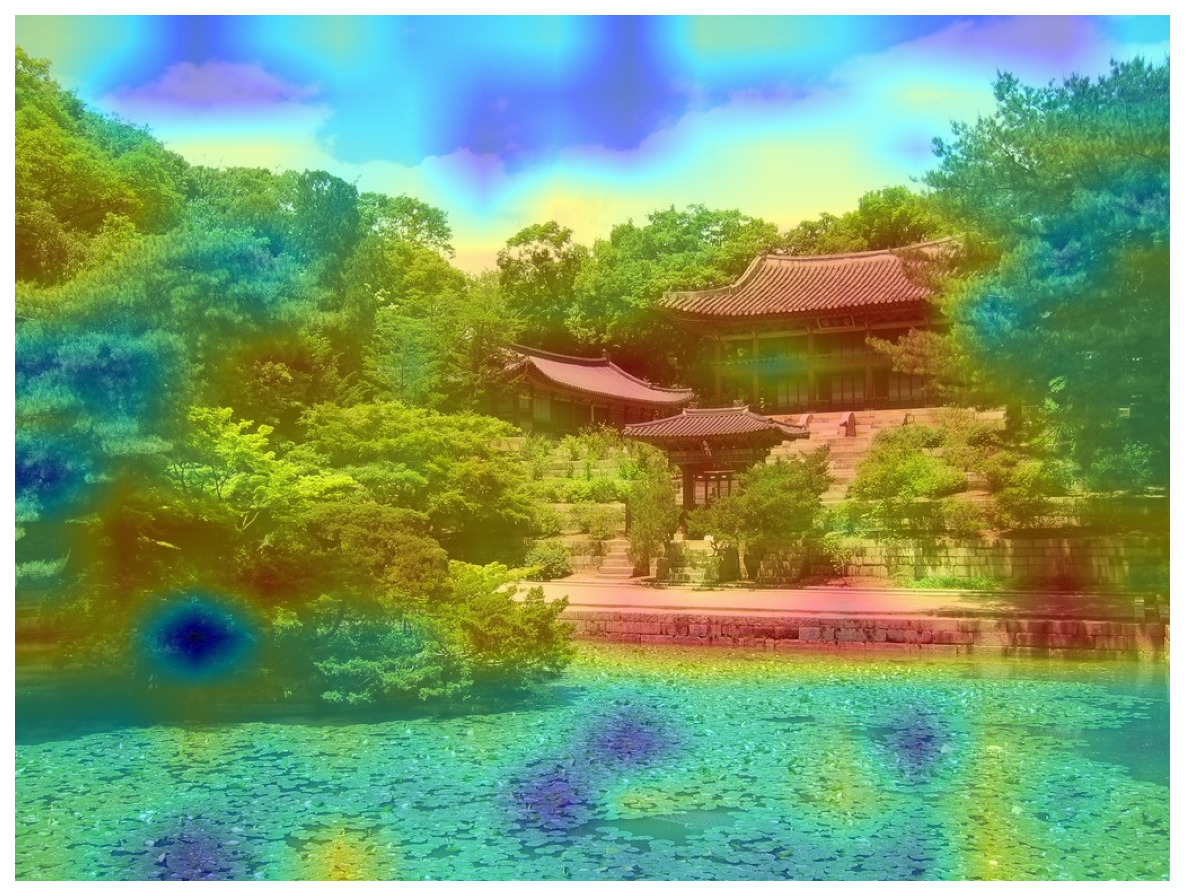} &
        \includegraphics[width=0.24\linewidth, height=0.15\textheight, keepaspectratio=false]{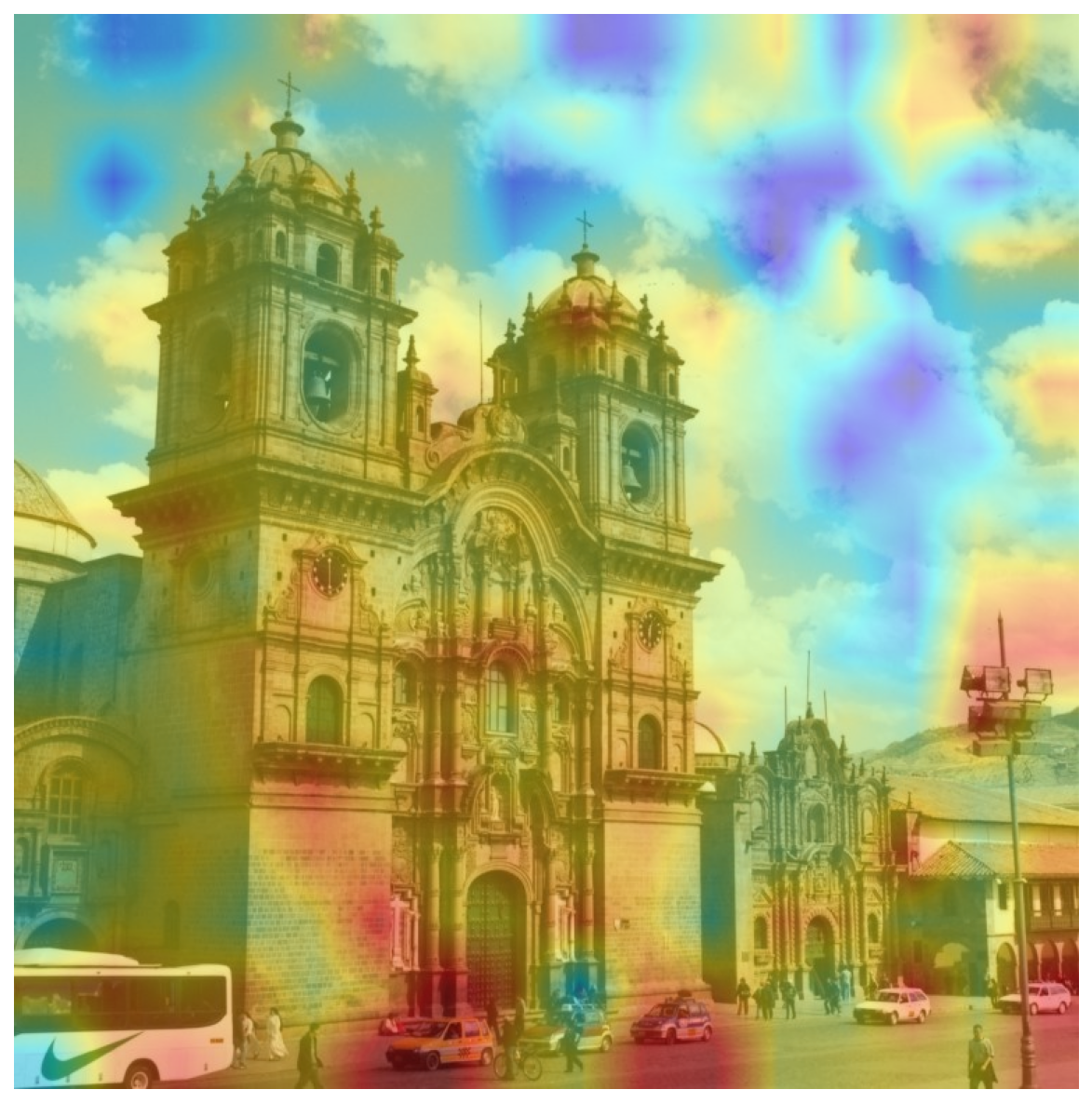} \\[-8pt]
        \multicolumn{1}{c}{\subcaptionbox{Florence}[0.22\linewidth]} &
        \multicolumn{1}{c}{\subcaptionbox{Paris}[0.22\linewidth]} &
        \multicolumn{1}{c}{\subcaptionbox{Seoul}[0.22\linewidth]} &
        \multicolumn{1}{c}{\subcaptionbox{Peru}[0.22\linewidth]}
    \end{tabular}
    \caption{\textbf{Natural image saliency maps taken from Im2GPS dataset from four locations: Florence, Paris, Seoul, Peru, shown without (top) and with (bottom) CLIP Surgery applied.} Confusing highlights on background elements, a known issue in CLIP-based models saliency maps, are noticeably reduced when CLIP Surgery is applied.}
    \label{fig:nosurgery-saliency}
\end{figure}

\begin{figure}[!htbp]
    \centering
   \includegraphics[width=0.8\linewidth]{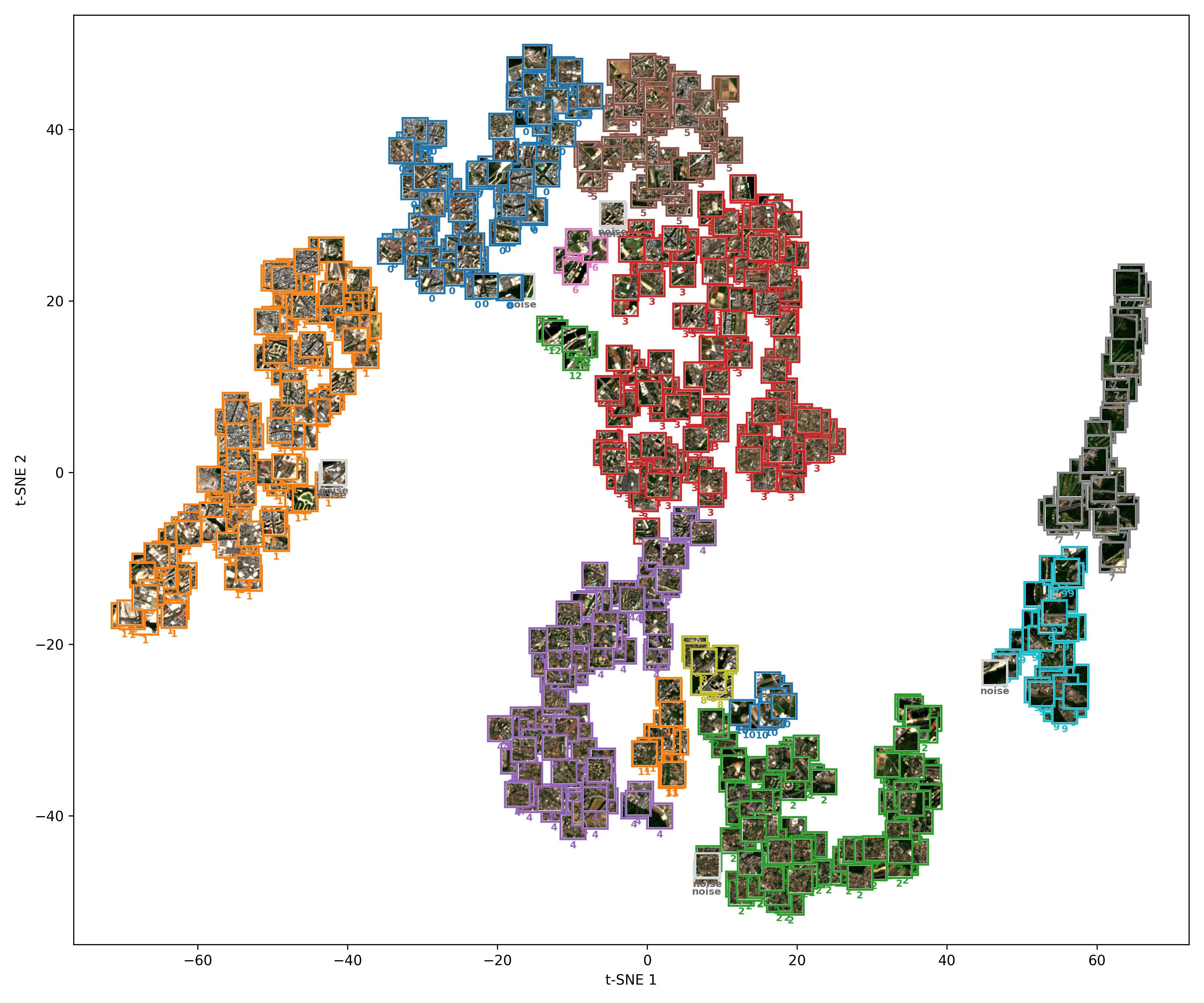}
    \caption{\textbf{Spatial clustering of salient region crops within the Paris area}. Crop embeddings were computed using a ResNet-50 pretrained on ImageNet weights, reduced to 2D via t-SNE, and clustered using DBSCAN. Colors denote distinct clusters. }
    \label{fig:paris-clusters}
\end{figure}

\end{document}